\documentclass{article}

\PassOptionsToPackage{numbers, compress}{natbib}
\usepackage[preprint]{neurips_2026}

\usepackage[utf8]{inputenc} % allow utf-8 input
\usepackage[T1]{fontenc}    % use 8-bit T1 fonts
\usepackage{hyperref}       % hyperlinks
\usepackage{url}            % simple URL typesetting
\usepackage{booktabs}       % professional-quality tables
\usepackage{amsfonts}       % blackboard math symbols
\usepackage{nicefrac}       % compact symbols for 1/2, etc.
\usepackage{microtype}      % microtypography
\usepackage{xcolor}         % colors

\usepackage{amsmath,amssymb,amsthm}
\usepackage{mathtools}
\usepackage{algorithm}
\usepackage{algpseudocode}
\usepackage{booktabs}
\usepackage{wrapfig}
\usepackage{natbib}
\usepackage{xcolor}
\usepackage{subcaption}
\usepackage{graphicx}
\usepackage{float}
\usepackage{multirow}
\usepackage{enumitem}
\usepackage{comment}
\usepackage{thmtools}
\usepackage{thm-restate}
\usepackage{tikz}
\usetikzlibrary{shadings}
\definecolor{densblue}{HTML}{0d4a8a}

\mathtoolsset{showonlyrefs}

\newcounter{dummy} 
\numberwithin{dummy}{section}

\newtheorem{remark}[dummy]{Remark}

\newtheorem{example}[dummy]{Example}

\DeclareMathOperator{\divg}{div}

\DeclareMathOperator{\vMF}{vMF}
\DeclareMathOperator*{\argmax}{arg\,max}

\newcommand{\R}{\mathbb{R}}
\newcommand{\E}{\mathbb{E}}

\newcommand{\dd}{\,\mathrm{d}}

\newcommand{\eH}[1]{\,\textcolor{gray}{\scriptsize(#1)}}

\usepackage{placeins}
\usepackage{etoolbox}
\pretocmd{\subsection}{\FloatBarrier}{}{}
\pretocmd{\subsubsection}{\FloatBarrier}{}{}
\usepackage[most]{tcolorbox}

\newenvironment{samplecard}[1]{%
  \par\bigskip\noindent
  \colorbox{gray!55!black}{\parbox{\dimexpr\linewidth-2\fboxsep}{\textcolor{white}{\bfseries\small\ #1}}}%
  \par\nobreak\smallskip
}{\par\medskip}

\newtcolorbox{sampletext}{%
  breakable,
  colback=gray!8, colframe=gray!25,
  boxrule=0.4pt, arc=2mm,
  fontupper=\small\raggedright,
  left=2.5mm, right=2.5mm, top=2mm, bottom=2mm,
}

\newcommand{\sampleheader}[2]{%
  \textbf{#1}\\
  #2\par\smallskip
}

\title{Spherical Flows for Sampling Categorical Data}

\author{%
  Jannis Chemseddine\thanks{Corresponding author.} \\
  Technische Universität Berlin \\
  \texttt{chemseddine@math.tu-berlin.de} \\
  \And
  Gregor Kornhardt \\
  Technische Universität Berlin \\
  \texttt{kornhardt@math.tu-berlin.de} \\
  \And
   Gabriele Steidl \\
  Technische Universität Berlin \\
  \texttt{steidl@math.tu-berlin.de} \\
}

\begin{document}

\maketitle

\begin{abstract}
We study the problem of learning generative models for discrete sequences in a continuous embedding space. Whereas prior approaches typically operate in Euclidean space or on the probability simplex, we instead work on the sphere $\mathbb S^{d-1}$. There the von Mises-Fisher (vMF) distribution induces a natural noise process and admits a closed-form conditional score. The conditional velocity is in general intractable. Exploiting the radial symmetry of the vMF density we reduce the continuity equation on $\mathbb S^{d-1}$ to a scalar ODE in the cosine similarity, whose unique bounded solution determines the velocity. The marginal velocity and marginal score on $(\mathbb S^{d-1})^L$ both decompose into posterior-weighted tangent sums that differ only by per-token scalar weights. This gives access to both ODE and predictor-corrector (PC) sampling. The posterior is the only learned object, trained by a cross-entropy loss. Experiments compare the vMF path against geodesic and Euclidean alternatives. The vMF path especially in combination with PC sampling significantly improves results on Sudoku, language modeling, and mathematical reasoning. The code is available on
\href{https://github.com/JChemseddine/SFCD}{\texttt{GitHub}}.
\end{abstract}

\section{Introduction}
\begin{wrapfigure}[17]{r}{0.42\textwidth}
    \vspace{-1.2em}
    \centering
    \includegraphics[width=0.42\textwidth]{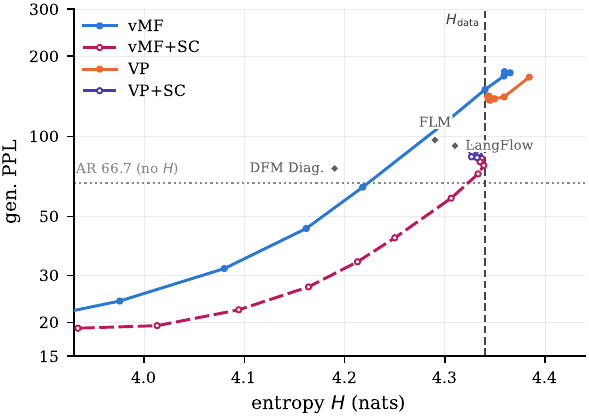}
        \caption{\textbf{LM1B} gen.\ PPL against entropy at $\mathrm{NFE} = 128$, sweeping $\varepsilon$; every curve is predictor-corrector. References that report an entropy are placed at it, AR is drawn as a horizontal line. Data entropy $H_{\mathrm{data}}$ as a vertical dashed line. Configuration in Appendix~\ref{sec:lm1b_grids}.}

    \label{fig:ppl_vs_entropy}
\end{wrapfigure}
Large language models (LLMs) based on autoregressive decoding are the prevailing approach to text generation. An alternative to sequential generation is to apply diffusion or flow-based generative modeling to produce the entire sequence at once. Such methods have proven remarkably successful for image generation, e.g. \cite{song2020score,lipman2022flow, geng2025meanflowsonestepgenerative}. Several works extend these methods to discrete data, such as text. As in the continuous setting, a forward process progressively corrupts an entire sequence and generation proceeds by learning to reverse this corruption.

The existing methods can be split into two families. \emph{Discrete diffusion models} \cite{austin2021structured, campbell2024generative, lou2023discrete,sahoo2024simple} operate directly on the finite state space, corrupting tokens by e.g. masking or replacing them according to a discrete Markov process. \emph{Continuous diffusion methods} instead embed tokens in a continuous space and define the noise process there. For example, Continuous Diffusion for Categorical Data (CDCD) \cite{dieleman2022continuous} adds Gaussian noise to learned embeddings in $\mathbb R^d$. 

In CDCD \cite{dieleman2022continuous}, the token embeddings $w_k$ are normalized to unit norm. The model produces a vector per position and predicts token probabilities via softmax of inner products with the $w_k$. This output decomposes into a direction on $\mathbb S^{d-1}$ and a magnitude. The direction determines the most likely token. The magnitude controls confidence. 

In this paper, we propose a continuous noise process directly on $\mathbb S^{d-1}$. We assign each token from a vocabulary a learned embedding $w_k$ on the unit sphere $\mathbb S^{d-1}$. Given a target sequence, a forward process corrupts each position independently by moving the embedding toward a reference distribution on $\mathbb S^{d-1}$. Generation reverses this process by integrating a velocity field on the product manifold $(\mathbb S^{d-1})^L$. 

What remains to specify is the conditional path $p_t(\cdot|w_k)$ at each position, which determines the marginal flow on $(\mathbb S^{d-1})^L$. Two conditional paths arise naturally on the sphere. The first is the geodesic interpolation, the analogue of linear interpolation in $\mathbb R^d$ \cite{chen2024flow}. We propose a new one based on the von Mises--Fisher (vMF) family of distributions on $\mathbb S^{d-1}$ indexed by a concentration parameter $\kappa$.

The vMF density $\varphi(x; w, \kappa) \propto \exp(\kappa\langle w, x\rangle)$ around mean $w \in \mathbb S^{d-1}$ with concentration $\kappa \ge 0$ is radially symmetric and depends on x only through the cosine similarity $\langle w, x\rangle$. While the Riemannian score follows directly from this form, the velocity must be derived from the continuity equation. We show that radial symmetry reduces the continuity equation of the vMF path on $\mathbb S^{d-1}$ to a scalar ODE in the cosine similarity (Theorem \ref{thm:conditional_velocity}). The solution of this ODE can be precomputed efficiently. Having both velocity and score in tractable form, the marginal velocity and the marginal score on $(\mathbb S^{d-1})^L$ decompose into per-position sums over the vocabulary weighted by the per-position posteriors $p_t^l(w_k|\mathbf x)$. The minimizer of the per-position cross-entropy loss is precisely this posterior (Proposition \ref{prop:loss}). The learned posteriors therefore determine both the velocity for ODE sampling and the score for predictor-corrector (PC) and SDE sampling. 

We validate our method on Sudoku-Extreme, on LM1B and OpenWebText for language modeling, and on TinyGSM for mathematical reasoning (Section~\ref{sec:exp}). Figure~\ref{fig:ppl_vs_entropy} previews the LM1B results, where vMF traces a smooth perplexity-entropy frontier under the predictor-corrector sampler, in contrast to VP. \\

\textbf{Contributions.}
1. For radially symmetric conditional paths on $\mathbb S^{d-1}$, the continuity equation reduces to a one-dimensional flux equation in the cosine similarity (Theorem \ref{thm:radial_continuity}). Applied to the vMF family, this yields the conditional velocity as the unique bounded solution of a scalar ODE (Theorem \ref{thm:conditional_velocity}). The solution factorizes as $\psi_t = \dot\kappa_t\,\tilde\psi_t$, where $\tilde\psi_t$ is schedule-independent and admits a numerically stable evaluation (Remark \ref{rem:46}).
\\[1ex]
2. We build a generative model on $(\mathbb S^{d-1})^L$ using both geodesic and vMF conditional paths. The per-position cross-entropy loss trains the model to output the marginal posteriors (Proposition \ref{prop:loss}). For the vMF path these posteriors determine the marginal velocity and the Riemannian score, giving access to ODE, SDE, and predictor-corrector sampling from a single model.
\\[1ex]
3. We evaluate the models on Sudoku-Extreme, LM1B, OpenWebText, and TinyGSM, comparing the
spherical paths against Euclidean baselines under ODE, SDE, and predictor-corrector
sampling (Section~\ref{sec:exp}). The experiments showcase the competitiveness of the
vMF path and analyze how self-conditioning, predictor-corrector sampler, and the sampling budget each
contribute.

For related work see Appendix~\ref{sec:related}. All proofs are in Appendix \ref{appa}.

\section{Background: Continuous Flows for Discrete Data} \label{sec:model}
%----------------------------------------------------------
To sample from a discrete data distribution \(p_{\mathrm{data}}\) via flow-based generation, we embed the discrete data into a continuous Riemannian manifold and construct a flow of measures. It has been shown, for example in \cite{lee2026flowmaplanguagemodels}, that for discrete data it is often advantageous to train a denoiser \(p_1(\cdot \mid \mathbf{x})\) with a cross-entropy loss. We therefore use the standard decomposition of the velocity as a conditional expectation, obtained by the law of total probability. We recall this construction here. The choice of manifold and conditional path is deferred to Section~\ref{sec:vM}.

\subsection{Setting and Continuous Embedding}
Let $\mathcal V \coloneqq \{v_1, \ldots, v_N\}$ be a vocabulary of tokens and $(\Omega, \mathcal A, \mathbb P)$ a probability space. Our aim is to sample from a discrete random variable $\mathbf Y = (Y^1, \ldots, Y^L) : \Omega \to \mathcal V^L$ given training samples. %{\color{red} In binary MNIST, $\mathcal V = \{0,1\}$ and $L = 28^2$; in language modeling, $\mathcal V$ is a tokenizer vocabulary with thousands of entries \cite{BST2026}.} 
For the concrete choice of $N$ and $L$, see Section \ref{sec:exp}.

To apply flow-based generative modeling, we embed $\mathcal V$ into a Riemannian manifold
$$\mathcal{M} \in \{\mathbb{R}^d,\, \mathbb{S}^{d-1}\}, \quad  d > 1,$$
and operate on its $L$-fold product $\mathcal{M}^L$. We work with extrinsic representations on $\mathbb{S}^{d-1}$, viewing its points as elements of $\mathbb{R}^d$ of unit norm. All norms and inner products in this paper are inherited from $\mathbb{R}^d$. The constructions below extend to general finite-dimensional smooth Riemannian manifolds, see \cite{AT2014, villani2008}.

\paragraph{Embedding and decoding.}
To each token $v_k \in \mathcal V$ we assign an element $w_k \in \mathcal M$, collected into $\mathcal W \coloneqq \{w_1, \ldots, w_N\} \subset \mathcal M$. This induces a token-wise embedding $W_E : \mathcal V^L \to \mathcal M^L$ defined by $W_E(\mathbf y)^l \coloneqq w_{y^l}$, for $\mathbf y \in \mathcal V^L$. The distribution induced by applying $W_E$ to $\mathbf Y$ is a $\mathcal W^L$-valued random variable $\mathbf W$ with probability mass function (pmf)
$$
p_{\mathrm{data}}(\mathbf w) \coloneqq \mathbb P(\mathbf W = \mathbf w),
$$
and this is the distribution we aim to sample from. The embeddings $\{w_k\}$ will be learnable parameters of the model, optimized jointly with the network in Section~\ref{sec:model_2}. To return token sequences, we equip the model with a decoder that maps a generated state on $\mathcal M^L$ back to $\mathcal V^L$. The specific decoder we use is defined in Section \ref{sec:model_2}.

\subsection{Conditional Flows of Measures}
We sample from $p_{\mathrm{data}}$ by constructing a flow of measures from a simple noise distribution to the data distribution. The standard machinery is summarized below, for details see \cite{albergo2023stochastic, chen2024flow, WS25}.

\paragraph{Measure Flows.}
Let $p_t : I \to \mathcal P(\mathcal M)$ be a curve of probability densities on $\mathcal M$ (with respect to the Riemannian volume measure) for $I = (0,1)$, and let $v : I \times \mathcal M \to \mathcal{TM}$ be a sufficiently smooth Borel-measurable vector field. The pair $(p_t, v_t)$ satisfies the \textit{continuity equation}
\begin{equation}\label{eq:ce}
\partial_t p_t + \mathrm{div}_{\mathcal M}(p_t v_t) = 0
\tag{CE}
\end{equation}
if and only if there exists a solution $\Phi : I \times \mathcal M \to \mathcal M$ of the \textit{flow ODE}
\begin{equation}\label{eq:flow_ode}
\partial_t \Phi(t, x) = v_t(\Phi(t, x)), \quad \Phi(0, x) = x,
\tag{Flow ODE}
\end{equation}
with $p_t = \Phi(t, \cdot)_\sharp p_0$. We can therefore sample from $p_1$ by drawing $x \sim p_0$ and integrating \eqref{eq:flow_ode} numerically up to $t = 1$. 

The same flow of measures admits a stochastic realization once the score $\nabla_{\mathcal M} \log p_t$ is available, which we defer to Appendix~\ref{sec:sde_app}. The differential operators $\nabla_{\mathcal M}, \mathrm{div}_{\mathcal M}, \Delta_{\mathcal M}$ for $\mathcal M = \mathbb S^{d-1}$ are recalled in Section~\ref{sec:vM}.

\medskip
For paths of measures on the product manifold $\mathcal M^L$, we construct $(p_t, v_t)$ in three steps: a conditional flow on $\mathcal M$, its product on $\mathcal M^L$, and the marginal flow obtained from $p_{\mathrm{data}}$.

\paragraph{1. Conditional Flows on $\mathcal M$.}
Fix a position $l \in \{1, \ldots, L\}$ and drop the superscript, writing $W : \Omega \to \mathcal W$ for $W^l$. Let $p_0 \in \mathcal P(\mathcal M)$ be easy to sample from (for instance, the standard Gaussian on $\mathbb R^d$ or the uniform measure on $\mathbb S^{d-1}$), and $Z \sim p_0$. For each fixed $w \in \mathcal W$, we suppose we can construct a path of conditional random variables $X_t \mid W = w$, $t \in [0,1]$, between $Z$ and $\delta_w$, with conditional densities $p_t(x \mid w)$ and velocity fields $v_t(x \mid w)$ fulfilling the conditional continuity equation
\begin{equation}\label{ce_cond}
\partial_t p_t(x \mid w) + \mathrm{div}_{\mathcal M}\bigl(p_t(x \mid w)\, v_t(x \mid w)\bigr) = 0, \quad t \in (0,1).
\end{equation}
This is the design choice of the method. We give two natural choices on $\mathbb S^{d-1}$ in Section~\ref{sec:vM}. For now, the standard $\mathbb R^d$ example serves as a reference.

\begin{example}[Linear interpolation on $\mathbb R^d$]\label{ex:M=R}
For $\mathcal M = \mathbb R^d$, the linear interpolation path $X_t \mid W = w \coloneqq tw + (1-t)Z$ with $Z \sim \mathcal N(0, \mathrm I_d)$ has conditional density $p_t(x \mid w) = \mathcal N(x; tw, (1-t)^2 \mathrm I_d)$ and velocity $v_t(x \mid w) = (w - x)/(1 - t)$, which fulfill \eqref{ce_cond} with $p_0(x \mid w) = \mathcal N(x; 0, \mathrm I_d)$.
\end{example}

\paragraph{2. Conditional Flows on $\mathcal M^L$.}
Conditional flows on single positions combine into a conditional flow on $\mathcal M^L$ by taking products.

\begin{restatable}{proposition}{productCE}\label{prop:multi}
Let $\mathbf w = (w^l)_{l=1}^L \in \mathcal W^L$ be fixed, and let $p_t^l(\cdot \mid w^l)$ be absolutely continuous measures on $\mathcal M$ with velocities $v_t^l(\cdot \mid w^l)$ satisfying the conditional continuity equation~\eqref{ce_cond} on $\mathcal M$ for each $l = 1, \ldots, L$. Then the product density on $\mathcal M^L$,
\begin{equation}\label{prod_p}
p_t(\mathbf x \mid \mathbf w) = \prod_{l=1}^L p_t^l(x^l \mid w^l), \qquad \mathbf x = (x^1, \ldots, x^L),
\end{equation}
with the velocity field $v_t(\mathbf x \mid \mathbf w) \coloneqq \bigl(v_t^1(x^1 \mid w^1),...,v_t^L(x^L \mid w^L)\bigr)$ satisfies the continuity equation on $\mathcal M^L$ for $t \in (0,1)$.
\end{restatable}

The proof is given in Appendix~\ref{appa}. Related factorization results appear in the generator-matching framework \cite{holderrieth2025generatormatchinggenerativemodeling} and in discrete flow matching \cite{lipman2024flowmatchingguidecode}, for general measures, see \cite{CJDS2026,DCFS2025}.

\paragraph{3. Marginal Flow on $\mathcal M^L$.}
By the law of total probability, the unconditional density on $\mathcal M^L$ at time $t$ is
\begin{equation}\label{final}
p_t(\mathbf x) = \sum_{\mathbf w \in \mathcal W^L} p_t(\mathbf x \mid \mathbf w)\, p_{\mathrm{data}}(\mathbf w),
\end{equation}
which interpolates between $p_0$ and $p_{\mathrm{data}}$ as $t$ goes from $0$ to $1$. Direct evaluation is intractable since $p_{\mathrm{data}}$ is unknown. The associated marginal velocity
\begin{equation}\label{velo}
v_t(\mathbf x) = \sum_{\mathbf w \in \mathcal W^L} v_t(\mathbf x \mid \mathbf w)\, p_t(\mathbf w \mid \mathbf x), \quad p_t(\mathbf w \mid \mathbf x) = p_t(\mathbf x \mid \mathbf w)\,\frac{p_{\mathrm{data}}(\mathbf w)}{p_t(\mathbf x)},
\end{equation}
fulfills the continuity equation \eqref{eq:ce} on $\mathcal M^L$ \cite{WS25}. By \eqref{prod_p}, its components $v_t = (v_t^l)_{l=1}^L$ simplify by marginalization to
\begin{align}%\label{velo}
v_t^l(\mathbf x) &= \sum_{\mathbf w \in \mathcal W^L} v_t^l(x^l \mid w^l)\, p_t(\mathbf w \mid \mathbf x) 
= 
\sum_{w^l \in \mathcal W} v_t^l(x^l \mid w^l) \!\! \sum_{\substack{\mathbf w \in \mathcal W^L \\
\mathbf w^l = w^l}}\!\! p_t(\mathbf w \mid \mathbf x)
\\
&= \sum_{w^l \in \mathcal W} v_t^l(x^l \mid w^l)\, p_t^l(w^l \mid \mathbf x), 
\end{align}
where we group sequences $\mathbf w \in \mathcal W^L$ by their $l$-th component, with $w^l$ first denoting that fixed $l$-th component and then ranging over $\mathcal W$ in the outer sum. 
The same per-position reduction applies to the score. Differentiating \eqref{final} with respect to $x^l$, only the $l$-th factor of $p_t(\mathbf x \mid \mathbf w)$ in \eqref{prod_p} contributes, so
\begin{align}\label{score}
\nabla_{\mathcal M, x^l} \log p_t(\mathbf x)= \sum_{w^l \in \mathcal W} \nabla_{\mathcal M, x^l} \log p_t^l(x^l \mid w^l)\, p_t^l(w^l \mid \mathbf x). 
\end{align}

\begin{example}\label{bsp2}
Continuing Example \ref{ex:M=R} with $\mathcal M^L = \mathbb R^{Ld}$, 
 the path
\begin{equation}\label{lin_path}
\mathbf X_t = t \mathbf W + (1-t) \mathbf Z, \quad \mathbf Z \sim \mathcal N(0, {\rm{I}}_{Ld})
\end{equation}
has laws \eqref{final}.
The components of the corresponding vector field 
\eqref{velo} read as
\begin{equation}
v_t^l (\mathbf x) = 
\sum_{w^l \in \mathcal W} \frac{w^l - x^l}{1-t}
p_t^l (w^l|\mathbf  x).
\end{equation}
The conditional density $p_t^l(\cdot \mid w^l) = \mathcal N(\cdot, tw^l, (1-t)^2 {\rm{I}}_d)$  has score
$$
\nabla_{x^l} \log p_t^l(x^l \mid w^l) = \frac{tw^l - x^l}{(1-t)^2}.
$$
Substituting into \eqref{score}, the score of the unconditional density at position $l$ is
\begin{equation}
\nabla_{x^l} \log p_t(\mathbf x) = \frac{1}{(1-t)^2} \Big( t \sum_{w^l \in \mathcal W} p_t^l(w^l|\mathbf x)\, w^l - x^l \Big).
\end{equation}
\end{example}

\paragraph{The posterior is the only learned object.}
In summary, both the velocity field \eqref{velo} and the score \eqref{score} of the unconditional density are weighted sums over $\mathcal W$ with the same marginal conditional pmfs $p_t^l(w^l|\mathbf x)$.
The conditional quantities $v^l_{t}(x^l|w^l)$ and $\nabla_{x^l} \log p_t^l(x^l|w^l)$ are determined by the choice of noise process.
The marginal posteriors $p_t^l(w^l|\mathbf x)$ must be learned.

What remains to specify is the conditional path $p_t(\cdot \mid w)$ on $\mathcal M$. In Section~\ref{sec:vM}, we work on $\mathcal M = \mathbb S^{d-1}$ and consider two natural choices: geodesic interpolation and a path through the von Mises-Fisher family. The latter yields a tractable conditional velocity via a radial-symmetry reduction and an informative range that scales correctly with the embedding dimension.

%--------------------------------------------------------
\section{Conditional von Mises-Fisher Paths}\label{sec:vM}
%--------------------------------------------------------

So far, we have only given concrete examples for $v^l_{t}(x^l|w^l)$ on $\mathcal M = \mathbb R^d$.  Assuming that the weights $w_k$ are normalized, i.e. $\|w_k\|=1$, $k=1,\ldots,N$, the target embeddings live on the $(d-1)$-sphere, with $d>1$. We propose to use a path of measures that stays on this manifold.

\subsection{Background on the Sphere}
\paragraph{Differential operators.}For the sphere $\mathbb S^{d -1} \coloneqq \{x \in \mathbb R^d: \|x\| = 1\}$, the geodesic distance between two points $x,z\in \mathbb S^{d -1}$ is given by
${\rm d} (x,z) \coloneqq \theta = \arccos(\langle x,z \rangle)$. The tangent space at a point $x\in \mathbb S^{d -1}$ is 
$$\mathcal T_x \mathbb S^{d -1} = \{v \in \R^d : \langle v, x \rangle  = 0\}$$
and the orthogonal projection onto $\mathcal T_x \mathbb S^{d -1}$ reads as 
$$%\begin{equation}\label{eq:proj_tan}$
    \mathrm{P}_x (w) := w - \langle w,x \rangle \, x,  \quad w \in \mathbb R^d
.$$
Let $\tilde g : \R^d \to \R$ be sufficiently smooth and $g = \tilde g\big|_{\mathbb S^{d -1}}$.
Then the Riemannian gradient of $g$ on $\mathbb S^{d -1}$ is  the orthogonal projection of the Euclidean gradient $\nabla  g$ onto $\mathcal T_x \mathbb S^{d -1}$, i.e.,
\begin{equation}
    \nabla_{\mathbb S^{d -1}} g(x) = \mathrm{P}_x (\nabla \tilde g(x)) 
    = \nabla \tilde g(x) - \langle \nabla \tilde g(x), x\rangle x.
\end{equation}
For a differentiable vector field $\tilde u : \R^d \to \R$ and $u = \tilde u\big|_{\mathbb S^{d -1}}$, the divergence is defined by $$\text{div}_{{\mathbb S^{d -1}}} u (x) = \text{tr} \left(\nabla \tilde u (x) \right) - x^\top   \nabla \tilde u(x)  \, x$$
and the Laplace-Beltrami operator becomes $$\Delta_{\mathbb S^{d -1}} g (x) = \text{tr} \left(({\rm I}_d - x x^\top) \nabla^2 \tilde g(x) \right) - (d-1) \langle \nabla \tilde g(x),x \rangle.$$
In the rest of this paper, we always assume that functions, resp. vector fields on the $(d-1)$-sphere have a smooth extension to functions, resp. vector fields living in the whole $\mathbb R^d$ and we
skip the 'tilde'.

\begin{remark}[Geodesic Path]\label{rem:geo_path}
Using the geodesic between two points, one can construct the slerp path   
\begin{equation} \label{int_geo}
    X_t = \frac{\sin((1 - t)\theta)}{\sin \theta} X_0 + \frac{\sin(t \theta)}{\sin \theta} X_1, \quad t \in [0,1],
    \end{equation}
where $\theta \coloneqq \arccos(\langle X_0,X_1 \rangle)$, $X_0 \sim \mathcal U(\mathbb S^{d-1})$. %and ${\rm d}(x_1,x_t) = \arccos(\langle  x_1,x_t\rangle) = \big((1-t)\theta\big)$.
%not unit speed, speed is theta
With the linear interpolation on $\mathbb R^d$
  from Examples~\ref{ex:M=R} and~\ref{bsp2} in mind, this is its spherical counterpart. This path is sometimes called spherical linear interpolation \textbf{\rmfamily(slerp)}. Using this path for generative modeling via flows is a straightforward application of Riemannian Flow Matching~\cite{chen2024flow}.  

\end{remark}

In this paper, we propose another path based on special radially symmetric functions on the sphere.

\textbf{Radial symmetry on the sphere.}
A  function $p: \mathbb S^{d -1}\to \mathbb R_{\ge 0}$ is called \textit{radially symmetric around} $w \in  \mathbb S^{d -1}$, if there exists a  function $\bar p: [-1,1] \to \mathbb R$ such that
\begin{equation} \label{rad_ab}
p(x) = \bar p(\langle w,x\rangle) \quad \text{for all} \quad x \in \mathbb S^{d-1}.
\end{equation}
Note that $|\langle w,x \rangle| \le \|w\| \, \|x \| = 1$ with equality iff $x=w$.
Extending $\bar p$  to the whole $\mathbb R$, the map $x \mapsto p (\langle w,x \rangle)$ extends to all of $\R^d$ and 
$$\nabla_{\mathbb S^{d -1}} p(x) 
    = \mathrm{P}_x\big( \bar p'(\langle w,x \rangle) \, w \big)
    = \bar p'(\langle w,x \rangle)\,  \mathrm{P}_x(w).
    $$
A tangent vector field $v : \mathbb S^{d -1} \to \mathcal T_x \mathbb S^{d -1}$ is \textit{radially symmetric  around} $w \in \mathbb S^{d -1}$,
if there exists $\psi : [-1,1] \to \R$ such that
\begin{equation}\label{eq:radial_vf}
    v(x) = \psi\big(\langle w,x\rangle \big) \,\mathrm{P}_x(w).
\end{equation}
Given a random variable $X$ with radially symmetric density $p$ around $w$,
the random variable $S = w^\top X \in [-1,1]$ 
has the density $C \, f$ with
\begin{equation}\label{eq:f_mu_def}
    f(s) := (1-s^2)^{(d-3)/2}\,\bar p(s),
    \qquad s \in (-1,1),
\end{equation}
and normalizing factor $C$. In particular, we have for $d=3$ that $f = \bar p$.
The following theorem  shows that the continuity equation of a radially symmetric density on the sphere
can be reduced to the one-dimensional flux  function on the
interval $[-1,1]$. 

\begin{restatable}{theorem}{radialcontinuity}\label{thm:radial_continuity}
Let $(p_t, v_t)$, $t\in I$ be a sequence of radially symmetric pairs of densities and velocities of the form $p_t(x) = \bar p_t(\langle x,w \rangle)$, resp. 
$v_t(x) = \psi_t(\langle x,w\rangle)\mathrm{P}_x(w)$,
where 
$\bar p_t \in C^1[I \times (-1,1)]$ 
and 
$\psi_t \in C^1(-1,1)$ for every $t \in I$. Let $f_t$, $t \in I$ be the corresponding densities defined by \eqref{eq:f_mu_def}.
Then $(p_t, v_t)$, $t\in I$ satisfy the continuity equation
\begin{equation}\label{eq:cont_sphere}
    \partial_t p_t + {\rm{div}}_{\mathbb S^{d-1}}(p_t\, v_t) = 0,
    \quad t \in I, x \in \mathbb S^{d-1}\setminus \{\pm w\}
\end{equation}
if and only if 
$(f_t, \psi_t)$ satisfy the one-dimensional flux equation
\begin{equation}\label{eq:1d_flux}
    \partial_t f_t + \partial_s\big(f_t \, \psi_t \cdot (1-s^2)\, \big) = 0, \quad t \in I, s \in (-1,1).
\end{equation}
%{\color{blue} If in addition $\psi_t \in C[-1,1]$ and $\bar p_t \in C^1[I \times [-1,1]]$, then the equivalence extends to $[-1,1]$.}
\end{restatable}

\subsection{von Mises-Fisher Distribution and Path}
%--------------------------------------------- 
The \textit{von Mises-Fisher distribution} with mean $w \in  \mathbb S^{d -1}$ and concentration $\kappa \geq 0$, denoted by $\text{vMF}(w,\kappa)$, has the density
\begin{equation}\label{eq:vmf_density}
    \varphi (x;\, w, \kappa) = C_d(\kappa)\,\exp(\kappa\,\langle w,x \rangle),
    \tag{vMF}
\end{equation}
where 
$$C_d(\kappa) \coloneqq \kappa^{d/2-1}/\bigl((2\pi)^{d/2}\, \mathcal I_{d/2-1}(\kappa)\bigr)$$ 
is the normalization constant
and 
$\mathcal I_{d/2-1}$ is the modified Bessel function of the first kind and order $d/2-1$.
At $\kappa = 0$ the distribution is uniform on $\mathbb S^{d -1}$, and as $\kappa \to \infty$ it concentrates to a point mass at $w$, see  Figure \ref{fig:vmf_sphere}.
The normalization constant $C_d(\kappa)$ depends only on $\kappa$ and not on the mean direction $w$.
The \textit{mean resultant length} of the vMF distribution is the Bessel ratio 
$$A_d(\kappa) :=\E_{x \sim \vMF(w,\,\kappa)}[\langle w,x \rangle] = \frac{\mathcal I_{d/2}(\kappa)}{\mathcal I_{d/2-1}(\kappa)},$$
which is equal to the expected cosine similarity, see \cite[Chapter 9]{mardia2000directional}.
In particular, 
$A_d$ is strictly monotone increasing, 
$A_d(0) = 0$, $A_d(\kappa) = \kappa / d + O(\kappa^3)$ as $\kappa \to 0$ and
$A_d(\kappa) \to 1$ as $\kappa \to \infty$.

\begin{figure}[t]
    \centering
    \def\R{1.9}
    \def\tilt{25}
    \def\spc{4.6}
    % Render a vMF density of concentration κ (=#3) over a (theta, phi) grid,
    % with #1=x-offset, #2=label, #4 unused (kept for backward signature).
    \newcommand{\vmfsphere}[4]{%
    \begin{scope}[shift={(#1,0)}]
        \node[above, font=\normalsize] at (0,\R+0.4) {#2};
        \pgfmathsetmacro{\muy}{\R*sin(\tilt)}
        \begin{scope}
            \clip (0,0) circle (\R);
            \fill[white!97!yellow] (0,0) circle (\R);
            \shade[inner color=white, outer color=gray!18, opacity=0.5]
                (0,0) circle (\R);
            \shade[inner color=white, outer color=white!0, opacity=0.35]
                (-0.4*\R, 0.55*\R) circle (0.7*\R);
            \pgfmathsetmacro{\gradR}{#3*\R}%
            \shade[inner color=densblue, outer color=densblue!0!white, opacity=#4]
                (0,\muy) circle [radius=\gradR];
            \foreach \colat in {30,60,90,120,150} {%
                \pgfmathsetmacro{\rx}{\R*sin(\colat)}%
                \pgfmathsetmacro{\ry}{\R*sin(\colat)*cos(\tilt)}%
                \pgfmathsetmacro{\cy}{\R*cos(\colat)*sin(\tilt)}%
                \draw[gray!45, line width=0.28pt]
                    (0,\cy) ellipse [x radius=\rx, y radius=\ry];
            }%
            \foreach \lon in {0,30,...,150} {%
                \draw[gray!45, line width=0.28pt]
                    plot[variable=\t, domain=0:180, samples=80, smooth]
                    ({\R*sin(\t)*cos(\lon)},
                     {\R*sin(\t)*sin(\lon)*cos(\tilt) + \R*cos(\t)*sin(\tilt)});
            }%
        \end{scope}
        \draw[gray!45, line width=0.4pt] (0,0) circle (\R);
        \fill[white] (0,\muy) circle (2pt);
        \fill[red!70!black] (0,\muy) circle (1.3pt);
        \node[right, font=\footnotesize, inner sep=1.5pt] at (0.12,\muy+0.1)
            {$w$};
    \end{scope}
    }
    \begin{tikzpicture}[scale=0.6, transform shape]
    \draw[->, line width=1pt, black] (0.6*\R,\R+1.0) -- ({3*\spc - 0.6*\R},\R+1.0)
        node[midway, above=2pt, font=\normalsize] {$\kappa \to \infty$};
    \vmfsphere{0}{}{4.0}{0.45}
    \vmfsphere{\spc}{}{1.2}{0.72}
    \vmfsphere{2*\spc}{}{0.48}{0.88}
    \vmfsphere{3*\spc}{}{0.18}{0.97}
    \end{tikzpicture}
    \caption{
    Illustration of von Mises--Fisher density on $\mathbb S^{2}$ for increasing concentration $\kappa$. At $\kappa = 0$ (left) the density is uniform. As $\kappa$ grows the mass concentrates around the mean direction $w$, collapsing to a point mass at $w$ as $\kappa \to \infty$. }
    %Shadings are illustrative; only the qualitative shape is rendered to scale.}}
    \label{fig:vmf_sphere}
\end{figure}
In the following, let
$\kappa_t \colon [0,1) \to [0,\infty)$
be a continuously differentiable, strictly monotone increasing bijection with
$\kappa_0 = 0$, so that $\kappa_t \to \infty$ as $t \to 1$ and $p_t(\cdot|w)$ converges to the point mass at $w$; the schedule will be specified later. 
Then we consider, for fixed $w \in \mathcal W$,  a sequence of random variables $X_t| W = w$ on $\mathbb S^{d-1}$, $t \in [0,1]$ with laws forming the \textit{conditional vMF path}
\begin{equation} \label{eq:vMFpath}
p_t (\cdot|w) \coloneqq \varphi(\cdot;w,\kappa_t) = \bar p_t(\langle w , \cdot \rangle).
\end{equation}
Clearly, $p_t$ are radially symmetric  functions
and the corresponding function $f_t$
in \eqref{eq:f_mu_def} reads as
\begin{equation}\label{eq:f_vMF}
f_t (s)  =  C_d(\kappa_t)\,(1-s^2)^{(d-3)/2}\,\exp(\kappa_t \,s), \quad s \in (-1,1).
\end{equation}

%--------------------------------
\begin{restatable}{lemma}{psiode}\label{lem:psi_ode}
Let $f_t$, $t \in I$ be defined by \eqref{eq:f_vMF}.
Then $\psi_t$, $t \in I$ solves the linear ODE \begin{equation}\label{eq:psi_ode}
    (1 - s^2)\, \psi_t' + \left( \kappa_t(1-s^2) - (d-1)s\right)\psi_t =  \left(A_d(\kappa_t) - s \right) \dot{\kappa}_t, \quad s \in (-1,1)
\end{equation}
if and only if $(f_t,\psi_t)$, $t \in I$ fulfill the flux equation \eqref{eq:1d_flux}.
\end{restatable}

We have to deal with the possible singularity of the ODE~\eqref{eq:psi_ode} at $x = \pm 1$. This can be handled similarly as for the Kac process, see e.g. \cite{DCFS2025},
by the following lemma.

\begin{restatable}{lemma}{psiwellposed}\label{prop:psi_wellposed}
For any $t \in I$, the ODE~\eqref{eq:psi_ode} has a unique solution $\psi_t$ in  $C[-1,1]$, which is automatically also in $C^\infty{(-1,1)}$ with
boundary values $\psi_t(1) = \frac{\left(1 - A_d(\kappa_t)\right) \dot{\kappa_t}}{d-1},$ and $  \psi_t(-1) = \frac{\left(1 + A_d(\kappa_t)\right)\dot \kappa_t}{d-1}$.
\end{restatable}

As a direct consequence of Theorem \ref{thm:radial_continuity}, Lemma \ref{lem:psi_ode} and Lemma \ref{prop:psi_wellposed}, we obtain access to the conditional velocities in our key formula \eqref{velo} for the vMF path. %{\color{blue}Note that the vMF radial density $\bar p_t(s) = C_d(\kappa_t)e^{\kappa_t s}$ is in $C^1[I \times [-1,1]]$, so the equivalence holds on $[-1,1]$ by Remark \ref{rem:boundary}.}

\begin{restatable}{theorem}{conditionalvelocity}\label{thm:conditional_velocity}
Let $p_t(\cdot| w)$ be the conditional vMF path defined by \eqref{eq:vMFpath} and let $\psi_t$, $t \in I$ be the 
unique $C[-1,1]$ solutions of the ODE~\eqref{eq:psi_ode}.
Then $p_t(\cdot| w)$ together with the velocities
\begin{equation}\label{eq:cond_velocity}
    v_t(x| w) = \,\psi_t(\langle w, x\rangle)\,\mathrm{P}_x(w)
\end{equation}
fulfill  the continuity equation ~\eqref{eq:cont_sphere} on $\mathbb S^{d-1}$.
\end{restatable}

\begin{remark} \label{rem:46}
The conditional velocity in Theorem \ref{thm:conditional_velocity} is expressed in terms of $\psi_t$, which is defined as the unique bounded solution of the ODE \eqref{eq:psi_ode}. For practical computation, we will use the following expression. The proof of Lemma \ref{prop:psi_wellposed} uses the integral representation \eqref{h1}. By \eqref{eq:f_vMF}, this can be rewritten as 
\begin{equation}\label{eq:psi_tilde}
\psi_t(s) =  \dot\kappa_t\,\tilde\psi_t(s) \quad \text{with} \quad
\tilde\psi_t(s) \coloneqq -\frac{\int_{-1}^{s} (y - A_d(\kappa_t))\, f_t(y)\, \mathrm{d} y}{f_t(s)\,(1-s^2)} .
\end{equation}
The function $\tilde\psi$ depends only on $\kappa_t$ and not on the schedule. This representation is both schedule-independent and amenable to numerically stable evaluation, see Section~\ref{sec:exp} and Section~\ref{sec:paths_app} for the implementation.
\end{remark}

The geodesic interpolation from Remark \ref{rem:geo_path} and the vMF path from Theorem \ref{thm:conditional_velocity} both define conditional paths starting from the uniform distribution on $\mathbb S^{d-1}$. We want to highlight that they differ in how quickly the noisy sample reveals the target with increasing dimension $d$. We quantify this by comparing the expected cosine similarity 
%$\mathbb E[\langle w, x_t\rangle]$ 
under both paths.

\begin{restatable}{proposition}{signal}%[Signal comparison]
\label{prop:signal}
Let $X_0 \sim \mathcal U(\mathbb S^{d-1})$ and  $w \in \mathbb S^{d-1}$ be arbitrary fixed. Then
the conditional geodesic interpolation path {\rm{(slerp)}} in \eqref{int_geo} fulfill
\begin{align}\label{eq:geo_signal}
&\mathbb E_{x_t \sim {\rm slerp}_t(x_0,w)}\big[\langle w, x_t\rangle\big] \to \sin\!\big(\tfrac{\pi}{2}\,t\big) \quad \text{as} \quad d \to \infty, 
\end{align}
the vMF path \eqref{eq:vMFpath},
\begin{align} \label{eq:vmf_signal}
&\mathbb E_{x_t \sim \vMF(w, \kappa_t)}\big[\langle w, x_t\rangle\big] = A_d(\kappa_t) = \frac{\kappa_t}{d} + O(\kappa_t^3).
\end{align}
\end{restatable}

Under geodesic interpolation, the expected cosine similarity at any fixed $t > 0$ is bounded away from zero independently of $d$. Under the vMF path, reaching the same signal level requires $\kappa_t$ proportional to $d$. The informative range of the noise parameter therefore grows with $d$.

The vMF construction is one instance of a broader family of radially symmetric paths; two further instances (power spherical and spherical Cauchy) are developed in Appendix~\ref{app:further_paths}.

%-----------------------------------------------------------------
\section{Generative Model} \label{sec:model_2}
%-----------------------------------------------------------------
We now specify how to build a generative model on $(\mathbb S^{d-1})^L$ from the conditional paths constructed in Section~\ref{sec:model}.

\subsection{Learning} % the posterior}
By \eqref{velo} and \eqref{score}, both the velocity field and the score of the unconditional density are determined by the marginal conditional pmfs $p_t^l(w^l|\mathbf x)$. We learn these pmfs using a neural network.

Let $T_{\theta,t}: \mathcal M^L \to (\R^d)^L$, $\mathbf x \mapsto \hat{\mathbf x} \coloneqq T_{\theta,t}(\mathbf x)$ be a backbone network. We define a factorized approximation 
\begin{equation}
    p_{\theta,t} (\mathbf w|\mathbf x) \coloneqq \prod_{l=1}^L p^l_{\theta,t}( w^l|\hat x^l),
\end{equation}
where for $w_k \in \mathcal W$,
\begin{equation}\label{eq:softmax}
    p_{\theta,t}^l ( w_k|\hat x^l) 
    = \frac{\exp(s^l_k)}{\sum_{j=1}^N \exp(s^l_j)},
    \quad
    s^l_k \coloneqq \langle w_k,\hat x^l \rangle + b_k
\end{equation}
with learnable biases $b_k \in \mathbb R$. 

Recall that the Kullback--Leibler divergence of two pmfs $p, q$ on a discrete space of size $N$ with $p(k) = 0$ whenever $q(k) = 0$ is $\mathrm{KL}(p \| q) = \sum_{k=1}^N \big( p(k) \log p(k) - p(k) \log q(k) \big)$
and their cross-entropy and entropy by $\mathrm{CE}(p, q) = -\sum_{k=1}^N p(k) \log q(k) = \mathrm{KL}(p \| q) + \mathrm H(p), \qquad \mathrm H(p) = -\sum_{k=1}^N p(k) \log p(k).$
Since $\mathrm{KL}(p \| q) \geq 0$ with equality if and only if $p = q$, the cross-entropy $\mathrm{CE}(p, q)$ is minimized in $q$ exactly at $q = p$.

Having this in mind, we train $p_\theta$ with the \textit{cross-entropy loss}
\begin{equation}\label{eq:ce_loss}
    \mathcal L (\theta) 
    = 
    - \mathbb E_{t \sim \mathcal U(0,1),\, \mathbf w \sim p_{\text{data}},\, \mathbf x_t \sim p_t(\cdot|\mathbf w)} 
    \left[ \sum_{l=1}^L \log p^l_{\theta,t} (w^l|\mathbf x_t) \right].
\end{equation}
%-------------------------------------
\begin{restatable}{proposition}{loss} \label{prop:loss}
The minimizer of \eqref{eq:ce_loss} is the per-position marginal posterior $p_t^l(\cdot|\mathbf x)$ from \eqref{velo}.
\end{restatable}

A short derivation is given in Appendix~\ref{appa}.

The embeddings $\{w_k\}_{k=1}^N$ are learned jointly with $\theta$, with each $w_k$ constrained to $\mathbb S^{d-1}$. Gradients of \eqref{eq:ce_loss} flow into $w_k$ through the conditional path $p_t(\cdot \mid w^l)$ and through the decoder logits \eqref{eq:softmax}. The pmf $p_{\mathrm{data}}$ on $\mathcal W^L$ is the pushforward of the fixed law of $\mathbf Y$ on $\mathcal V^L$ under $W_E$. For any $\mathcal W$, Proposition~\ref{prop:loss} characterizes the optimal $\theta$.

\begin{remark}[Noise schedule]\label{rem:schedule}
The loss \eqref{eq:ce_loss} draws $t \sim \mathcal U(0, 1)$ uniformly. In practice we replace $\mathcal U(0, 1)$ with a learned distribution following CDCD \cite{dieleman2022continuous}. A learnable monotone $\tilde F: [0, 1] \to \mathbb R_{\geq 0}$ is fit to the per-sample cross-entropy, and the normalized curve $F = \tilde F / \tilde F(1)$ serves as a CDF on $[0, 1]$ from which $t$ is drawn by inverse-transform sampling. We use the same scheme for every conditional path in our experiments.  Full details are in Appendix~\ref{sec:schedule_app}.

\end{remark}

%\paragraph{Marginal velocity and score of the vMF mixture path}
Once the network approximates the per-position marginal posteriors,
$p_t^l(\cdot|\mathbf x) \approx p_{\theta,t}^l (\cdot|\hat x_t^l),$
the marginal velocity \eqref{velo} and score \eqref{score} of the vMF path follow.
By Theorem \ref{thm:conditional_velocity}, the conditional velocity is $v_t(x^l|w) = \psi_t(\langle w, x^l\rangle)\, {\rm{P}}_{x^l}(w)$
and substituting  $\psi_t = \dot\kappa_t\,\tilde\psi_t$ from \eqref{eq:psi_tilde} into \eqref{velo}, the \textit{marginal velocity at position} $l$ becomes
\begin{equation}\label{eq:marginal_velo_vmf}
v_{\theta,t}^l(\mathbf x) = \dot\kappa_t \sum_{k=1}^N p_{\theta,t}^l(w_k|\mathbf x)\,\tilde\psi_t(\langle w_k, x^l\rangle)\,{\rm{P}}_{x^l}(w_k).
\end{equation}
For the conditional score, recall that by \eqref{eq:vmf_density} it holds
$\log p_t^l(x^l|w) = \log C_d(\kappa_t) + \kappa_t\langle w, x^l\rangle$. The normalization constant does not depend on $x^l$. 
Projecting the Euclidean gradient $\kappa_t w$ onto the tangent space gives $\nabla_{\mathbb S^{d-1}} \log p_t^l(x^l|w) = \kappa_t\,{\rm{P}}_{x^l}(w)$.
Substituting into \eqref{score}, the \textit{marginal score at position} $l$ becomes
\begin{equation}\label{eq:marginal_score_vmf}
  \nabla_{\mathbb S^{d-1}, x^l}  \log p_{\theta, t}(\mathbf x) = \kappa_t \sum_{k=1}^N p_{\theta,t}^l(w_k|\mathbf x)\,{\rm{P}}_{x^l}(w_k).
\end{equation}
Both the marginal velocity and the marginal score at position $l$ are posterior-weighted sums of tangent projections ${\rm{P}}_{x^l}(w_k)$, differing only in the scalar weights. The velocity carries the schedule-dependent prefactor $\dot\kappa_t\,\tilde\psi_t(\langle w_k,x^l\rangle)$, while the score carries $\kappa_t$.
In particular, the posteriors $p_t^l(w_k|\mathbf x)$ are the only learned quantities needed to evaluate velocity, score, and any linear combination of the two. %We note that the score \eqref{eq:marginal_score_vmf} is bounded in norm by $\kappa_t \le \kappa_{\max}$ uniformly in $\mathbf x$, in contrast to the Euclidean case (Example \ref{bsp2}) where the score diverges as $t \to 1$.

\subsection{Sampling}
The pair $(p_t,v_t)$ in \eqref{final} and \eqref{velo} with velocity components 
\eqref{eq:marginal_velo_vmf}  satisfies the continuity equation \eqref{eq:ce} on $(\mathbb S^{d-1})^L$. We focus on ODE sampling and predictor-corrector sampling. The score \eqref{eq:marginal_score_vmf} also makes SDE sampling available, which we outline in Appendix~\ref{sec:sde_app} and evaluate in Appendix~\ref{sec:sde_results} (Tables~\ref{tab:lm1b_sde}, \ref{tab:owt_sde}, and~\ref{tab:gsm8k_sde}). Driven by the same posterior, the SDE tracks the predictor-corrector results at matched budgets, with the diffusion scale acting as the same quality--entropy knob.

\paragraph{Flow ODE (Euler).} The associated flow ODE
\begin{equation}
\partial_t \Phi^l(t, \mathbf x) = v_t^l\left(\Phi(t,\mathbf x) \right), \quad \Phi^l(0,\mathbf x) = x^l \sim \mathcal{U}(\mathbb S^{d-1}), \quad l = 1,\ldots,L,
\end{equation}
transports $p_0 \sim \mathcal U(\mathbb S^{(d-1)L})$ to $p_1 = p_{\text{data}}$. The equations are coupled through the posterior $p_t^l(w_k|\mathbf x)$, which depends on the full state $\mathbf x = (x^1,\ldots,x^L)$ via the backbone. Since $\kappa_t$ is strictly monotone, we discretize the ODE with Euler steps of size $\Delta\kappa_n = \kappa_{t_{n+1}} - \kappa_{t_n}$ in concentration space, using the factorization $\psi_t = \dot\kappa_t\,\tilde\psi_t$ from \eqref{eq:psi_tilde} so that $\dot\kappa_t$ cancels and never appears explicitly. Numerical integration introduces discretization errors that accumulate over steps.

\paragraph{Predictor-corrector (PC).} The score provides a mechanism to do corrector steps during inference. After advancing the state by one ODE step, we apply $k$ Langevin diffusion steps at the current concentration $\kappa_t$, which leave $p_t$ invariant. The Langevin dynamics on $(\mathbb S^{d-1})^L$ read per position as
\begin{align}\label{eq:langevin}
\mathrm d X_\tau^l &= \nabla_{\mathbb S^{d-1},x^l} \log p_t(\mathbf X_\tau)\,\mathrm d \tau + \sqrt{2}\,\mathrm d B_\tau^l 
= \kappa_t \sum_{k=1}^N p_t^l(w_k|\mathbf X_\tau)\,{\rm P}_{x^l}(w_k)\,\mathrm d\tau + \sqrt{2}\,\mathrm d B_\tau^l,
\tag{PC}
\end{align}
where $B_\tau^l$ is Brownian motion on $\mathbb S^{d-1}$ at position $l$, independent across positions. The discretization is given in Algorithm~\ref{alg:ode}.

\paragraph{Self-conditioning (SC).} Following \cite{chen2023analogbitsgeneratingdiscrete},
we optionally let the network see its own prediction. Each position's
predicted posterior is summarized by its mean embedding
$\sum_{k} p_t^l(w_k \mid \mathbf x)\, w_k$, and a linear layer turns this estimate into
an additive correction of the network input. During training the estimate comes from a
preview forward pass without gradients on a fraction $p = 0.25$ of the steps and is
zero otherwise. During sampling we reuse the posterior from the previous integration
step.

\paragraph{Decoding.}
Recall from Section~\ref{sec:vM} that the schedule $\kappa_t$ is a continuously differentiable, strictly increasing bijection onto $[0, \infty)$ with $\kappa_0 = 0$, so that $p_1 = p_{\text{data}}$ holds in the limit $t \to 1$. In practice we integrate up to a finite terminal concentration $\kappa_{\max} > 0$ and decode from the posterior at the terminal state via \eqref{eq:decoding}. 
Sampling consists of solving the flow ODE \eqref{eq:flow_ode} from $t=0$ to $t=1$,
optionally interleaved with Langevin corrections, transporting the uniform distribution on $(\mathbb S^{d-1})^L$ to a terminal density on $(\mathbb S^{d-1})^L$. At the terminal state $\mathbf x_1$, we pass through the backbone to obtain $\hat{\mathbf x} = T_\theta(\mathbf x_1)$ and decode the token at each position $l \in \{1,\ldots,L\}$ as
\begin{equation}\label{eq:decoding}
\hat y^l = v_{k^*}, \quad k^* \coloneqq \argmax_{k=1,\ldots,N}\, s_k^l, \quad s_k^l = \langle w_k, \hat x^l \rangle + b_k.
\end{equation}

\section{Experiments}\label{sec:exp}
%----------------------------------------------
We evaluate the vMF path on Sudoku, the One Billion Word Benchmark \cite{lm1b}, TinyGSM \cite{liu2023tinygsm}, and OpenWebText \cite{Gokaslan2019OpenWeb}.
We compare four continuous conditional paths, two spherical and two Euclidean ones: 
\begin{enumerate}
    \item 
    vMF path (Section~\ref{sec:vM}), 
    \item 
     geodesic (slerp) interpolation (Remark~\ref{rem:geo_path}),
    \item  variance-preserving (VP), i.e.\ the linear interpolation from Example~\ref{bsp2},
given as $h_t = (1 - t) Z + t w_k$ with $Z \sim \mathcal N(0, I)$ and
\item variance-exploding (VE), given as $h_t = w_k + \sigma_t Z$ with $\sigma_t$ growing from $0$ to a terminal $\sigma_{\max}$.
\end{enumerate} 
For more details see Appendix \ref{sec:paths_app}.

\paragraph{Metrics.} There is a trade-off between generation perplexity and entropy. We therefore report the entropy $H$ next to every perplexity and compare at matched entropy. Some reference values are published without an entropy. Sudoku validity and GSM8K report accuracy.

\paragraph{Sampling ablation.} At inference, we run two samplers per method at fixed $\mathrm{NFE} = 128$ network evaluations ($\mathrm{NFE} = 1024$ for TinyGSM and OpenWebText), except where a table states another budget. The first is an Euler integrator of the flow ODE \eqref{eq:flow_ode} with $n$ predictor steps. The second is the predictor-corrector scheme \eqref{eq:langevin}, in which each predictor step is followed by $k$ Langevin corrector steps, so that $\mathrm{NFE} = n(1 + k)$. We sweep the predictor-corrector split, predictor spacing (uniform or warp-aware, Appendix~\ref{sec:schedule_app}), and corrector step size at fixed $\mathrm{NFE}$. Tables~\ref{tab:sudoku_main}, \ref{tab:lm1b-comparison}, \ref{tab:gsm8k_main}, and~\ref{tab:owt_main}, and Figures~\ref{fig:ppl_vs_entropy} and~\ref{fig:owt_ppl_vs_entropy}, report the best result per entry over the sweep. Full grids, configurations, and the sweep description are in Appendix~\ref{sec:inference_ablation}. Inference-seed variability for the headline entries is reported in Appendix~\ref{sec:seed_bars}. In Appendix~\ref{sec:sde_results} we additionally evaluate an SDE sampler (Algorithm~\ref{alg:sde_app}) on the same checkpoints across all tasks, which tracks the predictor-corrector results at matched budgets. 

\paragraph{Networks.} All continuous methods share a DiT-style transformer \cite{peebles2023scalable} backbone with adaLN conditioning; masked diffusion uses the same backbone with a discrete embedding layer in place of the tied embeddings. Per-task hyperparameters and the training recipe are in Appendix~\ref{sec:arch_app}.

\subsection{Sudoku}\label{sec:exp_sudoku}
Sudoku is a standard puzzle played on a $9\times 9$ grid, partitioned into nine $3\times 3$ subgrids. 
Some cells are pre-filled as \emph{clues}, and the goal is to fill the remaining cells with digits $1$--$9$ 
such that each row, each column, and each $3\times 3$ subgrid contains every digit exactly once.

\paragraph{Setup.} We use the Sudoku-Extreme dataset from \cite{wang2025hierarchicalreasoningmodel}, a collection of
3,104,157 challenging puzzles that typically require substantially
more search/backtracking (i.e., “guesses”) for classical solvers. Generation is conditional: the model receives a partial puzzle and fills the missing cells, with clue positions pinned at every sampling step. We measure validity, the fraction of generated solutions satisfying all constraints, on $1{,}000$ held-out partial puzzles at $\mathrm{NFE} = 128$. As a discrete baseline, we use masked diffusion model  (MDM)~\cite{sahoo2024simple}, where $p$ is the power in the mask schedule. The probability that a position is masked at corruption time $t$ is $1 - (1-t)^p$. We report $p = 1$ and $p = 0.5$. Because the vocabulary is small ($V=10$), we use $\mathbb S^{10}$ for the sphere paths and $\mathbb R^{10}$ for the Euclidean paths so that the intrinsic dimension matches across paths.

\begin{wraptable}[14]{r}{0.35\textwidth} % use 0.48\columnwidth in 2-col papers
\vspace{-1.5em}
\centering
\small
\setlength{\tabcolsep}{5pt}
\begin{tabular}{lcc}
\toprule
Method & ODE (\%) & PC (\%) \\
\midrule
\multicolumn{3}{l}{\textit{Reference}} \\
MDM$_{p = 1}$   & $22.4$ & $38.8$ \\
MDM$_{p = 0.5}$ & $35.1$ & $56.7$ \\

\midrule
\multicolumn{3}{l}{\textit{Ours}} \\
vMF      & $\mathbf{65.8}$ & $\mathbf{74.5}$ \\
Geodesic & $49.4$ & -- \\
VP       & $53.4$ & $69.1$ \\
VE       & $48.0$ & $46.5$ \\
\bottomrule
\end{tabular}
\vspace{2pt}
\caption{\textbf{Sudoku Extreme} \cite{wang2025hierarchicalreasoningmodel} validity ($\uparrow$) under ODE and predictor-corrector sampling.}
\label{tab:sudoku_main}
\vspace{-0.8em}
\end{wraptable}
\paragraph{Results.} Using the PC sampler, see Table~\ref{tab:sudoku_main}, improves accuracy substantially for vMF and VP. For VE there is no useful PC improvement. We note that VP requires time dependent scaling of the epsilon used for PC steps (Appendix~\ref{sec:sweep_choices}). PC cells report the best configuration over the full sweep grid (Appendix~\ref{sec:sudoku_grids}). We also report results using non-time-conditioned networks in Table~\ref{tab:sudoku_nontc}. Masked trails the continuous methods at both schedule powers. Two inference-time samplers close some of that gap, neither needing retraining. Remasking \cite{wang2025remasking} lets the model revise tokens it has already committed , it fills the masked PC column. Committing the most confident positions first, using the sampler from \cite{bie2025llada20scalingdiffusionlanguage}, reaches $75.5$ at $p = 0.5$, using an adaptive budget. Grids for both are in Appendix~\ref{sec:sudoku_grids}.

\subsection{Language Modeling (LM1B)}\label{sec:exp_lm1b}

The One Billion Word Benchmark \cite{lm1b} is a dataset based on crawled news data.  

\paragraph{Setup.} We use the BERT-base-uncased tokenizer \cite{devlin2019bertpretrainingdeepbidirectional} ($N = 30{,}522$, $L = 128$). Training and evaluation follow  MDLM~\cite{sahoo2024simple} and LangFlow \cite{chen2026langflowcontinuousdiffusionrivals} with a DiT-style transformer backbone, $1$M training steps, and GPT-2-large \cite{radford2019language} scoring over $1024$ unconditional samples. 
We report generation perplexity (gen.\ PPL) and the average per-sample entropy $H$. Geodesic admits no closed-form Riemannian score and uses ODE only.The sweep choices are described in Appendix~\ref{sec:sweep_choices} and the selected configurations in Appendix~\ref{sec:lm1b_grids}

\begin{wraptable}[24]{r}{0.42\textwidth}
\vspace{-0.5em}
\centering
\small
\setlength{\tabcolsep}{5pt}
\begin{tabular}{lc}
\toprule
Method & gen.\ PPL ($\downarrow$) \eH{$H$ ($\uparrow$)}\\
\midrule
\multicolumn{2}{l}{\textit{Reference}} \\
AR \cite{chen2026langflowcontinuousdiffusionrivals}           & $\mathbf{66.7}$ \\
Plaid \cite{gulrajani2023likelihood}        & $77.3$ \\
LangFlow \cite{chen2026langflowcontinuousdiffusionrivals}     & $92.2$\eH{4.31} \\
MDLM \cite{sahoo2024simple}         & $103.9$ \\
SEDD \cite{lou2023discrete}         & $115.9$ \\
Duo \cite{sahoo2025diffusionduality}          & $97.6$ \\
DFM Diagonal \cite{potaptchik2026discreteflowmaps} & $75.82$\eH{4.19} \\
FLM \cite{lee2026flowmaplanguagemodels}          & $96.9$\eH{4.29} \\
\midrule
\multicolumn{2}{l}{\textit{Ours (Euler sampling)}} \\
vMF      & $171.8$\eH{4.35} \\
Geodesic & $167.9$\eH{4.34} \\
VP       & $140.4$\eH{4.34} \\
VE       & $149.7$\eH{4.35} \\
\midrule
\multicolumn{2}{l}{\textit{+ self-conditioning ($p = 0.25$)}} \\
vMF      & $92.4$\eH{4.34} \\
Geodesic & $107.1$\eH{4.33} \\
VP       & $92.3$\eH{4.33} \\
VE       & $163.4$\eH{4.36} \\
\bottomrule
\end{tabular}
\vspace{2pt}
\caption{\textbf{LM1B} at $\mathrm{NFE} = 128$ under the ODE sampler; gen.\ PPL ($\downarrow$) with entropy in light gray. Predictor-corrector results are in Table~\ref{tab:lm1b_pc}. Best per column in bold.}
\label{tab:lm1b-comparison}
\vspace{-0.8em}
\end{wraptable}
\paragraph{Baselines.} As baselines we use
autoregressive transformer (AR) \cite{chen2026langflowcontinuousdiffusionrivals} which is sequential, LangFlow \cite{chen2026langflowcontinuousdiffusionrivals}, Plaid \cite{gulrajani2023likelihood}, MDLM \cite{sahoo2024simple}, Duo \cite{sahoo2025diffusionduality}
 and SEDD from \cite{lou2023discrete}.  As an additional reference point, we report the diagonal variant (configuration without flow-map distillation)  of \cite{potaptchik2026discreteflowmaps} at the same NFE. Note that LangFlow uses self-conditioning. Without  self-conditioning they report a gen.\ PPL of $154.2$.

\paragraph{Results.}
Table~\ref{tab:lm1b-comparison} compares each method against the reference baselines
under ODE sampling. Table~\ref{tab:lm1b_pc} and Figure~\ref{fig:ppl_vs_entropy}
report PC results across entropy floors.

\begin{table}[t]
\centering
\small
\setlength{\tabcolsep}{5pt}
\begin{tabular}{lccccc}
\toprule
 & \multicolumn{5}{c}{gen.\ PPL ($\downarrow$) \eH{$H$ ($\uparrow$)}} \\
\cmidrule(lr){2-6}
Method & PC$_{H \geq 4.34}$ & PC$_{H \geq 4.30}$ & PC$_{H \geq 4.25}$ & PC$_{H \geq 4.20}$ & PC$_{H \geq 4.15}$ \\
\midrule
vMF & $165.4$\eH{4.36} & $112.7$\eH{4.30} & $66.0$\eH{4.25} & $52.4$\eH{4.21} & $45.1$\eH{4.16} \\
VP & $138.4$\eH{4.35} & $138.4$\eH{4.35} & $138.4$\eH{4.35} & $138.4$\eH{4.35} & $138.4$\eH{4.35} \\
VE & $151.7$\eH{4.35} & $151.7$\eH{4.35} & $151.7$\eH{4.35} & $151.7$\eH{4.35} & $151.7$\eH{4.35} \\
\midrule
\multicolumn{6}{l}{\textit{+ self-conditioning ($p = 0.25$)}} \\
vMF & $\mathbf{77.5}$\eH{4.34} & $\mathbf{55.6}$\eH{4.30} & $\mathbf{38.4}$\eH{4.25} & $\mathbf{33.8}$\eH{4.21} & $\mathbf{27.0}$\eH{4.17} \\
VP & $88.9$\eH{4.34} & $83.1$\eH{4.33} & $83.1$\eH{4.33} & $83.1$\eH{4.33} & $83.1$\eH{4.33} \\
VE & $164.6$\eH{4.36} & $164.6$\eH{4.36} & $164.6$\eH{4.36} & $164.6$\eH{4.36} & $164.6$\eH{4.36} \\
\bottomrule
\end{tabular}
\vspace{2pt}
\caption{\textbf{LM1B} predictor-corrector (PC) at $\mathrm{NFE} = 128$: best gen.\ PPL per entropy floor, the leading column held to $H_{\mathrm{data}} = 4.34$. Best per column in bold. Configurations in Table~\ref{tab:lm1b-configs}.}
\label{tab:lm1b_pc}
\end{table}

Under the ODE, vMF does not lead. VP attains the lowest gen.\ PPL. The methods
  separate once the corrector is switched on, where vMF smoothly descends, trading
  perplexity with entropy (Figure~\ref{fig:ppl_vs_entropy}, Table~\ref{tab:lm1b_pc}). At
$H = 4.21$ vMF reaches $52.4$, improving on DFM Diagonal on both axes ($75.82$ at
$H = 4.19$). We do not have a complete account of why the gap is this large. One
possibility is that vMF profits from the noisy state and the target embeddings
having the same norm.

Self-conditioning ($p = 0.25$) helps every method except VE and sharpens the same
picture. Under the ODE vMF and VP are level ($92.4$ and $92.3$), both matching
LangFlow ($92.2$), which also uses self-conditioning. Under PC vMF is best at every
floor, $77.5$ at $H = 4.34$ and $33.8$ at $H = 4.21$, against VP's $88.9$ and $83.1$.

Non-time-conditioned results (Table~\ref{tab:lm1b_nontc}) and an NFE ablation (Table~\ref{tab:lm1b_nfe}) are in Appendix~\ref{sec:lm1b_grids}.

\subsection{Mathematical Reasoning (TinyGSM)}\label{sec:exp_tinygsm}
\begin{wrapfigure}[10]{r}{0.35\textwidth}
\vspace{-3.5em}
\centering
\includegraphics[width=\linewidth]{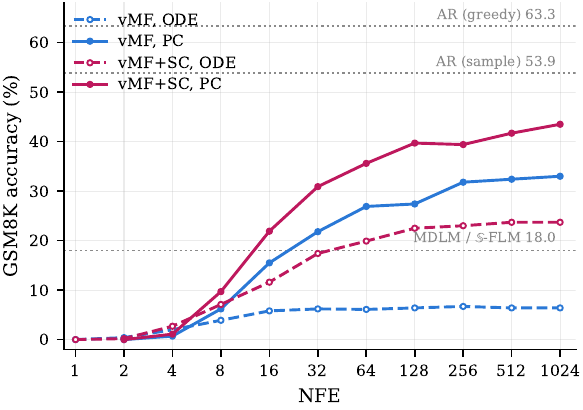}
\caption{\textbf{TinyGSM} accuracy against sampling budget. Dotted: references at $1024$ steps.}
\label{fig:gsm8k_nfe}
\end{wrapfigure}
TinyGSM \cite{liu2023tinygsm} is a synthetic dataset of roughly $11.8$M GSM8K-style \cite{cobbe2021gsm8k} grade-school math problems paired with Python solutions. Following \cite{deschenaux2026hyperspherical}, we use it to test whether the models transfer from language modeling to mathematical reasoning.

\paragraph{Setup.} Training data, tokenizer, and the execution-based evaluation protocol follow \cite{deschenaux2026hyperspherical}: we use the SmolLM-135M tokenizer ($N = 49{,}152$, $L = 512$) and train for $250$k steps with the models of Table~\ref{tab:arch_hparams}. 
We evaluate on the full GSM8K test set ($1{,}319$ problems): for each problem the model generates one Python solution, which is executed in a sandbox and we report the fraction of problems whose executed solution yields the correct answer. 

\paragraph{Results.}
\begin{wraptable}[19]{r}{0.42\textwidth}
\centering
\small
\setlength{\tabcolsep}{5pt}
\begin{tabular}{lcc}
\toprule
Method & \multicolumn{2}{c}{Acc.\ ($\uparrow$, \%)} \\
\midrule
\multicolumn{3}{l}{\textit{Reference} \cite{deschenaux2026hyperspherical}} \\
AR (sample)  & \multicolumn{2}{c}{$53.9$} \\
AR (greedy)  & \multicolumn{2}{c}{$63.3$} \\
MDLM         & \multicolumn{2}{c}{$18.0$} \\
DUO          & \multicolumn{2}{c}{$17.2$} \\
FLM          & \multicolumn{2}{c}{$0.3$}  \\
$\mathbb{S}$-FLM (top-$k{=}1$) & \multicolumn{2}{c}{$18.0$} \\
\midrule
\textit{Ours} & ODE & PC \\
\midrule
vMF      & $6.4$  & $33.0$ \\
Geodesic & $5.0$  & -- \\
VP       & $7.7$  & $12.7$ \\
VE       & $2.1$  & $1.9$ \\
\midrule
\multicolumn{3}{l}{\textit{+ self-conditioning ($p = 0.25$)}} \\
vMF      & $\mathbf{23.7}$ & $\mathbf{43.5}$ \\
Geodesic & $14.9$          & -- \\
VP       & $18.3$          & $28.5$ \\
VE       & $3.5$           & $4.1$ \\
\bottomrule
\end{tabular}
\caption{\textbf{GSM8K} test accuracy ($\uparrow$, \%) after training on TinyGSM, at
$\mathrm{NFE} = 1024$; ODE/PC denotes the sampler. Reference values from
\cite{deschenaux2026hyperspherical}. Best per column in bold.}
\label{tab:gsm8k_main}
\end{wraptable}
Table~\ref{tab:gsm8k_main} reports GSM8K accuracy at $\mathrm{NFE} = 1024$ against the
reference values of \cite{deschenaux2026hyperspherical}. The autoregressive baselines
decode $L = 512$ tokens in $512$ sequential passes, half our budget. 

Without self-conditioning, ODE sampling leaves every path in the single digits, and the
corrector is what separates them. vMF goes from $6.4\%$ to $33.0\%$, VP only from
$7.7\%$ to $12.7\%$.

Self-conditioning adds another ten points under PC and leaves vMF best in both columns,
$23.7\%$ under the ODE and $43.5\%$ under PC. That is above MDLM and $\mathbb{S}$-FLM
(both $18.0\%$) and still below the autoregressive baseline.

We further ablate the sampling budget in Appendix~\ref{sec:gsm8k_nfe_app} (Figure~\ref{fig:gsm8k_nfe}, Table~\ref{tab:gsm8k_nfe}). Without self-conditioning the ODE sampler saturates near
$\mathrm{NFE} = 32$, and past that point only corrector steps turn further network
evaluations into accuracy.

\subsection{Results on OpenWebText}\label{sec:exp_owt}
\begin{figure}[t]
    \centering
    \begin{subfigure}[t]{0.49\textwidth}
        \centering
        \includegraphics[width=\textwidth]{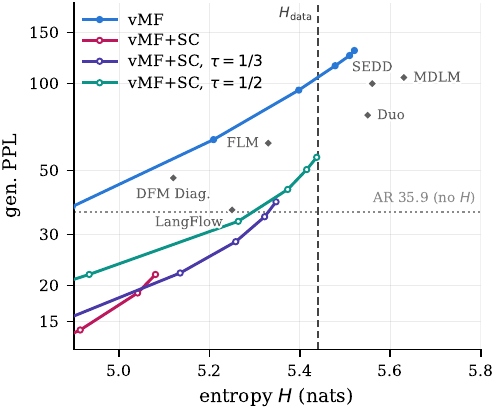}
        \caption{vMF}\label{fig:owt_ppl_vs_entropy_vmf}
    \end{subfigure}
    \hfill
    \begin{subfigure}[t]{0.49\textwidth}
        \centering
        \includegraphics[width=\textwidth]{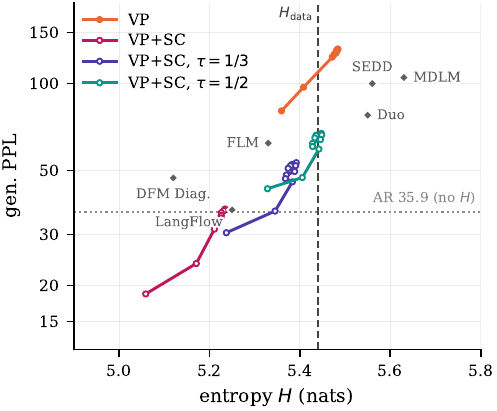}
        \caption{VP}\label{fig:owt_ppl_vs_entropy_vp}
    \end{subfigure}
    \caption{\textbf{OpenWebText} gen.\ PPL against entropy at $\mathrm{NFE} = 1024$, sweeping $\varepsilon$ at self-conditioning onsets $\tau \in \{0, 1/3, 1/2\}$; every curve is predictor-corrector. References that report an entropy are placed at it; AR reports none and is drawn as a line. Dashed: $H_{\mathrm{data}}$. Both panels share axes. Configurations in Appendix~\ref{sec:owt_nfe}.}
    \label{fig:owt_ppl_vs_entropy}
\end{figure}

We report results on OpenWebText \cite{Gokaslan2019OpenWeb}, a web-crawled corpus commonly used to train GPT-2-scale models.

\paragraph{Setup.} We use the GPT-2 tokenizer ($N = 50{,}257$, $L = 1024$) and train
for $1$M steps with the models of Table~\ref{tab:arch_hparams}. Evaluation follows the
LM1B protocol. We drop VE, which trails the other paths under PC on every earlier task.
We report $\mathrm{NFE} = 128$ and $1024$, under the ODE sampler and under the best PC configuration per
entropy floor (Table~\ref{tab:owt_pc}, Figure~\ref{fig:owt_ppl_vs_entropy}). Each ODE
cell gives the better of the uniform and warp-aware variants. The delayed-onset cells
additionally sweep the onset $\tau$, the time from which the model receives its own
previous prediction rather than a zero input, and report the best value near the data
entropy. $\tau = 0$ is the standard variant. Per-variant values are in
Table~\ref{tab:owt_ode_variants}.

\paragraph{Results.}
\begin{wraptable}[25]{r}{0.50\textwidth}
\vspace{-1.5em}

\centering
\footnotesize
\setlength{\tabcolsep}{5pt}

\begin{tabular}{lcc}

\toprule
 & \multicolumn{2}{c}{gen.\ PPL ($\downarrow$) \eH{$H$ ($\uparrow$)}} \\
\cmidrule(lr){2-3}
Method & $\mathrm{NFE}\,{=}\,128$ & $\mathrm{NFE}\,{=}\,1024$ \\
\midrule
\multicolumn{3}{l}{\textit{Reference ($1024$ sampling steps)}} \\
AR \cite{chen2026langflowcontinuousdiffusionrivals}
    & -- & $35.9$ \\
LangFlow \cite{chen2026langflowcontinuousdiffusionrivals}
    & -- & $36.5$\eH{5.25} \\
FLM \cite{lee2026flowmaplanguagemodels}
    & -- & $62.2$\eH{5.33} \\
DFM Diagonal \cite{potaptchik2026discreteflowmaps}
    & -- & $47.1$\eH{5.12} \\
Duo \cite{sahoo2025diffusionduality}
    & -- & $77.7$\eH{5.55} \\
SEDD Uniform \cite{lou2023discrete}
    & -- & $99.9$\eH{5.56} \\
MDLM \cite{sahoo2024simple}
    & -- & $104.9$\eH{5.63} \\
\midrule
\multicolumn{3}{l}{\textit{Ours}} \\
vMF      & $146.9$\eH{5.54} & $139.3$\eH{5.52} \\
Geodesic & $144.4$\eH{5.54} & $139.0$\eH{5.51} \\
VP       & $128.6$\eH{5.48} & $123.0$\eH{5.46} \\
\midrule
\multicolumn{3}{l}{\textit{+ self-conditioning ($p = 0.25$)}} \\
vMF      & $53.6$\eH{5.35} & $\mathbf{36.0}$\eH{5.20} \\
Geodesic & $55.0$\eH{5.38} & $36.3$\eH{5.23} \\
VP       & $\mathbf{50.1}$\eH{5.35} & $36.3$\eH{5.22} \\
\midrule
\multicolumn{3}{l}{\textit{+ self-conditioning, delayed onset}} \\
vMF      & $70.7$\eH{5.45} & $73.9$\eH{5.46} \\
Geodesic & $67.5$\eH{5.45} & $70.3$\eH{5.45} \\
VP       & $76.1$\eH{5.48} & $66.2$\eH{5.43} \\
\bottomrule
\end{tabular}

\caption{\textbf{OpenWebText} under the ODE sampler
($n = \mathrm{NFE}$). Best per budget in bold, taken across all rows;
references are not eligible. Delayed-onset rows select over
$\tau \in \{1/3, 1/2\}$, per-onset values in
Table~\ref{tab:owt_ode_variants}.}
\label{tab:owt_main}
\end{wraptable}
Under plain ODE sampling (Table~\ref{tab:owt_main}), VP leads as on LM1B ($123.0$
against $139.3$ for vMF at $\mathrm{NFE} = 1024$). Self-conditioning changes this, and
the gain grows with the NFE budget. vMF goes from $53.6$ to $36.0$ over the same range.
At $L = 1024$ this carries a cost. Self-conditioned ODE sampling sheds entropy as the
budget grows, down to $H = 5.20$, below $H_{\mathrm{data}} = 5.44$, before the
corrector runs. It is not specific to our paths. VP and Geodesic show it, and LangFlow,
which also uses self-conditioning, reports $H = 5.25$. Delaying the onset returns entropy,
with $\tau \in \{1/3, 1/2\}$ reaching $H = 5.45$.

Against the references, vMF at $36.0$ is close to LangFlow ($36.5$) at comparable
entropy and ahead of DFM Diagonal ($47.1$ at $5.12$). At the data entropy the delayed
onset gives $70.7$, ahead of Duo ($77.7$), SEDD Uniform ($99.9$) and MDLM ($104.9$),
which all sample above it. AR gives $35.9$ without a reported entropy.

PC (also using delayed onset) improves this throughout, and vMF leads at the data entropy
(Figure~\ref{fig:owt_ppl_vs_entropy}, Tables~\ref{tab:owt_pc}
and~\ref{tab:owt_pc_128}). It reaches $52.7$ at $\mathrm{NFE} = 128$ and $55.5$ at
$1024$, slightly ahead of VP at both budgets and with and without self-conditioning,
and ahead of every reference that samples at least as much entropy. At lower floors it
passes LangFlow and DFM Diagonal as well, though VP overtakes vMF there. Qualitative
samples are in Appendix~\ref{sec:qual_samples}.

\section*{Limitations and Broader Impact}
Our work shares the limitations of continuous diffusion approaches to discrete generation. The main technical specificity is that the conditional velocity and the radial CDF must be computed numerically, which requires care at high concentration and high embedding dimension to avoid underflow in the radial density. We address this with a flux-based ODE formulation and log-shifted density evaluation (Appendix~\ref{sec:paths_app}). We are not aware of any negative societal impact specific to this work beyond those that apply to generative modeling of discrete sequences in general. All experiments use public datasets.

\section{Conclusion}\label{sec_conclusion}

We introduced a vMF path on $\mathbb S^{d-1}$ for continuous generative modeling of discrete sequences. More generally, we showed that for radially symmetric paths on the sphere the continuity equation reduces to a one dimensional flux equation in the cosine similarity. For the vMF path, this reduction yields a linear ODE whose unique bounded solution gives a tractable conditional velocity. The Riemannian score follows directly from the vMF density, making ODE and predictor-corrector sampling available solely from the posterior.

We applied this construction by embedding tokens as learned points on $\mathbb S^{d-1}$ and training a posterior with a cross-entropy loss. This posterior determines the marginal velocity and score as well as the terminal decoding probabilities. On Sudoku, LM1B, TinyGSM, and OpenWebText, we compared vMF paths to geodesic and Euclidean baselines and showed the predictor-corrector scheme, especially for vMF, produces strong results. We believe the results show much promise and open several directions for future work.

\paragraph{Acknowledgments.} GS acknowledges funding by
the German Research Foundation (DFG) within the Excellence Cluster MATH+ and JC by project \mbox{STE 571/17-2} within the DFG-SPP 2298. GK acknowledges funding by the BMBF project VIScreenPRO (ID:
100715327).

\bibliographystyle{bib_style}
\bibliography{bibliography}

\appendix
\tableofcontents
\setcounter{topnumber}{4}
\setcounter{bottomnumber}{3}
\setcounter{totalnumber}{6}
\renewcommand{\topfraction}{0.92}
\renewcommand{\bottomfraction}{0.85}
\renewcommand{\textfraction}{0.05}
\renewcommand{\bottomfraction}{0.9}
\renewcommand{\floatpagefraction}{0.5}

%------------------------------------------------------
\section{Related Work}\label{sec:related}
%------------------------------------------------------

Adapting diffusion and flow-based generative models to discrete data is an active area of research. Existing approaches either work directly in the discrete state space or embed the data into a continuous space.

Our approach is closest in spirit to CDCD \cite{dieleman2022continuous}. We use normalized embeddings, cross-entropy training, and an adaptive schedule. The key difference is that we keep the noisy state on $\mathbb S^{d-1}$ throughout the generative process. In the following we highlight further related work.

\paragraph{Discrete state space.}
Discrete state space methods define Markov processes over the finite vocabulary. In general these models split into masked diffusion \cite{sahoo2024simple, campbell2024generative, shi2025simplifiedgeneralizedmaskeddiffusion, nie2025largelanguagediffusionmodels} and uniform diffusion \cite{austin2021structured,schiff2025simpleguidancemechanismsdiscrete, sahoo2025diffusionduality}, which converge to a special mask token or uniform random tokens respectively. D3PM \cite{austin2021structured} designed discrete time forward processes. This was extended to continuous time based on the CTMC by \cite{campbell2024generative}. SEDD \cite{lou2023discrete} build a score entropy loss. \cite{sahoo2024simple, ou2026absorbingdiscretediffusionsecretly, shi2025simplifiedgeneralizedmaskeddiffusion} introduce theoretical results and LLaDA \cite{nie2025largelanguagediffusionmodels} scaled mask diffusion to 8 billion parameters. Recent work, GDDS \cite{zekri2026generalized} uses GPT-2 embeddings to induce semantic structure in discrete noising. Continuous noising in the learned embedding space follows a related intuition.

\paragraph{Continuous state space.}
Continuous embedding methods assign each token a vector and apply noise in the embedding space. Diffusion-LM \cite{li2022diffusion} uses Gaussian DDPM with learned embeddings. \cite{gulrajani2023likelihood} uses a discrete time process without normalized embeddings. \cite{lee2026flowmaplanguagemodels,roos2026categoricalflowmaps} use a linear interpolant to the one-hot target in the large $\mathbb R^{V}$ space with a softmax denoiser and flow distillation for few-step language generation. LangFlow \cite{chen2026langflowcontinuousdiffusionrivals} works in the continuous embedding space and introduces an ODE based NLL. 
In all of these the noisy state lives in $\mathbb R^d$ or $\mathbb R^{V}$.

A separate line of work defines noise processes on the probability simplex or on spheres of dimension $|\mathcal V| - 1$. Generative Assignment Flows \cite{boll2024generativemodelingdiscretejoint} construct flows on the probability simplex by defining linear paths in logit space and mapping them to simplex-valued distributions via softmax. Dirichlet Flow Matching \cite{stark2024dirichletflowmatchingapplications} constructs paths from Dirichlet distributions on the simplex. RDLM \cite{jo2025continuousdiffusionmodellanguage} uses the diffeomorphism between the simplex and the positive orthant of the sphere $\mathbb{S}^{|\mathcal{V}|-1}$ for language modeling and derives an NLL bound via Girsanov's theorem. To simulate the bridge SDE, they split the sphere into a product of smaller spheres, since the bridge on high dimensional spheres is difficult. \cite{davis2024fisherflowmatchinggenerative, cheng2025categoricalflowmatchingstatistical} map the simplex to the positive orthant of $\mathbb S^{|\mathcal V|-1}$ via the standard diffeomorphism and construct geodesic flows there. Concurrent to our work, \cite{deschenaux2026hyperspherical} also propose flows on the sphere for language modeling; they focus on the geodesic path and a custom spherical network architecture.

\paragraph{Mixing discrete and continuous methods.} A different line of work is the combination of discrete and continuous diffusion models for categorical data. \cite{pynadath2025candihybriddiscretecontinuousdiffusion} propose CANDI, a hybrid discrete--continuous diffusion model with a structured forward noising process that combines a discrete masking process with Gaussian corruption. During reverse sampling, the model starts from a fully corrupted Gaussian state: positions selected to become clean are sampled from the model's token distribution and then carried forward/fixed, while the remaining noisy positions continue to be updated by the continuous reverse ODE. Duo by \cite{sahoo2025diffusionduality} shows that uniform-state discrete diffusion arises as the $\operatorname{argmax}$ push-forward of an underlying Gaussian diffusion process. They exploit this duality to improve training and enable much faster few-step sampling via discrete consistency distillation.

\paragraph{Sudoku.} For non-generative neural network approaches on Sudoku-Extreme, see \cite{ wang2025hierarchicalreasoningmodel, KCWS2026, jolicoeurmartineau2025morerecursivereasoningtiny}.

%------------------------------------------------------
\section{Further Theory and Proofs}\label{app:theory}
%------------------------------------------------------
\subsection{Proofs}\label{appa}
\productCE*
\paragraph{Proof.}
Using the product rule we obtain
\begin{equation}
    \partial_t p_t(\mathbf x|\mathbf w)
= \sum_{l=1}^L \Big( \prod_{j \neq l}^L p_t^j(x^j|w^j) \Big)\, \partial_t p_t^l(x^l|w^l)
\end{equation}
and further by the continuity equation for $p_t^l$,
\begin{equation}
   \partial_t p_t(\mathbf x|\mathbf w)
= - \sum_{l=1}^L \Big( \prod_{j \neq l}^L p_t^j(x^j|w^j) \Big)
\operatorname{div}_{\mathcal M}^L\big( p_t^l(x^l|w^l)\, v_t^l(x^l|w^l) \big).
\end{equation}
Since the factor $\prod_{j \neq l} p_t^j(x^j|w^j)$ does not depend on $x^l$, we can pull it into the divergence operator, which acts only on the $l$-th variable which results in
\begin{equation}
   \partial_t p_t(\mathbf x|\mathbf w)
= - \sum_{l=1}^L \operatorname{div}_{\mathcal M}^l \Big(
\Big( \prod_{j=1}^L p_t^j(x^j|w^j) \Big) v_t^l(x^l|w^l)
\Big).
\end{equation}
Recognizing that $p_t(\mathbf x|\mathbf w) = \prod_{j=1}^L p_t^j(x^j|w^j)$, this yields
\begin{equation}
   \partial_t p_t(\mathbf x|\mathbf w)
= - \sum_{l=1}^L \operatorname{div}_{\mathcal M} \big( p_t(\mathbf x|\mathbf w)\, v_t^l(x^l|w^l) \big).
\end{equation}
Finally, using that 
the divergence on the product manifold $\mathcal M^L$ is given as
$\operatorname{div}_{\mathcal M^L} (u)\coloneqq \sum_{l=1}^L \operatorname{div}_{\mathcal M}(u^l)$, we conclude
\[
\partial_t p_t(\mathbf x|\mathbf w)
= 
- \operatorname{div}_{\mathcal M^L} 
\big( p_t(\mathbf x|\mathbf w)\, v_t(\mathbf x|\mathbf w) \big).
%\qquad \Box
\]
\vspace{0.5cm}

\radialcontinuity*
\paragraph{Proof.}Let $s \coloneqq \langle w,x \rangle$. Using the divergence on the sphere
we want to reformulate \eqref{eq:cont_sphere}.
With 
$$u_t (x) \coloneqq p_t(x) v_t (x) = g_t (s) (w-sx), \quad g_t \coloneqq \bar p_t \psi_t,$$
we obtain 
$$
\nabla u_t (x)  = g_t'(s)  w (w-sx )^\top 
- g_t(s)\big( x {\rm I}_d + x w^\top \big),
$$
and further
\begin{align}
\text{tr} \left(\nabla u_t (x) \right) &= g_t'(s) (1-s^2) - g_t(s) (d+1) s,\\
x^\top \, \nabla u_t(x) \,  x &= - 2s g_t(s).
\end{align}
Thus, the continuity equation \eqref{eq:cont_sphere} can be rewritten as
\begin{align}
0 &= \partial_t p_t + \divg_{\mathbb S^{d-1}}(p_t\, v_t) =
\nabla_t \bar p_t + \text{tr} (\nabla u) - x^\top  \nabla u_t \, x\\
&=
\partial_t \bar p_t + g_t' (1-s^2) - g_t (d-1) s. \label{to_get}
\end{align}
On the other hand, the flux equation in \eqref{eq:1d_flux} becomes
\begin{align}
0&=\partial_t f_t + \partial_s \big(g_t  \, (1-s^2)^{(d-1)/2} \big)\\
&= 
\Big( (1-s^2)^{(d-3)/2} \partial_t \bar p_t 
+ g_t' (1-s^2)^{(d-1)/2} + g_t (d-1) s (1-s^2)^{(d-3)/2} \Big).
\end{align}
Dividing by $(1-s^2)^{(d-3)/2}$, $s \in (-1,1)$ yields equation 
\eqref{to_get} and we are done. \hfill $\Box$
\vspace{0.5cm}

%{\color{blue}\begin{remark}\label{rem:boundary}
%EINBAUEN: The proof establishes the equivalence on $(-1,1)$. If $\psi_t \in C^1(-1,1) \cap C[-1,1]$ and $\bar p_t \in C^1[I \times [-1,1]]$, the equivalence extends to $[-1,1]$ as follows. Both directions of the proof use the residual
%$$
%\bar R(s) = \partial_t \bar p_t(s) + g_t'(s)(1-s^2) - %(d-1)\,s\,g_t(s), \quad g_t = \bar p_t\,\psi_t,
%$$
%which is defined on $(-1,1)$ where the continuity equation and the flux equation each imply $\bar R = 0$. Since $\bar p_t \in C[-1,1]$ and $\psi_t \in C[-1,1]$, their product $g_t$ is in $C[-1,1]$. On $(-1,1)$, the identity $\bar R = 0$ gives
%$$
%g_t'(s)(1-s^2) = -\partial_t \bar p_t(s) + (d-1)\,s\,g_t(s).
%$$
%The right-hand side is in $C[-1,1]$, so each term in $\bar R$ has a limit at $s = 1$ and at $s = -1$. Since $\bar R = 0$ on $(-1,1)$, these limits are zero.
%\end{remark}
%}

\psiode*
\paragraph{Proof.}
For the vMF setting, the flux equation~\eqref{eq:1d_flux} reads as
\begin{align}
 0&=   \partial_t f_t + \partial_s\big(f_t \, \psi_t \,(1-s^2)\, \big) \\
 &=
 (1-s^2)^{(d-3)/2} \partial_t \Big(C_d(\kappa_t)\,\exp(\kappa_t \,s) \Big)
 + C_d(\kappa_t) \partial_x \Big( (1-s^2)^{(d-1)/2} \psi_t (s) \exp(\kappa_t \,s) \Big).\label{eq:vmf_flux_expanded}
\end{align}
Concerning the first summand, we obtain
\begin{align}\label{eq:vmf_flux_dt}
\partial_t \Big(C_d(\kappa_t)\,\exp(\kappa_t \,s) \Big) 
&=
C_d'(\kappa_t) \, \dot \kappa_t \,\exp(\kappa_t \,s) + s \dot \kappa_t C_d(\kappa_t)\,\exp(\kappa_t \,s).
\end{align}
Noting that 
\begin{equation} \label{helper}
\mathcal I_{d/2-1}'(\kappa_t) = \mathcal I_{d/2}(\kappa_t) + \frac{d/2-1}{\kappa_t} \mathcal I_{d/2-1}(\kappa_t),
\end{equation}
we conclude
\begin{align}
C_d'(\kappa_t) &= 
\frac{(\frac{d}{2} -1)\kappa_t^{d/2-2}}{(2\pi)^{d/2} \mathcal I_{d/2-1}(\kappa_t)}
- 
\frac{\kappa_t^{d/2-1} \mathcal I_{d/2-1}'(\kappa_t)}{(2\pi)^{d/2}\mathcal I_{d/2-1}^2(\kappa_t) }
\\
&=
\frac{(\frac{d}{2}-1)}{\kappa_t} C_d(\kappa_t)
- \frac{(\frac{d}{2}-1)}{\kappa_t} C_d(\kappa_t)
-  A_d(\kappa_t) C_d(\kappa_t)  \\
&= -A_d(\kappa_t) C_d(\kappa_t)
\end{align}
and consequently
\begin{align}\label{eq:vmf_flux_term1}
\partial_t \Big(C_d(\kappa_t)\,\exp(\kappa_t \,s) \Big) 
&=
C_d(\kappa_t) \,\exp(\kappa_t \,s)  \dot{\kappa}_t \left(
-A_d(\kappa_t) 
+ s \right).
\end{align}
The second summand can be rewritten as
\begin{align}\label{eq:vmf_flux_term2}
&\partial_s \Big( (1-s^2)^{(d-1)/2} \psi_t (s) \exp(\kappa_t \,s) \Big)
 =
-(d-1) (1-s^2) ^{(d-3)/2} \, s \, \psi_t(s) \exp(\kappa_t \,s)\\
&\quad + (1-s^2)^{(d-1)/2}\psi_t' (s) \exp(\kappa_t \,s)
+\kappa_t (1-s^2)^{(d-1)/2}\psi_t(s)\exp(\kappa_t \,s)
\end{align}
Plugging in \eqref{eq:vmf_flux_term1} and \eqref{eq:vmf_flux_term2} into \eqref{eq:vmf_flux_expanded}
and dividing by $C_d(\kappa_t) (1-s^2)^{(d-3)/2} \exp(\kappa_t \,s)$ yields
\begin{align}
\dot{\kappa}_t \left(
A_d(\kappa_t)- s \right)
&=
\big(-(d-1)  s + (1-s^2)\kappa_t\big) \, \psi_t(s) 
+ (1-s^2) \psi_t' (s).
\end{align}
This finishes the proof. \hfill $\Box$
\vspace{0.5cm}

\psiwellposed*
\paragraph{Proof.}
The general solution of the linear ODE
\begin{equation}
     \psi' + \left( \kappa - (d-1)\frac{s}{1-s^2}\right)\psi =  \left(A_d - s \right) \frac{\dot{\kappa}}{1-s^2}, 
     \quad s \in (-1,1)
\end{equation}
(with skipped index $t$ for convenience)
can be computed by the method of integrating factors as
$$
\psi(s) = \exp(-\kappa s) (1-s^2)^{-(d-1)/2} 
\Big( C + \dot \kappa \int_{-1}^s \exp(\kappa y) (1-y^2)^{(d-3)/2} (A_d-y) \, \rm{d} y \Big),
$$
where the constant $C$ is determined by  conditions on  $\psi$.
We are interested in solutions with finite $\lim_{s \to \pm 1}\psi(s)$. 
Since $(1-s^2)^{-(d-1)/2}$ goes to infinity as $s \to \pm 1$,
this can  only be achieved for $C=0$ and the solution
\begin{equation} \label{h1}
\psi(s) = \dot \kappa \, \exp(-\kappa s) (1-s^2)^{-(d-1)/2} 
 \int_{-1}^s \exp(\kappa y) (1-y^2)^{(d-3)/2} (A_d-y) \, \rm{d} y .
\end{equation}
Indeed the integral behaves for $s$ near $-1$ as $(s+1)^{(d-1)/2}$, so that $\psi(-1)$ is finite and the concrete value follows immediately from
\eqref{eq:psi_ode}. On the other hand, considering $s \to 1$, we have
$$
\int_{-1}^1 \exp(\kappa y) (1-y^2)^{(d-3)/2} (A_d-y) \, {\rm{d}} y
=
A_d G(\kappa) -   G'(\kappa)
$$
where
$$
G(\kappa) \coloneqq \int_{-1}^1 \exp(\kappa y) (1-y^2)^{(d-3)/2} \, {\rm{d}} y
= \mathcal I_{d/2-1}(\kappa) \alpha(\kappa), \quad \alpha(\kappa) \coloneqq
\frac{\Gamma((d-1)/2) \sqrt{\pi}}{(\kappa/2)^{(d/2-1)}}
$$
and using \eqref{helper} and the definition of $A_d$ further
\begin{align*}
A_d G(\kappa) -  G'(\kappa) &= \mathcal I_{d/2} (\kappa) \alpha(\kappa) 
- \big(  \mathcal I_{d/2}(\kappa) + \frac{d/2-1}{\kappa}  \mathcal I_{d/2-1}(\kappa)\big) \alpha(\kappa) - I_{d/2-1}(\kappa) \alpha'(\kappa)
=0.
\end{align*}
The asymptotics of the integral in \eqref{h1} as $s \to 1$ is $(s-1)^{(d-1)/2}$.
Finally, the representation \eqref{h1} is a product of functions that are smooth on $(-1,1)$, so that $\psi \in C^\infty(-1,1)$ 
and we are done. \hfill $\Box$
\vspace{0.5cm}

\conditionalvelocity*
\paragraph{Proof.}
By the previous proofs we have for the corresponding expressions in \eqref{eq:1d_flux}, \eqref{eq:psi_ode} and \eqref{eq:cont_sphere} that
\begin{align}
\eqref{eq:1d_flux} &= C_d(\kappa_t) (1-s^2)^{(d-3)/2} \exp(\kappa_t s) \cdot \eqref{eq:psi_ode} \\
&=(1-s^2)^{(d-3)/2} \cdot \eqref{eq:cont_sphere},
\end{align}
where $\eqref{eq:cont_sphere} = R(s) = \partial_t \bar p_t(s) + g_t'(s) (1-s^2) - g_t(s) (d-1) s$ and $g_t(s) = C_d(\kappa_t)  \exp(\kappa_t s) \psi_t(s)$. This gives the assertion for $s \in (-1,1)$. For the chosen $\psi$, we have that $R$ is continuous on $[-1,1]$ (note that $\psi_t'(\pm 1)$ is also finite) and since $R$ is zero on $(-1,1)$ by \eqref{eq:psi_ode},
it remains zero at the boundary. So we have an extension of the relation to the whole $\mathbb S^{d-1}$. \hfill $\Box$
\vspace{0.5cm}

\signal*
\paragraph{Proof.}
Since \eqref{eq:vmf_signal} was already stated, it remains to verify \eqref{eq:geo_signal}. For
$$
X_t = \frac{\sin((1-t)\theta)}{\sin\theta}\, X_0 + \frac{\sin(t\theta)}{\sin\theta}\, w,
$$
let $Z_t \coloneqq \langle w,X_t \rangle$.
By Remark \ref{rem:geo_path}, we know that
$$
Z_t = \cos \left( (1-t) \Theta\right), \quad \Theta = \arccos \left(\langle w,X_0 \rangle \right).
$$
It is straightforward to check that $Z_0$ is a random variable on $[-1,1]$ with 
expectation value zero  and variance $1/d$. Thus, $Z_0 \to 0$ as $d \to \infty$ in probability and consequently $\Theta \to \pi/2$ in probability as $d \to \infty$, and finally
$$
Z_t \to \cos \left( (1-t) \frac{\pi}{2} \right) \quad \text{as} \quad d \to \infty
$$
in probability. Then the bounded convergence theorem gives
$$
\mathbb E[Z_t] \to \cos \Big( (1-t) \frac{\pi}{2} \Big) = \sin \Big(\frac{\pi}{2}\,t\Big) \quad \text{as} \quad d \to \infty. \quad \Box
$$

\vspace{0.5cm}

\loss*
\paragraph{Proof.}
We show that the loss \eqref{eq:ce_loss} decomposes as
\begin{equation}\label{eq:loss_ce}
\mathcal L(\theta) = \sum_{l=1}^L \mathbb E_{t \sim \mathcal U(0,1),\, \mathbf x_t \sim p_t}\Big[\mathrm{CE}\big(p_t^l(\cdot|\mathbf x_t),\, p_{\theta,t}^l(\cdot|\mathbf x_t)\big)\Big],
\end{equation}
from which the claim follows by the minimization property above. Write $\mathcal L = \sum_{l=1}^L \mathcal L^l$, where
\begin{align}
\mathcal L^l (\theta)
    &\coloneqq  - \int \int
    \sum_{\mathbf w \in \mathcal W^L} 
    \log p_{\theta,t}^l (w^l|\hat x_t^l) \,  p_{t}(\mathbf x_t|\mathbf w)
    \, p_{\text{data}} (\mathbf w)
    \, \mathrm d \mathbf x_t \, \mathrm d t.
\end{align}    
By Bayes' rule and since $\log p_\theta^l(w^l|\hat x_t^l)$ depends on $\mathbf w$ only through $w^l$, this can be rewritten as
\begin{align}
\mathcal L^l (\theta)
&= - \int \int
    \sum_{\mathbf w \in \mathcal W^L} 
    \log p_{\theta,t}^l (w^l|\hat x_t^l) \, p_t(\mathbf w|\mathbf x_t) \, p_t(\mathbf x_t)
    \, \mathrm d \mathbf x_t \, \mathrm d t \\
    &= - \int \int
    \sum_{w^l \in \mathcal W} 
    \log p_{\theta,t}^l (w^l|\hat x_t^l) 
    \sum_{w^j \in \mathcal W \atop j \not = l} p_t(\mathbf w|\mathbf x_t) \, p_t(\mathbf x_t)
    \, \mathrm d \mathbf x_t \, \mathrm d t \\
    &= \mathbb E_{t,\,\mathbf x_t \sim p_{t}}
\Big[
\mathrm{CE}\big(p_t^l(\cdot|\mathbf x_t),\, p_{\theta,t}^l (\cdot|\hat x_t^l) \big) \Big],
\end{align}
which is \eqref{eq:loss_ce}.
\hfill $\Box$

\subsection{SDE Sampling on the Sphere}\label{sec:sde_app}

\subsubsection{Fokker--Planck and the flow SDE}
The same flow of measures from Section~\ref{sec:model} admits a stochastic realization once the score is available. Substituting $v_t = u_t - \tfrac{\sigma^2}{2} \nabla_{\mathcal M} \log p_t$ for $\sigma > 0$ and using $\mathrm{div}_{\mathcal M}(\nabla_{\mathcal M} g) = \Delta_{\mathcal M} g$, the continuity equation \eqref{eq:ce} becomes the \textit{Fokker--Planck equation}
\begin{equation}\label{eq:fp}
\partial_t p_t + \mathrm{div}_{\mathcal M}(p_t u_t) - \tfrac{\sigma^2}{2} \Delta_{\mathcal M} p_t = 0.
\end{equation}
The solution $p_t$ is the law of a random variable $X_t$ determined by the \textit{flow SDE}
\begin{equation}\label{eq:sde_allg}
\mathrm d X_t = u_t(X_t)\,\mathrm d t + \sigma\,\mathrm d B_t,
\tag{Flow SDE}
\end{equation}
where $B_t$ is the Brownian motion on $\mathcal M$. Once $v_t$ and $\nabla_{\mathcal M} \log p_t$ are known, sampling from $p_1$ can be done either deterministically via the flow ODE \eqref{eq:flow_ode} or stochastically via the SDE \eqref{eq:sde_allg}.

\subsubsection{Marginal SDE for the vMF path}
The score \eqref{eq:marginal_score_vmf} enables stochastic sampling along the same marginal path. The flow SDE on $(\mathbb S^{d-1})^L$ reads per position
\begin{equation}\label{eq:sde_app}
\mathrm d X^l_t = \Big[v_t^l(\mathbf X_t) + \tfrac{\sigma^2}{2} \nabla_{\mathbb S^{d-1},x^l} \log p_t(\mathbf X_t)\Big]\,\mathrm d t + \sigma\,\mathrm d B_t^l, \quad l = 1,\ldots,L,
\end{equation}
where $\sigma > 0$ is a free diffusion coefficient and $B_t^l$ is Brownian motion on $\mathbb S^{d-1}$. The derivation holds verbatim for a time-dependent diffusion coefficient $\sigma_t$, since the Fokker--Planck cancellation between the score part of the drift and the diffusion term in \eqref{eq:fp} holds pointwise in $t$ provided the drift uses the matching $\sigma_t^2/2$; in the experiments below we use $\sigma_t = \sigma (1 - u)^q$ with the path progress variable $u$ of Appendix~\ref{sec:sweep_choices} ($u = \kappa_t / \kappa_{\max}$ for vMF). Substituting \eqref{eq:marginal_velo_vmf} and \eqref{eq:marginal_score_vmf}, the drift of \eqref{eq:sde_app} is itself a posterior-weighted tangent sum,
\begin{equation}\label{eq:sde_drift_app}
v_t^l + \tfrac{\sigma^2}{2}\nabla_{\mathbb S^{d-1},x^l}\log p_t = \sum_{k=1}^N p_t^l(w_k|\mathbf x)\,\Big(\dot\kappa_t\,\tilde\psi_t(\langle w_k, x^l\rangle) + \tfrac{\sigma^2}{2}\,\kappa_t\Big)\,\mathrm P_{x^l}(w_k),
\end{equation}
with the same posteriors and tangent projections as the ODE. No separate score network is required. Algorithm~\ref{alg:sde_app} discretizes \eqref{eq:sde_app} via Euler--Maruyama with a tangent retraction.

\begin{algorithm}[H]
\caption{SDE sampling on $(\mathbb S^{d-1})^L$}
\label{alg:sde_app}
\begin{algorithmic}[1]
\Require Time grid $0 = t_0 < t_1 < \cdots < t_S = 1$, diffusion coefficient $\sigma > 0$
\State Sample $x_0^l \sim \mathcal{U}(\mathbb S^{d-1})$ independently for $l = 1,\ldots,L$
\For{$n = 0, \ldots, S-1$}
    \State $\Delta t \leftarrow t_{n+1} - t_n$
    \State $(\hat x^1, \ldots, \hat x^L) \leftarrow T_\theta(x_n^1, \ldots, x_n^L)$
    \State $p_k^l \leftarrow p_\theta^l(w_k|\hat x^l)$ for all $l,k$
    \State $d^l \leftarrow \displaystyle\sum_{k=1}^N p_k^l\,\Big(\dot\kappa_{t_n}\,\tilde\psi_{t_n}(\langle w_k, x_n^l\rangle) + \tfrac{\sigma^2}{2}\,\kappa_{t_n}\Big)\,\mathrm P_{x_n^l}(w_k)$
        \hfill $\triangleright$ \eqref{eq:sde_drift_app}
    \State Sample $\xi^l \sim \mathcal N(0, I_d)$ for $l = 1,\ldots,L$
    \State $x_{n+1}^l \leftarrow \dfrac{x_n^l + \Delta t\, d^l + \sigma\sqrt{\Delta t}\,\mathrm P_{x_n^l}(\xi^l)}{\|x_n^l + \Delta t\, d^l + \sigma\sqrt{\Delta t}\,\mathrm P_{x_n^l}(\xi^l)\|}$ for $l = 1,\ldots,L$
\EndFor
\State $(\hat x^1, \ldots, \hat x^L) \leftarrow T_\theta(x_S^1, \ldots, x_S^L)$
\State \Return $\hat y^l = \argmax_{k=1,\ldots,N}\, \langle w_k, \hat x^l \rangle + b_k$ for $l = 1,\ldots,L$
\end{algorithmic}
\end{algorithm}

\subsection{Further Radially Symmetric Paths}\label{app:further_paths}
%-----------------------------------------------------------------
Beyond the vMF path of Section~\ref{sec:vM}, the reduction of Theorem~\ref{thm:radial_continuity} can be read in two directions: fixing the radial density $\bar p_t$ and solving the flux equation \eqref{eq:1d_flux} for the velocity, or fixing the velocity and solving for the density. In this appendix we first state the closed velocity formula obtained in the density-first direction, and then instantiate it for two further conditional paths, the power spherical path and the spherical Cauchy path, summarized together with the geodesic and vMF paths in Table~\ref{tab:paths}.

\subsubsection{Velocity of a radial density}
Fixing the radial density $\bar p_t$ and solving for the velocity gives a closed formula, obtained by integrating the flux equation \eqref{eq:1d_flux} once in $s$.

\begin{restatable}[Velocity of a radial path]{proposition}{mastervelocity}\label{prop:master_velocity}
Let $(p_t)_{t \in I}$ be radially symmetric densities with profiles $\bar p_t \in C^1(I \times (-1,1))$, and let $f_t$ be as in \eqref{eq:f_mu_def}. Assume $f_t(s)\,\psi_t(s)\,(1-s^2) \to 0$ as $s \to -1$. Then the velocity coefficient for which the pair $(p_t, v_t)$ with $v_t(x) = \psi_t(\langle w,x\rangle)\,\mathrm P_x(w)$ satisfies the continuity equation \eqref{eq:cont_sphere} is
\begin{equation}\label{eq:master_velocity}
    \psi_t(s) = -\frac{1}{f_t(s)\,(1-s^2)} \int_{-1}^{s} \partial_t f_t(y)\, \dd y.
\end{equation}
\end{restatable}

The corresponding radial score is immediate from \eqref{rad_ab}, since the Riemannian gradient of a radial density is
\begin{equation}\label{eq:radial_score}
    \nabla_{\mathbb S^{d-1}} \log p_t(x) = (\log \bar p_t)'(\langle w,x\rangle)\, \mathrm P_x(w).
\end{equation}
The integral in \eqref{eq:master_velocity} is the unique velocity whose flux vanishes at both endpoints; it is smooth on $(-1,1)$, and its endpoint values follow from the flux equation once the profile is fixed. Conversely, fixing the velocity coefficient $\psi_t$ and solving the flux equation \eqref{eq:1d_flux} for $f_t$ recovers the radial density transported by that velocity.

When the profile is indexed by a scalar concentration schedule, $\bar p_t = \bar p(\cdot\,; c_t)$ with $c_t$ monotone and $c_0 = 0$, the time derivative factorizes as $\partial_t f_t = \dot c_t\,\partial_c f$, and so does the velocity,
\begin{equation}\label{eq:schedule_factor}
    \psi_t(s) = \dot c_t\,\tilde\psi(s; c_t), \qquad
    \tilde\psi(s; c) = -\frac{1}{f(s;c)(1-s^2)}\int_{-1}^s \partial_c f(y;c)\,\dd y,
\end{equation}
where $\tilde\psi$ depends on the concentration value alone, not on the schedule. For the vMF path, specializing to $c = \kappa$ recovers exactly the representation \eqref{eq:psi_tilde} of Remark~\ref{rem:46}. The factor $\tilde\psi$ is tabulated once in $(s, c)$ and evaluated by a numerically stable cumulative sum (Section~\ref{sec:paths_app}), and in concentration-space integration $\dot c_t$ cancels, so the schedule enters sampling only through the step sizes $\Delta c$. We use the density-first direction for both the power spherical and the spherical Cauchy paths below; for the spherical Cauchy path the resulting velocity coefficient is constant in $s$.

\subsubsection{Two further instances}
Each of the following paths has velocity $\psi_t(s)\,\mathrm P_x(w)$ and, by \eqref{eq:radial_score}, score $(\log \bar p_t)'(s)\,\mathrm P_x(w)$ with $s = \langle w,x\rangle$.

\paragraph{Power spherical.}
As a further density-first instance, we take the power spherical family, whose radial profile is a power of $1+s$. For a monotone schedule $\alpha_t$ with $\alpha_0 = 0$ we set
\begin{equation}\label{eq:power_density}
    p_t(\cdot \mid w) \coloneqq C(\alpha_t)\,(1 + \langle w, \cdot\rangle)^{\alpha_t}, \qquad
    C(\alpha)^{-1} = 2^{\,d-1+\alpha}\,\pi^{(d-1)/2}\,\frac{\Gamma(\alpha + \tfrac{d-1}{2})}{\Gamma(\alpha + d - 1)},
\end{equation}
so that $p_0$ is uniform. This is radially symmetric, and evaluating the velocity formula \eqref{eq:master_velocity} is equivalent to a linear ODE.

\begin{restatable}[Power spherical velocity]{lemma}{powervelocity}\label{lem:power_velocity}
For the profile \eqref{eq:power_density}, the velocity coefficient in \eqref{eq:master_velocity} is the unique solution bounded as $s \to 1$ of
\begin{equation}\label{eq:power_ode}
    (1-s^2)\,\psi_t' + \big[\alpha_t(1-s) - (d-1)s\big]\,\psi_t
    = -\dot\alpha_t\Big[\log\tfrac{1+s}{2} + h(d-1+\alpha_t) - h\big(\tfrac{d-1}{2}+\alpha_t\big)\Big],
\end{equation}
where $h = (\log\Gamma)'$ is the digamma function, with boundary value $\psi_t(1) = \tfrac{\dot\alpha_t}{d-1}\big[h(d-1+\alpha_t) - h(\tfrac{d-1}{2}+\alpha_t)\big]$.
\end{restatable}

By \eqref{eq:radial_score} the score is $\nabla_{\mathbb S^{d-1}}\log p_t(x) = \tfrac{\alpha_t}{1+\langle w,x\rangle}\,\mathrm P_x(w)$. As for vMF the coefficient $\psi_t$ has no elementary closed form; it is precomputed on the same $(\mu, \alpha)$ grid as the vMF table via the flux representation \eqref{eq:master_velocity}, using the same max-shifted log-density trick to avoid underflow, with the analytical boundary value $\psi_t(1)$ from Lemma~\ref{lem:power_velocity} handled as for vMF (Section~\ref{sec:paths_app}). Unlike vMF the profile vanishes at the antipode $s = -1$, where $\psi_t$ diverges logarithmically; the velocity field itself stays bounded, since $|\mathrm P_x(w)| = \sqrt{1-s^2}$ gives velocity magnitude $|\psi_t(s)|\sqrt{1-s^2} \to 0$ as $s \to -1$. Training-time samples $x \sim p_t(\cdot \mid w)$ use the same radial-plus-tangent decomposition as vMF, with $\mu$ drawn by inverse CDF of $(1-\mu^2)^{(d-3)/2}(1+\mu)^{\alpha_t}$.

\paragraph{Spherical Cauchy.}
The \emph{Cauchy distribution} on $\mathbb S^{d-1}$
with location  $w \in \mathbb S^{d-1}$ and concentration $r \in [0,1)$ has the density
\begin{align}
\varphi(x; w,r) = \frac{1}{\omega_{d-1}} \frac{( 1-  r^2)^{d-1}}{(1+ r^2 - 2 r \langle w,x\rangle)^{d-1}}
\qquad \omega_{d-1} = \frac{2\pi^{d/2}}{\Gamma(d/2)},
\end{align}.
For $r = 0$, we obtain the uniform distribution, while $r \to 1$ pushes the distribution to $w$.
It is the image of the multivariate Cauchy distribution under stereographic projection.
Setting $a \coloneqq  (2r)/(1+r^2)$, this can be rewritten as
\begin{equation}
\varphi(x; w,r) =
\frac{1}{\omega_{d-1}}\,\frac{(1 - a^2)^{(d-1)/2}}{\big(1 - a\langle w,x\rangle\big)^{d-1}}.
\end{equation}
With a schedule $a_t: [0,\infty) \to [0,1)$, we consider
\begin{equation}\label{eq:cauchy_density}
    p_t(x \mid w) = \frac{1}{\omega_{d-1}}\,\frac{(1 - a_t^2)^{(d-1)/2}}{\big(1 - a_t\langle w,x\rangle\big)^{d-1}}.
\end{equation}
By the following theorem, the corresponding velocity coefficient
in the CE \eqref{eq:1d_flux} is especially simple, since it does not depend on $s$.

\begin{restatable}[Spherical Cauchy velocity]{proposition}{cauchypatha}\label{thm:cauchya}
For the profile \eqref{eq:cauchy_density}, the velocity coefficient $\psi$ in \eqref{eq:master_velocity} is the unique solution which is bounded as $s \to 1$ of
\begin{equation}\label{eq:cauchy_ode}
\partial_s \psi_t(s)
+
(d-1)\left(
\frac{a_t}{1-a_t s}
-
\frac{s}{1-s^2}
\right)\psi_t(s)
=
-\frac{(d-1)\dot a_t}{1-s^2}
\left(
\frac{s}{1-a_t s}
-
\frac{a_t}{1-a_t^2}
\right).
\end{equation}
This solution is given by $$\psi_t(s) = \dot \beta_t = \frac{\dot a_t}{1-a_t^2},$$
where
\begin{equation} \label{schedule_cauchy}
\beta_t \coloneqq \tfrac{1}{2}\log \frac{1+ a_t}{1-a_t}, \quad a_t \coloneqq \tanh\beta_t.
\end{equation}
The score reads as
\begin{equation}\label{eq:cauchy_score}
    \nabla_{\mathbb S^{d-1}}\log p_t(x) = \frac{(d-1)\,a_t}{1 - a_t\langle w,x\rangle}\,\mathrm P_x(w).
\end{equation}
\end{restatable}

Training-time samples use $x = \mu\,w + \sqrt{1-\mu^2}\,v$ with $v$ a uniform tangent direction and $\mu$ drawn from the radial marginal $\propto (1-\mu^2)^{(d-3)/2}(1 - a_t\mu)^{-(d-1)}$, again a one-dimensional inverse-CDF draw.

\begin{table}[t]
\centering
\small
\setlength{\tabcolsep}{5pt}
\caption{Radially symmetric conditional paths on $\mathbb S^{d-1}$ as instances of Theorem~\ref{thm:radial_continuity}.}
\label{tab:paths}
\small
\begin{tabular}{@{}llll@{}}
\toprule
Path & Radial profile $\bar p_t(s)$ & Velocity coeff.\ $\psi_t(s)$ & Score coeff.\ $(\log \bar p_t)'(s)$ \\
\midrule
Geodesic & -- & $\dfrac{\arccos s}{(1-t)\sqrt{1-s^2}}$ & -- \\[1.4ex]
vMF & $\propto e^{\kappa_t s}$ & ODE \eqref{eq:psi_ode} & $\kappa_t$ \\[0.6ex]
Power sph. & $\propto (1+s)^{\alpha_t}$ & ODE \eqref{eq:power_ode} & $\dfrac{\alpha_t}{1+s}$ \\[1.4ex]
Cauchy & $\propto (1 - a_t s)^{-(d-1)}$ & $\dot\beta_t$ & $\dfrac{(d-1)\,a_t}{1 - a_t s}$ \\
\bottomrule
\end{tabular}
\end{table}
In Table~\ref{tab:paths}, every velocity has the form $\psi_t(s)\,\mathrm P_x(w)$ and every closed-form score the form $(\log \bar p_t)'(s)\,\mathrm P_x(w)$ with $s = \langle w,x\rangle$. vMF and power spherical obtain $\psi_t$ from the quadrature \eqref{eq:master_velocity}; for the Cauchy path $\psi_t = \dot\beta_t$ is constant in $s$, with $a_t = \tanh\beta_t$. Geodesic and Cauchy have $\psi_t$ in closed form.

\subsubsection{Proofs}

\mastervelocity*
\paragraph{Proof.}
By Theorem~\ref{thm:radial_continuity}, the pair $(p_t, v_t)$ solves \eqref{eq:cont_sphere} if and only if $(f_t, \psi_t)$ solves the flux equation \eqref{eq:1d_flux}, that is $\partial_t f_t + \partial_s\big(f_t \psi_t (1-s^2)\big) = 0$. Integrating in $s$ from $-1$ to $s$ gives
$$
\int_{-1}^s \partial_t f_t(y)\,\dd y + f_t(s)\psi_t(s)(1-s^2) - \lim_{y \to -1^+} f_t(y)\psi_t(y)(1-y^2) = 0.
$$
The boundary term vanishes by assumption, and solving for $\psi_t$ yields \eqref{eq:master_velocity}. \hfill $\Box$
\vspace{0.5cm}

\powervelocity*
\paragraph{Proof.}
For the profile \eqref{eq:power_density} we have $f_t(s) = C(\alpha_t)(1-s^2)^{(d-3)/2}(1+s)^{\alpha_t}$, and the flux equation \eqref{eq:1d_flux} reads
$$
(1-s^2)^{(d-3)/2}\,\partial_t\big(C(\alpha_t)(1+s)^{\alpha_t}\big) + C(\alpha_t)\,\partial_s\big((1-s^2)^{(d-1)/2}\psi_t(s)(1+s)^{\alpha_t}\big) = 0.
$$
For the first term, $\partial_t\big(C(\alpha_t)(1+s)^{\alpha_t}\big) = \dot\alpha_t\big[\tfrac{C'(\alpha_t)}{C(\alpha_t)} + \log(1+s)\big]C(\alpha_t)(1+s)^{\alpha_t}$. From \eqref{eq:power_density}, $\log C(\alpha) = -(d-1+\alpha)\log 2 - \tfrac{d-1}{2}\log\pi - \log\Gamma(\alpha+\tfrac{d-1}{2}) + \log\Gamma(\alpha+d-1)$, so with $h = (\log\Gamma)'$,
$$
\frac{C'(\alpha)}{C(\alpha)} = -\log 2 + h(d-1+\alpha) - h\big(\tfrac{d-1}{2}+\alpha\big).
$$
Expanding the derivative in the second term and dividing the whole equation by $C(\alpha_t)(1-s^2)^{(d-3)/2}(1+s)^{\alpha_t}$, using $(1-s^2)(1+s)^{-1} = 1-s$, gives
$$
\dot\alpha_t\Big[\tfrac{C'(\alpha_t)}{C(\alpha_t)} + \log(1+s)\Big] + (1-s^2)\psi_t' + \big[\alpha_t(1-s) - (d-1)s\big]\psi_t = 0,
$$
which is \eqref{eq:power_ode} after substituting $\tfrac{C'(\alpha_t)}{C(\alpha_t)}$ and collecting $\log(1+s) - \log 2 = \log\tfrac{1+s}{2}$. As in Lemma~\ref{prop:psi_wellposed}, boundedness at $s = 1$ selects a unique solution; letting $s \to 1$ in \eqref{eq:power_ode} and using $\log 1 = 0$ gives $-(d-1)\psi_t(1) = -\dot\alpha_t[h(d-1+\alpha_t) - h(\tfrac{d-1}{2}+\alpha_t)]$, the stated boundary value. \hfill $\Box$
\vspace{0.5cm}

\cauchypatha*
\paragraph{Proof.}
Consider the radially symmetric velocity field $v_t(x) = \dot\beta_t\,\mathrm P_x(w)$ of the form \eqref{eq:radial_vf}, with constant coefficient $\psi_t(s) = \dot\beta_t$ and $\beta_t = \operatorname{artanh} a_t$ as in \eqref{schedule_cauchy}. Under this velocity the cosine similarity $s_t = \langle w, X_t\rangle$ obeys $\dot s_t = \langle w, v_t(X_t)\rangle = \psi_t(s_t)\langle w, \mathrm P_{X_t}(w)\rangle = \dot\beta_t(1-s_t^2)$. Separating variables gives $\operatorname{artanh}(s_t) = \beta_t + \operatorname{artanh}(s_0)$, which by the addition formula for $\tanh$ is the fractional-linear (M\"obius) map
$$
s_t = \frac{s_0 + a_t}{1 + a_t\, s_0}, \qquad a_t = \tanh\beta_t.
$$

It remains to identify the law of $s_t$. For $X_0 \sim \mathcal U(\mathbb S^{d-1})$ the cosine $s_0$ has density $f_0(s) \propto (1-s^2)^{(d-3)/2}$ by \eqref{eq:f_mu_def}. Writing $\Phi(s_0) = (s_0 + a_t)/(1 + a_t s_0)$ for this map, its inverse is $\Phi^{-1}(s) = (s - a_t)/(1 - a_t s)$ with
$$
1 - \Phi^{-1}(s)^2 = \frac{(1-a_t^2)(1-s^2)}{(1-a_t s)^2}, \qquad
\big(\Phi^{-1}\big)'(s) = \frac{1-a_t^2}{(1-a_t s)^2}.
$$
By the change of variables formula the density of $s_t = \Phi(s_0)$ is
\begin{align*}
f_t(s) = f_0\big(\Phi^{-1}(s)\big)\,\big(\Phi^{-1}\big)'(s)
&\;\propto\; \Big(\tfrac{(1-a_t^2)(1-s^2)}{(1-a_t s)^2}\Big)^{(d-3)/2}\frac{1-a_t^2}{(1-a_t s)^2} \\
&\;\propto\; (1-s^2)^{(d-3)/2}\,(1 - a_t s)^{-(d-1)}.
\end{align*}
Dividing by $(1-s^2)^{(d-3)/2}$, the radial profile is $\bar p_t(s) \propto (1 - a_t s)^{-(d-1)}$. As a pushforward of a probability density, $f_t$ is normalized, and matching the profile to the normalized form identifies the constant, so the transported density is exactly \eqref{eq:cauchy_density}.

As pushforward densities of the flow of $\dot s_t = \dot\beta_t(1-s_t^2)$, the family $f_t$ satisfies $\partial_t f_t + \partial_s\big(f_t\,\dot\beta_t\,(1-s^2)\big) = 0$, which is the flux equation \eqref{eq:1d_flux} with $\psi_t = \dot\beta_t$. For the profile \eqref{eq:cauchy_density} we have $f_t(s) \propto (1-a_t^2)^{(d-1)/2}(1-s^2)^{(d-3)/2}(1-a_t s)^{-(d-1)}$, so
$$
\frac{\partial_t f_t}{f_t} = (d-1)\,\dot a_t\left(\frac{s}{1-a_t s} - \frac{a_t}{1-a_t^2}\right), \qquad
\frac{\partial_s\big(f_t\,(1-s^2)\big)}{f_t\,(1-s^2)} = (d-1)\left(\frac{a_t}{1-a_t s} - \frac{s}{1-s^2}\right),
$$
and dividing \eqref{eq:1d_flux} by $f_t(s)(1-s^2)$ gives \eqref{eq:cauchy_ode}. Hence the constant $\psi_t(s) = \dot\beta_t = \dot a_t/(1-a_t^2)$ solves \eqref{eq:cauchy_ode}, and it is bounded as $s \to 1$ since it does not depend on $s$.

For uniqueness, the difference of two solutions of the linear equation \eqref{eq:cauchy_ode} solves the homogeneous equation, whose solutions are proportional to
$$
\exp\left(-(d-1)\int \Big(\frac{a_t}{1-a_t s} - \frac{s}{1-s^2}\Big) \dd s\right) = (1-a_t s)^{d-1}\,(1-s^2)^{-(d-1)/2},
$$
which is unbounded as $s \to 1$ for $d \ge 2$ and $0 \le a_t < 1$. Hence the bounded solution is unique.

Finally, by \eqref{eq:radial_score} the score is $(\log \bar p_t)'(s)\,\mathrm P_x(w) = \tfrac{(d-1)\,a_t}{1-a_t s}\,\mathrm P_x(w)$, which is \eqref{eq:cauchy_score}. \hfill $\Box$
\vspace{0.5cm}

\newpage
\section{Implementation Details}\label{sec:impl_details}
\begin{algorithm}[h]
\caption{ODE / predictor-corrector sampling on $(\mathbb S^{d-1})^L$}
\label{alg:ode}
\begin{algorithmic}[1]
\Require Time grid $0 = t_0 < t_1 < \cdots < t_n = 1$, corrector steps $k \geq 0$, step size $\varepsilon > 0$
\State Sample $x_0^l \sim \mathcal U(\mathbb S^{d-1})$ independently for $l = 1, \ldots, L$
\For{$i = 0, \ldots, n - 1$}
    \State $(\hat x^1, \ldots, \hat x^L) \leftarrow T_\theta(x_i^1, \ldots, x_i^L)$
    \State $p_j^l \leftarrow p_\theta^l(w_j \mid \hat x^l)$ for all $l, j$
    \State $v^l \leftarrow \dot\kappa_{t_i} \displaystyle\sum_{j=1}^N p_j^l\,\tilde\psi_{t_i}(\langle w_j, x_i^l\rangle)\,\mathrm P_{x_i^l}(w_j)$ for $l = 1, \ldots, L$
        \hfill $\triangleright$ \eqref{eq:marginal_velo_vmf}
    \State $x_{i+1}^l \leftarrow \dfrac{x_i^l + (t_{i+1} - t_i)\, v^l}{\|x_i^l + (t_{i+1} - t_i)\, v^l\|}$ for $l = 1, \ldots, L$
    \For{$j = 1, \ldots, k$} \Comment{Langevin corrector}
        \State $(\hat x^1, \ldots, \hat x^L) \leftarrow T_\theta(x_{i+1}^1, \ldots, x_{i+1}^L)$
            \hfill $\triangleright$ fresh forward at corrected state
        \State $p_m^l \leftarrow p_\theta^l(w_m \mid \hat x^l)$ for all $l, m$
        \State Sample $\xi^l \sim \mathcal N(0, I_d)$ for $l = 1, \ldots, L$
        \State $g^l \leftarrow \kappa_{t_{i+1}} \displaystyle\sum_{m=1}^N p_m^l\,\mathrm P_{x_{i+1}^l}(w_m)$ for $l = 1, \ldots, L$
            \hfill $\triangleright$ score, \eqref{eq:marginal_score_vmf}
        \State $\eta^l \leftarrow \mathrm P_{x_{i+1}^l}\big(\varepsilon\, g^l + \sqrt{2\varepsilon}\,\xi^l\big)$ for $l = 1, \ldots, L$
            \hfill $\triangleright$ project to tangent
        \State $x_{i+1}^l \leftarrow \dfrac{x_{i+1}^l + \eta^l}{\|x_{i+1}^l + \eta^l\|}$ for $l = 1, \ldots, L$
            \hfill $\triangleright$ retract to sphere
    \EndFor
\EndFor
\State $(\hat x^1, \ldots, \hat x^L) \leftarrow T_\theta(x_n^1, \ldots, x_n^L)$
\State \Return $\hat y^l = \argmax_{j = 1, \ldots, N}\, \langle w_j, \hat x^l \rangle + b_j$ for $l = 1, \ldots, L$
    \hfill $\triangleright$ \eqref{eq:decoding}
\end{algorithmic}
\end{algorithm}
\subsection{Architecture and Training}\label{sec:arch_app}

All continuous methods share a DiT-style transformer backbone with adaLN conditioning. When time conditioning is off, adaLN reduces to a learnable per-block scale, shift, and gate. Otherwise it receives the schedule parameter normalized to $[0, 1]$. Embeddings are tied to the softmax classifier head as described in Section \ref{sec:model_2}. We use AdamW with mixed-precision (bf16) training, a linear warmup followed by a cosine learning-rate schedule, and an exponential moving average on the network weights, which is the version used at sampling time. Masked diffusion uses the same backbone but with a discrete embedding layer of width equal to the hidden size, replacing the tied learned embeddings on the manifold. Per-task hyperparameters are listed in Table \ref{tab:arch_hparams}. The TinyGSM model size and training recipe ($4\times$A100, $250$k steps) follow \cite{deschenaux2026hyperspherical}.

\begin{table}[h]
\centering
\small
\setlength{\tabcolsep}{5pt}
\begin{tabular}{lcccc}
\toprule
 & Sudoku & LM1B & TinyGSM & OpenWebText \\
\midrule
Hidden size       & 512    & 768 & 768 & 768 \\
Blocks            & 8      & 12  & 12  & 12 \\
Parameters (non-embed.) & 28.6M  & 93.0M & 93.0M & 93.0M \\
Sequence length $L$ & 81     & 128 & 512 & 1{,}024 \\
Vocabulary $N$    & 10     & 30{,}522 & 49{,}152 & 50{,}257 \\
\midrule
Training steps    & 1{,}000{,}000 & 1{,}000{,}000 & 250{,}000 & 1{,}000{,}000 \\
Batch size (effective) & 128 & 512 & 512 & 512 \\
Learning rate     & $3 \times 10^{-4}$ & $3 \times 10^{-4}$ & $3 \times 10^{-4}$ & $3 \times 10^{-4}$ \\
Warmup steps      & 1{,}000 & 2{,}500 & 2{,}500 & 2{,}500 \\
EMA decay         & 0.999 & 0.9999 & 0.9999 & 0.9999 \\
Dropout           & 0.1   & 0.1 & 0.1 & 0.1 \\
Gradient clip     & 1.0   & 1.0 & 1.0 & 1.0 \\
Precision         & bf16  & bf16 & bf16 & bf16 \\
\bottomrule
\end{tabular}
\vspace{2pt}
\caption{Architecture and training hyperparameters per task.}
\label{tab:arch_hparams}
\end{table}

\subsection{Compute Resources}
For the Sudoku experiments, we use an NVIDIA RTX 6000 Ada, and for LM1B we use 2$\times$ NVIDIA H200 (4$\times$ for the self-conditioned runs) for approximately 2 days. TinyGSM models train on a single node with 4$\times$ NVIDIA A100 for roughly two days each, and OpenWebText models on a single node with 4$\times$ NVIDIA H200 for roughly ten days each.
\subsection{Conditional paths}\label{sec:paths_app}

\paragraph{vMF path.}
Embeddings are constrained to the unit sphere by normalization. The terminal concentration $\kappa_{\max}$ is set per task. We use $\kappa_{\max} = 15$ for Sudoku and $\kappa_{\max} = 250$ for LM1B, TinyGSM, and OpenWebText. We precompute two 2D tables on a uniform grid $(\mu_i, \kappa_j) \in [-1,1] \times [0, \kappa_{\max}]$, one for the velocity scalar $\tilde\psi(\mu, \kappa)$ and one for the radial CDF used in sampling. Both are queried at runtime by bilinear interpolation, so the per-step cost of a vMF velocity evaluation matches that of a Euclidean path up to the $L \times N$ posterior sum that all paths share.

\textit{Velocity lookup.} The function $\tilde\psi(\mu, \kappa)$ in \eqref{eq:psi_tilde} is schedule-independent and depends only on the cosine similarity $\mu = \langle w, x\rangle$ and the concentration $\kappa$. The flux representation
\begin{equation*}
\tilde\psi(\mu, \kappa) 
= -\frac{\int_{-1}^{\mu} (y - A_d(\kappa))\, 
f(y;\kappa)\, \mathrm{d} y}
{f(\mu;\kappa)\,(1-\mu^2)}
\end{equation*}
is evaluated by a cumulative trapezoid sum along the $\mu$-axis. The unnormalized radial density $f(\mu;\kappa) = (1{-}\mu^2)^{(d-3)/2}\exp(\kappa\mu)$ varies over many orders of magnitude for large $d$ or $\kappa$. We avoid underflow by working with $g(\mu) = \exp(\log f(\mu) - \max_\mu \log f)$, where the shift cancels in the ratio. Near $\mu \to \pm 1$ where $g$ underflows, the analytical boundary values from Lemma~\ref{prop:psi_wellposed} are blended smoothly with the interior solution. The Bessel ratio $A_d(\kappa) = \mathcal I_{d/2}(\kappa)/\mathcal I_{d/2-1}(\kappa)$ is evaluated by a backward continued fraction.

\textit{Training-time sampling.} Each position requires a sample $x^l \sim \mathrm{vMF}(w^l, \kappa_t)$. By radial symmetry, every sample decomposes as $x = \mu\, w + \sqrt{1{-}\mu^2}\, v$, where $v$ is a random unit vector in the tangent plane at $w$. The tangent component $v$ is obtained by projecting a standard Gaussian onto the tangent plane and normalising. The cosine $\mu$ is sampled by inverse CDF: we precompute the CDF of $f(\mu;\kappa)$ on the same 2D grid via cumulative integration, draw $u \sim \mathcal U(0,1)$, and invert by interpolation at runtime.

\paragraph{Geodesic path.}
Embeddings are constrained to the unit sphere. The conditional path is the slerp interpolant of Remark \ref{rem:geo_path} between a uniformly sampled point and the target embedding, indexed by $t \in [0, 1]$. There is no closed-form Riemannian score for this path, so the predictor-corrector sampler is unavailable.

\paragraph{VP path.}
Embeddings are unconstrained vectors in $\mathbb R^d$. The conditional path is $h_t = (1 - t) Z + t w$ with $Z \sim \mathcal N(0, I_d)$, indexed by $t \in [0, 1]$. The marginal score \eqref{score} diverges as $t \to 1$. At inference we cap the predictor and corrector below $t = 1 - 10^{-3}$ to avoid division by zero.

\paragraph{VE path.}
Following \cite{dieleman2022continuous} embeddings are vectors in $\mathbb R^d$ rescaled to norm $\sqrt{d}$ so that the per-coordinate variance of clean embeddings matches the additive Gaussian noise. The conditional path is $h_t = w + \sigma_t Z$ with $Z \sim \mathcal N(0, I_d)$ and $\sigma_t \in [0, \sigma_{\max}]$. We use $\sigma_{\max} = 300$ for Sudoku, LM1B, and TinyGSM.
 Following \cite{karras,dieleman2022continuous}, we apply input preconditioning, the network sees $h_t / \sqrt{\sigma_t^2 + 1}$ rather than $h_t$ directly. Combined with the $\sqrt{d}$ embedding scale, this gives the network input a per-coordinate variance independent of $\sigma_t$. The marginal score diverges as $\sigma_t \to 0$, so at inference we floor $\sigma_t$ at $10^{-3}$. 

\paragraph{Masked diffusion baseline.}
Tokens are corrupted by an absorbing-state continuous-time Markov chain at rate $t \in [0, 1]$, following \cite{sahoo2024simple}. The model predicts the clean token from a partially masked sequence and is trained with a cross-entropy loss restricted to masked positions. The mask schedule is fixed (not learned) as $m(t)=1-(1-t)^p$ for a power $p>0$, where $m(t)$ is the probability that a position is masked at corruption time $t$. Thus $p$ controls the curvature of the corruption schedule: $p=1$ is linear, $p<1$ masks fewer positions at intermediate times, and $p>1$ masks more positions at intermediate times. On Sudoku we report $p=1$ and $p=0.5$. The optimal posterior is time-independent \cite{gat2024discrete}, so we do not condition the backbone on time for the masked baseline.

\subsection{Adaptive Schedule}\label{sec:schedule_app}

We give details of the adaptive schedule introduced in Remark~\ref{rem:schedule}. The loss \eqref{eq:ce_loss} expects $t \sim \mathcal U(0, 1)$. In practice we sample $t$ from a learned distribution $F$ on $[0, 1]$ that concentrates mass where the per-sample cross-entropy varies fastest along $t$. The same scheme is used for every conditional path.

\paragraph{Parametrization.} The warp $\tilde F : [0, 1] \to \mathbb R_{\geq 0}$ is piecewise linear with $N$ bins, parametrized by two sets of logits $\ell^{(\mathrm{in})}, \ell^{(\mathrm{out})} \in \mathbb R^N$, both initialized to $-\log N$ so that the warp starts as the identity. Input bin widths are $\mathrm{softmax}(\ell^{(\mathrm{in})})$. Output heights come from $\mathrm{softmax}(\ell^{(\mathrm{out})})$ for the normalized CDF $F$ and from $\exp(\ell^{(\mathrm{out})})$ for the unnormalized $\tilde F$. Values $\tilde F(t)$ and $F(t)$ are obtained by linear interpolation within the bin containing $t$.

\paragraph{Fitting.} Let
\begin{equation*}
\ell(\mathbf w, \mathbf x_t) = -\frac{1}{L}\sum_{l=1}^L \log p_{\theta,t}^l(w^l \mid \hat x_t^l)
\end{equation*}
be the per-sample cross-entropy at the current $t$. We fit $\tilde F$ by minimizing
\begin{equation}
\mathcal L_{\mathrm{warp}}
= \mathbb E_{t}\Big[\big(F(t) \cdot \tilde F(1) - \ell(\mathbf w, \mathbf x_t)\big)^2\Big],
\end{equation}
with $F = \tilde F / \tilde F(1)$, so that $F$ approximates the cumulative loss curve normalized to $[0, 1]$. A more flexible variant raises $F$ to a power $\beta > 0$ before regression, which sharpens or flattens the resulting CDF. We use the linear case throughout.

\paragraph{Sampling.} Given the fitted $F$, we draw $t$ by inverse-transform sampling: $u \sim \mathcal U(0, 1)$ and $t = F^{-1}(u)$. Inversion is exact and amounts to evaluating the same piecewise-linear interpolator with input and output edge sequences swapped.

\section{Extended Experimental Results}

\subsection{Inference Ablation}\label{sec:inference_ablation}

This section documents the predictor-corrector sweep behind every PC cell of the main-text and appendix tables. Section~\ref{sec:sweep_choices} defines the common protocol and the per-dataset grids; Sections~\ref{sec:sudoku_grids} (Sudoku), \ref{sec:lm1b_grids} (LM1B), \ref{sec:gsm8k_nfe_app} (TinyGSM), and~\ref{sec:owt_nfe} (OpenWebText) report the per-dataset details and configuration tables.

\subsubsection{Sweep Choices}\label{sec:sweep_choices}

The predictor-corrector sampler of Algorithm~\ref{alg:ode} has three components that we sweep uniformly across all paths admitting a closed-form score.

\paragraph{NFE split.} Each predictor step is followed by $k$ Langevin corrector steps at corrector interval $1$, so $\mathrm{NFE} = n(1 + k)$, and the integration window is $(0, 1]$. At fixed $\mathrm{NFE}$ we sweep the splits $(n, k)$ with $n(1 + k) = \mathrm{NFE}$, e.g.\ $(n, k) \in \{(64, 1), (32, 3), (16, 7)\}$ at $\mathrm{NFE} = 128$.

\paragraph{Predictor spacing.} By default the $n$ predictor steps are placed uniformly in training time $t$. The warp-aware variant places them uniformly in the warp coordinate $F(t) \in [0, 1]$ instead, so that steps concentrate where the loss curve is steepest. The flag $\mathrm{w}$ marks warp-aware spacing, its absence marks uniform spacing.

\paragraph{Corrector step size.} The corrector uses step size $\varepsilon$ on a per-dataset grid, identical for all methods (paragraphs below). With damping the effective step size is $\varepsilon_{\mathrm{eff}} = \varepsilon (1 - u)^2$, where $u \in [0, 1]$ is a path-specific progress variable running from $0$ at the noise end to $1$ at the clean end. We set $u = \kappa_t / \kappa_{\max}$ for vMF, $u = t$ for VP, and $u = 1 - \sigma_t / \sigma_{\max}$ for VE, so that the arrow of progress is unified across paths. Damping suppresses Langevin correction near the clean end, where Euclidean scores can be singular. Empirically, damping is required for VP and VE to PC-sample at all. Without it, the corrector collapses across the entire $\varepsilon$ grid (Appendix~\ref{sec:sudoku_grids}). On vMF the two formulas are within roughly one point of each other. The flag $\mathrm{d}$ marks damping on, its absence marks damping off. The figures draw a denser $\varepsilon$ sweep, the table entries select over the grids of Table~\ref{tab:sweep_grids} only. Each grid was chosen wide enough to saturate at both ends, which we verified by sweeping past them.

A configuration is specified by a tuple $((n, k),\, \varepsilon,\, \texttt{flags})$, where $\texttt{flags} \in \{\mathrm{w},\, \mathrm{d},\, \mathrm{wd},\, \mathrm{-}\}$ encodes the two  choices. Here $\mathrm{w}$ stands for warp-aware spacing alone, $\mathrm{d}$ for damping alone, $\mathrm{wd}$ for both, and $\mathrm{-}$ for neither. Geodesic and masked diffusion are sampled with the ODE only. Entries report the best configuration over this sweep: lowest gen.\ PPL subject to the entropy floor for the text datasets, highest accuracy for GSM8K.

Table~\ref{tab:sweep_grids} gives the per-dataset grids. Sudoku validity is measured on
$1{,}000$ held-out puzzles, LM1B and OpenWebText on $1{,}024$ unconditional samples scored by
GPT-2-large, and TinyGSM on the full GSM8K test set ($1{,}319$ problems) at temperature $1.0$
with a $5$s execution timeout. Per-dataset results are in
Sections~\ref{sec:sudoku_grids}--\ref{sec:sde_results}.

\begin{table}[t]
\centering
\small
\setlength{\tabcolsep}{5pt}
\begin{tabular}{lll}
\toprule
Dataset & Corrector step size $\varepsilon$ & Further axis \\
\midrule
Sudoku      & $10^{-3}, 10^{-2}, 10^{-1}, 1, 2$ & --- \\
LM1B        & $10^{-4}, 3 \cdot 10^{-4}, 10^{-3}, 3 \cdot 10^{-3}, 10^{-2}$ & --- \\
TinyGSM     & $10^{-3}, 10^{-2}, 10^{-1}$ & budget ladder $\mathrm{NFE} \in \{1, 2, \ldots, 1024\}$ \\
OpenWebText & $10^{-5}, 10^{-4}, 3 \cdot 10^{-4}, 10^{-3}, 10^{-2}, 10^{-1}$ & onset $\tau \in \{0, 1/3, 1/2\}$ \\
\addlinespace
SDE         & $\sigma \in \{0.003, \ldots, 3\}$, $q \in \{1, 2\}$ & --- \\
\bottomrule
\end{tabular}
\vspace{2pt}
\caption{Per-dataset sweep grids. Every dataset additionally sweeps the splits, both spacings
and both damping states. The SDE row replaces the corrector step size by the diffusion scale
$\sigma$ and schedule exponent $q$ of $\sigma_t = \sigma (1 - u)^q$, and takes one
Euler--Maruyama step per network evaluation.}
\label{tab:sweep_grids}
\end{table}

\subsubsection{Sudoku}\label{sec:sudoku_grids}

We report two views of the sweep here: the per-method validity grids in Tables~\ref{tab:sudoku_grid_vmf}, \ref{tab:sudoku_grid_vp}, and~\ref{tab:sudoku_grid_ve}, and the configuration achieving each PC entry of Table~\ref{tab:sudoku_main} in Table~\ref{tab:sudoku-configs}. The corresponding grids for the two inference-time masked samplers are Tables~\ref{tab:sudoku_remdm} and~\ref{tab:sudoku_conf}.

The PC sweep on Sudoku uses the per-task corrector step size $\varepsilon \in \{10^{-3}, 10^{-2}, 10^{-1}, 1, 2\}$ at $\mathrm{NFE} = 128$, with $\varepsilon = 2$ included as a saturation check. Tables~\ref{tab:sudoku_grid_vmf}--\ref{tab:sudoku_grid_ve} report the validity at each $((n, k), \varepsilon)$, taking the best across the warp and damping flags per cell. Table~\ref{tab:sudoku-configs} lists the configuration achieving each PC entry of Table~\ref{tab:sudoku_main}.

\begin{table}[t]
\centering
\small
\setlength{\tabcolsep}{5pt}
\begin{tabular}{lcc}
\toprule
Method & ODE (\%) & PC (\%) \\
\midrule
\multicolumn{3}{l}{\textit{Non-time-conditioned}} \\
vMF      & ${56.4}$ & $\mathbf{78.2}$ \\
Geodesic & $52.2$ & -- \\
VP       & $54.8$ & $73.4$ \\
VE       & $53.4$ & $52.3$ \\
\bottomrule
\end{tabular}
\vspace{2pt}
\caption{Sudoku validity ($\uparrow$) for the non-time-conditioned models, companion to Table~\ref{tab:sudoku_main}. PC reports the best configuration over the full sweep grid (Appendix~\ref{sec:sudoku_grids}).}
\label{tab:sudoku_nontc}
\end{table}

\begin{table}[H]
\centering
\small
\begin{minipage}{0.46\textwidth}
\centering
\setlength{\tabcolsep}{5pt}
\begin{tabular}{lcl}
\toprule
Method & Validity & Configuration \\
\midrule
\multicolumn{3}{l}{\textit{Time-conditioned}} \\
vMF      & $0.747$ & $(32, 3)$, $\varepsilon = 1$,       $\mathrm{d}$  \\
VP       & $0.678$ & $(16, 7)$, $\varepsilon = 1$,       $\mathrm{d}$  \\
VE       & $0.506$ & $(16, 7)$, $\varepsilon = 10^{-3}$, $\mathrm{wd}$ \\
\midrule
\multicolumn{3}{l}{\textit{Non-time-conditioned}} \\
vMF      & $0.792$ & $(64, 1)$, $\varepsilon = 1$,       $\mathrm{wd}$ \\
VP       & $0.736$ & $(64, 1)$, $\varepsilon = 1$,       $\mathrm{d}$  \\
VE       & $0.544$ & $(16, 7)$, $\varepsilon = 10^{-3}$, $\mathrm{d}$  \\
\bottomrule
\end{tabular}
\vspace{2pt}
\caption{Configurations behind the PC entries of Table~\ref{tab:sudoku_main}.}
\label{tab:sudoku-configs}
\end{minipage}
\hfill
\begin{minipage}{0.50\textwidth}
\centering
\begin{tabular}{l|cccc}
\toprule
$(n, k)$ \textbackslash\ $\varepsilon$ & $10^{-3}$ & $10^{-2}$ & $10^{-1}$ & $1$ \\
\midrule
\multicolumn{5}{l}{\textit{Time-conditioned}} \\
$(64, 1)$ & $0.681$ & $0.712$ & $0.741$ & $0.735$ \\
$(32, 3)$ & $\mathbf{0.693}$ & $0.730$ & $0.745$ & $0.747$ \\
$(16, 7)$ & $0.688$ & $\mathbf{0.732}$ & $0.737$ & $0.658$ \\
\midrule
\multicolumn{5}{l}{\textit{Non-time-conditioned}} \\
$(64, 1)$ & $0.583$ & $0.650$ & $\mathbf{0.755}$ & $\mathbf{0.792}$ \\
$(32, 3)$ & $0.589$ & $0.650$ & $0.689$ & $0.668$ \\
$(16, 7)$ & $0.593$ & $0.660$ & $0.668$ & $0.570$ \\
\bottomrule
\end{tabular}
\vspace{2pt}
\caption{Sudoku PC grid for vMF. ODE baselines: tc $0.658$ (uniform) / $0.626$ (warp-aware). Non-tc $0.564$ (uniform) / $0.556$ (warp-aware).}
\label{tab:sudoku_grid_vmf}
\end{minipage}
\end{table}

\begin{table}[H]
\centering
\small
\setlength{\tabcolsep}{5pt}
\setlength{\tabcolsep}{5pt}
\begin{minipage}{0.48\textwidth}
\centering
\begin{tabular}{l|cccc}
\toprule
$(n, k)$ \textbackslash\ $\varepsilon$ & $10^{-3}$ & $10^{-2}$ & $10^{-1}$ & $1$ \\
\midrule
\multicolumn{5}{l}{\textit{Time-conditioned}} \\
$(64, 1)$ & $0.531$ & $0.543$ & $0.601$ & $0.658$ \\
$(32, 3)$ & $0.528$ & $0.546$ & $0.616$ & $0.668$ \\
$(16, 7)$ & $0.529$ & $0.547$ & $0.633$ & $0.678$ \\
\midrule
\multicolumn{5}{l}{\textit{Non-time-conditioned}} \\
$(64, 1)$ & $0.542$ & $0.560$ & $0.610$ & $\mathbf{0.736}$ \\
$(32, 3)$ & $0.543$ & $0.561$ & $0.637$ & $0.731$ \\
$(16, 7)$ & $\mathbf{0.551}$ & $\mathbf{0.562}$ & $\mathbf{0.645}$ & $0.733$ \\
\bottomrule
\end{tabular}
\vspace{2pt}
\caption{Sudoku PC grid for VP. ODE baselines: tc $0.534$ (uniform) / $0.526$ (warp-aware). Non-tc $0.548$ (uniform) / $0.531$ (warp-aware).}
\label{tab:sudoku_grid_vp}
\end{minipage}
\hfill
\begin{minipage}{0.48\textwidth}
\centering
\begin{tabular}{l|cccc}
\toprule
$(n, k)$ \textbackslash\ $\varepsilon$ & $10^{-3}$ & $10^{-2}$ & $10^{-1}$ & $1$ \\
\midrule
\multicolumn{5}{l}{\textit{Time-conditioned}} \\
$(64, 1)$ & $0.484$           & $0.419$ & $0.375$ & $0.422$ \\
$(32, 3)$ & $0.495$           & $0.277$ & $0.059$ & $0.022$ \\
$(16, 7)$ & $0.506$  & $0.002$ & $0.078$ & $0.000$ \\
\midrule
\multicolumn{5}{l}{\textit{Non-time-conditioned}} \\
$(64, 1)$ & $0.528$           & $\mathbf{0.499}$ & $\mathbf{0.471}$ & $\mathbf{0.490}$ \\
$(32, 3)$ & $0.536$           & $0.336$ & $0.018$ & $0.019$ \\
$(16, 7)$ & $\mathbf{0.544}$  & $0.007$ & $0.028$ & $0.000$ \\
\bottomrule
\end{tabular}
\vspace{2pt}
\caption{Sudoku PC grid for VE. ODE baselines: tc $0.464$ (uniform) / $0.480$ (warp-aware). Non-tc $0.518$ (uniform) / $0.534$ (warp-aware).}
\label{tab:sudoku_grid_ve}
\end{minipage}
\end{table}

\paragraph{Masked baseline.} The two inference-time samplers of Section~\ref{sec:exp_sudoku} are swept on the same checkpoints, $1{,}000$ samples and seed $1234$ throughout. Table~\ref{tab:sudoku_remdm} gives the ReMDM grid and Table~\ref{tab:sudoku_conf} the confidence-based grid. The two are not comparable cell by cell: ReMDM spends one forward pass per step and is fixed at $\mathrm{NFE}=128$, whereas the confidence sampler halts once nothing is masked and its $\mathrm{NFE}$ is therefore adaptive and reported per cell.

\begin{table}[t]
\centering
\small
\setlength{\tabcolsep}{5pt}
\begin{tabular}{llcc}
\toprule
$\sigma$ rule & $\eta$ & $p = 0.5$ & $p = 1$ \\
\midrule
none (ancestral) & --     & $0.350$          & $0.224$ \\
\midrule
cap     & $0.02$ & $0.374$          & $0.250$ \\
cap     & $0.05$ & $0.427$          & $0.267$ \\
cap     & $0.10$ & $0.468$          & $0.294$ \\
cap     & $0.20$ & $0.501$          & $0.316$ \\
cap     & $0.40$ & $0.514$          & $0.335$ \\
cap     & $0.60$ & $0.538$          & $0.340$ \\
cap     & $0.80$ & $0.558$          & $0.372$ \\
cap     & $1.00$ & $0.567$          & $0.372$ \\
\midrule
rescale & $0.10$ & $0.467$          & $0.289$ \\
rescale & $0.30$ & $0.503$          & $0.323$ \\
rescale & $0.50$ & $0.543$          & $0.356$ \\
rescale & $0.70$ & $0.550$          & $0.344$ \\
rescale & $0.90$ & $0.557$          & $0.377$ \\
rescale & $1.00$ & $0.567$          & $0.363$ \\
\bottomrule
\end{tabular}
\vspace{2pt}
\caption{Sudoku validity ($\uparrow$) under ReMDM remasking, $\mathrm{NFE} = 128$. The two rules coincide at $\eta = 1$; a repeat of that cell read $0.388$ for $p = 1$, the value in Table~\ref{tab:sudoku_main}.}
\label{tab:sudoku_remdm}
\end{table}

\begin{table}[t]
\centering
\small
\setlength{\tabcolsep}{5pt}
\begin{tabular}{l|ccccc}
\toprule
$\tau$ \textbackslash\ $S$ & $9$ & $16$ & $27$ & $41$ & $81$ \\
\midrule
\multicolumn{6}{l}{\textit{$p = 0.5$}} \\
$0.90$ & $0.487$~(6) & $0.517$~(8)  & $0.525$~(11) & $0.523$~(12) & $0.522$~(19) \\
$0.95$ & $0.575$~(6) & $0.611$~(8)  & $0.621$~(12) & $0.627$~(14) & $0.635$~(23) \\
$0.99$ & $0.639$~(7) & $0.693$~(10) & $0.713$~(15) & $0.727$~(16) & $0.734$~(34) \\
$1.01$ & $\mathbf{0.648}$~(8) & $\mathbf{0.712}$~(13) & $\mathbf{0.739}$~(22) & $\mathbf{0.751}$~(32) & $\mathbf{0.755}$~(64) \\
\midrule
\multicolumn{6}{l}{\textit{$p = 1$}} \\
$0.90$ & $0.417$~(6) & $0.430$~(7)  & $0.443$~(10) & $0.441$~(12) & $0.446$~(20) \\
$0.95$ & $0.487$~(6) & $0.515$~(9)  & $0.544$~(11) & $0.546$~(14) & $0.555$~(26) \\
$0.99$ & $0.503$~(7) & $0.565$~(10) & $0.599$~(13) & $0.612$~(19) & $0.634$~(27) \\
$1.01$ & $0.498$~(8) & $0.572$~(13) & $0.609$~(22) & $0.614$~(32) & $0.642$~(64) \\
\bottomrule
\end{tabular}
\vspace{2pt}
\caption{Sudoku validity ($\uparrow$) under the confidence-based sampler, with the measured $\mathrm{NFE}$ in parentheses. $\tau$ is the confidence threshold; $S$ is a budget, not a cost, since decoding halts once nothing is masked.}
\label{tab:sudoku_conf}
\end{table}

\subsubsection{LM1B}\label{sec:lm1b_grids}

The LM1B sweep follows the grids of Table~\ref{tab:sweep_grids}. Figure~\ref{fig:ppl_vs_entropy} draws the $(64, 1)$ split with warp-aware spacing and damping. The non-time-conditioned models use the wider grid $\varepsilon \in \{10^{-5}, \ldots, 10^{-1}\}$, extended to $\varepsilon = 0.3$ for VP. Table~\ref{tab:lm1b_nontc} reports their results, Table~\ref{tab:lm1b_nfe} the budget ablation, and Table~\ref{tab:lm1b-configs} the configuration behind each predictor-corrector entry.

\paragraph{Document-separator alignment with reference dataloaders.} Our data preparation script uses \texttt{[SEP]} (id $102$) as the document separator, while the reference dataloaders of \cite{sahoo2024simple, chen2026langflowcontinuousdiffusionrivals} derive the marker via \texttt{tokenizer.encode()[0]} and use \texttt{[CLS]} (id $101$). The two tokens re-tokenize to different GPT-2 BPE sequences and would otherwise bias the perplexity comparison. We remap $102 \to 101$ at decode time before scoring.

\begin{table}[htbp]
\centering
\small
\setlength{\tabcolsep}{5pt}
\begin{tabular}{llc}
\toprule
Method & Sampler & gen.\ PPL ($\downarrow$) \eH{$H$ ($\uparrow$)}\\
\midrule
\multicolumn{3}{l}{\textit{Non-time-conditioned}} \\
\multirow{3}{*}{vMF}
 & ODE               & $205.5$\eH{4.38} \\
 & PC$_{H \geq 4.30}$ & $104.8$\eH{4.31} \\
 & PC$_{H \geq 4.15}$ & $77.2$\eH{4.28} \\
 \addlinespace
Geodesic           & ODE & $215.5$\eH{4.36} \\
\addlinespace
\multirow{5}{*}{VP}
 & ODE               & $138.8$\eH{4.34} \\
 & PC$_{H \geq 4.30}$ & $109.4$\eH{4.31} \\
 & PC$_{H \geq 4.25}$ & $88.3$\eH{4.26} \\
 & PC$_{H \geq 4.20}$ & $77.0$\eH{4.20} \\
 & PC$_{H \geq 4.15}$ & $70.7$\eH{4.15} \\
\addlinespace
\multirow{2}{*}{VE}
 & ODE               & $195.1$\eH{4.40} \\
 &$\text{PC}_{H \geq 4.15}$               & $176.7$\eH{4.29} \\
\bottomrule
\end{tabular}
\vspace{2pt}
\caption{LM1B comparison at $\mathrm{NFE} = 128$ for the non-time-conditioned models, companion to Table~\ref{tab:lm1b-comparison} and Figure~\ref{fig:ppl_vs_entropy}. ODE is the Euler sampler at $n = 128$; PC$_{H \geq h}$ is the best PC configuration with entropy at or above threshold.}
\label{tab:lm1b_nontc}
\end{table}

\begin{table}[t]
\centering
\small
\setlength{\tabcolsep}{5pt}
\begin{tabular}{llcl}
\toprule
Method & Sampler & gen.\ PPL ($\downarrow$) \eH{$H$ ($\uparrow$)} & Configuration \\
\midrule
\multicolumn{4}{l}{\textit{Time-conditioned}} \\
\multirow{6}{*}{vMF}
 & ODE                  & $171.8$\eH{4.35} & warp-aware, $n = 128$ \\
 & PC, $H \geq 4.34$    & $165.4$\eH{4.36} & $(32, 3)$, $\varepsilon = 3 \cdot 10^{-4}$, $\mathrm{wd}$ \\
 & PC, $H \geq 4.30$    & $112.7$\eH{4.30} & $(64, 1)$, $\varepsilon = 10^{-3}$, $\mathrm{wd}$ \\
 & PC, $H \geq 4.25$    & $66.0$\eH{4.25}  & $(32, 3)$, $\varepsilon = 10^{-3}$, $\mathrm{-}$  \\
 & PC, $H \geq 4.20$    & $52.4$\eH{4.21}  & $(32, 3)$, $\varepsilon = 10^{-3}$, $\mathrm{w}$  \\
 & PC, $H \geq 4.15$    & $45.1$\eH{4.16}  & $(64, 1)$, $\varepsilon = 3 \cdot 10^{-3}$, $\mathrm{wd}$ \\
\addlinespace
Geodesic                & ODE     & $167.9$\eH{4.34} & warp-aware, $n = 128$ \\
\addlinespace
\multirow{2}{*}{VP}
 & ODE                  & $140.4$\eH{4.34} & warp-aware, $n = 128$ \\
 & PC, $H \geq 4.15$    & $138.4$\eH{4.35} & $(64, 1)$, $\varepsilon = 10^{-2}$, $\mathrm{wd}$ \\
\addlinespace
\multirow{2}{*}{VE}
 & ODE                  & $149.7$\eH{4.35} & warp-aware, $n = 128$ \\
 & PC, $H \geq 4.15$    & $151.7$\eH{4.35} & $(64, 1)$, $\varepsilon = 10^{-4}$, $\mathrm{wd}$ \\
\midrule
\multicolumn{4}{l}{\textit{+ self-conditioning ($p = 0.25$)}} \\
\multirow{6}{*}{vMF}
 & ODE                  & $92.4$\eH{4.34} & warp-aware, $n = 128$ \\
 & PC, $H \geq 4.34$    & $77.5$\eH{4.34} & $(16, 7)$, $\varepsilon = 3 \cdot 10^{-4}$, $\mathrm{wd}$ \\
 & PC, $H \geq 4.30$    & $55.6$\eH{4.30} & $(32, 3)$, $\varepsilon = 10^{-3}$, $\mathrm{wd}$ \\
 & PC, $H \geq 4.25$    & $38.4$\eH{4.25} & $(32, 3)$, $\varepsilon = 10^{-3}$, $\mathrm{w}$  \\
 & PC, $H \geq 4.20$    & $33.8$\eH{4.21} & $(64, 1)$, $\varepsilon = 3 \cdot 10^{-3}$, $\mathrm{wd}$ \\
 & PC, $H \geq 4.15$    & $27.0$\eH{4.17} & $(64, 1)$, $\varepsilon = 3 \cdot 10^{-3}$, $\mathrm{w}$ \\
\addlinespace
Geodesic                & ODE     & $107.1$\eH{4.33} & warp-aware, $n = 128$ \\
\addlinespace
\multirow{3}{*}{VP}
 & ODE                  & $92.3$\eH{4.33} & warp-aware, $n = 128$ \\
 & PC, $H \geq 4.34$    & $88.9$\eH{4.34} & $(16, 7)$, $\varepsilon = 3 \cdot 10^{-3}$, $\mathrm{wd}$ \\
 & PC, $H \geq 4.15$    & $83.1$\eH{4.33} & $(16, 7)$, $\varepsilon = 10^{-2}$, $\mathrm{wd}$ \\
\addlinespace
\multirow{2}{*}{VE}
 & ODE                  & $163.4$\eH{4.36} & warp-aware, $n = 128$ \\
 & PC, $H \geq 4.15$    & $164.6$\eH{4.36} & $(64, 1)$, $\varepsilon = 10^{-3}$, $\mathrm{wd}$ \\
\midrule
\multicolumn{4}{l}{\textit{Non-time-conditioned}} \\
\multirow{3}{*}{vMF}
 & ODE                  & $205.5$\eH{4.38} & warp-aware, $n = 128$ \\
 & PC, $H \geq 4.30$    & $104.8$\eH{4.31} & $(16, 7)$, $\varepsilon = 10^{-2}$, $\mathrm{d}$  \\
 & PC, $H \geq 4.15$    & $77.2$\eH{4.28}  & $(16, 7)$, $\varepsilon = 10^{-2}$, $\mathrm{wd}$ \\
\addlinespace
Geodesic                & ODE     & $215.5$\eH{4.36} & warp-aware, $n = 128$ \\
\addlinespace
\multirow{5}{*}{VP}
 & ODE                  & $138.8$\eH{4.34} & warp-aware, $n = 128$ \\
 & PC, $H \geq 4.30$    & $109.4$\eH{4.31} & $(64, 1)$, $\varepsilon = 10^{-1}$, $\mathrm{wd}$ \\
 & PC, $H \geq 4.25$    & $88.3$\eH{4.26}  & $(64, 1)$, $\varepsilon = 0.3$,    $\mathrm{wd}$ \\
 & PC, $H \geq 4.20$    & $77.0$\eH{4.20}  & $(32, 3)$, $\varepsilon = 0.3$,    $\mathrm{wd}$ \\
 & PC, $H \geq 4.15$    & $70.7$\eH{4.15}  & $(16, 7)$, $\varepsilon = 0.3$,    $\mathrm{wd}$ \\
\addlinespace
\multirow{3}{*}{VE}
 & ODE                  & $195.1$\eH{4.40} & warp-aware, $n = 128$ \\
 & PC, $H \geq 4.30$    & $190.3$\eH{4.37} & $(32, 3)$, $\varepsilon = 10^{-5}$, $\mathrm{d}$ \\
 & PC, $H \geq 4.15$    & $176.7$\eH{4.29} & $(16, 7)$, $\varepsilon = 10^{-3}$, $\mathrm{d}$ \\
\bottomrule
\end{tabular}
\vspace{2pt}
\caption{Configurations behind every (method, sampler, entropy floor) cell of Tables~\ref{tab:lm1b-comparison} and~\ref{tab:lm1b_pc} and Figure~\ref{fig:ppl_vs_entropy}. Methods that plateau carry one PC row, at the lowest floor they reach.}
\label{tab:lm1b-configs}
\end{table}

\begin{table}[htbp]
\centering
\small
\setlength{\tabcolsep}{5pt}
\begin{tabular}{lcccc}
\toprule
& \multicolumn{2}{c}{vMF} & \multicolumn{2}{c}{vMF +SC} \\
\cmidrule(lr){2-3} \cmidrule(lr){4-5}
$\mathrm{NFE}$ & ODE & PC & ODE & PC \\
\midrule
$8$   & $243.7$\eH{4.40} & $132.7$\eH{4.31} & $234.5$\eH{4.40} & $163.2$\eH{4.32} \\
$16$  & $203.7$\eH{4.37} & $190.7$\eH{4.36} & $154.9$\eH{4.38} & $95.4$\eH{4.30} \\
$32$  & $185.0$\eH{4.36} & $150.5$\eH{4.34} & $119.3$\eH{4.36} & $86.2$\eH{4.30}  \\
$64$  & $176.3$\eH{4.35} & $123.1$\eH{4.32} & $101.7$\eH{4.34} & $76.7$\eH{4.32}  \\
$128$ & $171.8$\eH{4.35} & $112.7$\eH{4.30} & $92.4$\eH{4.34}  & $55.6$\eH{4.30}  \\
\bottomrule
\end{tabular}
\vspace{2pt}
\caption{Sampling-budget ablation on LM1B for time-conditioned vMF, without and with self-conditioning; cells report gen.\ PPL ($H$). PC is the best configuration subject to $H \geq 4.30$ (Appendix~\ref{sec:sweep_choices}).}
\label{tab:lm1b_nfe}
\end{table}

\subsubsection{TinyGSM}\label{sec:gsm8k_nfe_app} 

The TinyGSM sweep follows the grids of Table~\ref{tab:sweep_grids}. Table~\ref{tab:gsm8k_nfe} reports the accuracy at every budget, and Table~\ref{tab:gsm8k-configs} lists the configuration behind each predictor-corrector entry.

\begin{table}[htbp]
\centering
\small
\setlength{\tabcolsep}{5pt}
\begin{tabular}{llcl}
\toprule
Method & Cell & Acc.\ (\%) & Configuration \\
\midrule
\multicolumn{4}{l}{\textit{Table~\ref{tab:gsm8k_main} ($\mathrm{NFE} = 1024$)}} \\
vMF & PC & $33.0$ & $(512, 1)$, $\varepsilon = 10^{-2}$, $\mathrm{w}$ \\
VP  & PC & $12.7$ & $(512, 1)$, $\varepsilon = 10^{-1}$, $\mathrm{wd}$ \\
VE  & PC & $1.9$  & $(512, 1)$, $\varepsilon = 10^{-3}$, $\mathrm{d}$ \\
\addlinespace
\multicolumn{4}{l}{\textit{+ self-conditioning ($p = 0.25$)}} \\
vMF & PC & $43.5$ & $(512, 1)$, $\varepsilon = 10^{-2}$, $\mathrm{w}$ \\
VP  & PC & $28.5$ & $(512, 1)$, $\varepsilon = 10^{-1}$, $\mathrm{d}$ \\
VE  & PC & $4.1$  & $(512, 1)$, $\varepsilon = 10^{-1}$, $\mathrm{d}$ \\
\midrule
\multicolumn{4}{l}{\textit{Table~\ref{tab:gsm8k_nfe} ladder; the $\mathrm{NFE} = 1024$ rows repeat the vMF cells above}} \\
\multirow{9}{*}{vMF}
 & PC, $\mathrm{NFE} = 2$   & $0.0$  & $(1, 1)$, $\varepsilon = 10^{-1}$, $\mathrm{d}$ \\
 & PC, $\mathrm{NFE} = 4$   & $0.7$  & $(2, 1)$, $\varepsilon = 10^{-3}$, $\mathrm{w}$ \\
 & PC, $\mathrm{NFE} = 8$   & $6.2$  & $(4, 1)$, $\varepsilon = 10^{-2}$, $\mathrm{w}$ \\
 & PC, $\mathrm{NFE} = 16$  & $15.5$ & $(8, 1)$, $\varepsilon = 10^{-2}$, $\mathrm{w}$ \\
 & PC, $\mathrm{NFE} = 32$  & $21.8$ & $(16, 1)$, $\varepsilon = 10^{-2}$, $\mathrm{w}$ \\
 & PC, $\mathrm{NFE} = 64$  & $26.9$ & $(32, 1)$, $\varepsilon = 10^{-2}$, $\mathrm{w}$ \\
 & PC, $\mathrm{NFE} = 128$ & $27.4$ & $(64, 1)$, $\varepsilon = 10^{-2}$, $\mathrm{wd}$ \\
 & PC, $\mathrm{NFE} = 256$ & $31.8$ & $(128, 1)$, $\varepsilon = 10^{-2}$, $\mathrm{w}$ \\
 & PC, $\mathrm{NFE} = 512$ & $32.4$ & $(256, 1)$, $\varepsilon = 10^{-2}$, $\mathrm{w}$ \\
\addlinespace
\multirow{9}{*}{vMF +SC}
 & PC, $\mathrm{NFE} = 2$   & $0.0$  & $(1, 1)$, $\varepsilon = 10^{-3}$, $\mathrm{w}$ \\
 & PC, $\mathrm{NFE} = 4$   & $1.1$  & $(2, 1)$, $\varepsilon = 10^{-3}$, $\mathrm{w}$ \\
 & PC, $\mathrm{NFE} = 8$   & $9.7$  & $(4, 1)$, $\varepsilon = 10^{-2}$, $\mathrm{w}$ \\
 & PC, $\mathrm{NFE} = 16$  & $21.9$ & $(8, 1)$, $\varepsilon = 10^{-2}$, $\mathrm{w}$ \\
 & PC, $\mathrm{NFE} = 32$  & $30.9$ & $(16, 1)$, $\varepsilon = 10^{-2}$, $\mathrm{wd}$ \\
 & PC, $\mathrm{NFE} = 64$  & $35.6$ & $(32, 1)$, $\varepsilon = 10^{-2}$, $\mathrm{w}$ \\
 & PC, $\mathrm{NFE} = 128$ & $39.7$ & $(64, 1)$, $\varepsilon = 10^{-2}$, $\mathrm{w}$ \\
 & PC, $\mathrm{NFE} = 256$ & $39.4$ & $(128, 1)$, $\varepsilon = 10^{-2}$, $\mathrm{w}$ \\
 & PC, $\mathrm{NFE} = 512$ & $41.7$ & $(256, 1)$, $\varepsilon = 10^{-2}$, $\mathrm{w}$ \\
\bottomrule
\end{tabular}
\vspace{2pt}
\caption{Configurations behind the PC cells of Tables~\ref{tab:gsm8k_main} and~\ref{tab:gsm8k_nfe}. Notation as in Section~\ref{sec:sweep_choices}.}
\label{tab:gsm8k-configs}
\end{table}

\begin{table}[htbp]
\centering
\small
\setlength{\tabcolsep}{5pt}
\begin{tabular}{lcccc}
\toprule
& \multicolumn{2}{c}{vMF} & \multicolumn{2}{c}{vMF +SC} \\
\cmidrule(lr){2-3} \cmidrule(lr){4-5}
$\mathrm{NFE}$ & ODE & PC & ODE & PC \\
\midrule
$1$    & $0.0$ & --     & $0.0$  & --     \\
$2$    & $0.4$ & $0.0$  & $0.2$  & $0.0$  \\
$4$    & $2.0$ & $0.7$  & $2.7$  & $1.1$  \\
$8$    & $3.9$ & $6.2$  & $7.1$  & $9.7$  \\
$16$   & $5.8$ & $15.5$ & $11.6$ & $21.9$ \\
$32$   & $6.2$ & $21.8$ & $17.4$ & $30.9$ \\
$64$   & $6.1$ & $26.9$ & $19.9$ & $35.6$ \\
$128$  & $6.4$ & $27.4$ & $22.5$ & $39.7$ \\
$256$  & $6.7$ & $31.8$ & $23.0$ & $39.4$ \\
$512$  & $6.4$ & $32.4$ & $23.7$ & $41.7$ \\
$1024$ & $6.4$ & $33.0$ & $23.7$ & $43.5$ \\
\bottomrule
\end{tabular}
\vspace{2pt}
\caption{Sampling-budget ablation on GSM8K: test accuracy ($\uparrow$, \%) for time-conditioned vMF, without and with self-conditioning. PC is the best configuration per budget (Appendix~\ref{sec:sweep_choices}).}
\label{tab:gsm8k_nfe}
\end{table}

\subsubsection{OpenWebText}\label{sec:owt_nfe}

\begin{table}[t]
\centering
\small
\setlength{\tabcolsep}{5pt}
\begin{tabular}{lccccc}
\toprule
 & \multicolumn{5}{c}{gen.\ PPL ($\downarrow$) \eH{$H$ ($\uparrow$)}} \\
\cmidrule(lr){2-6}
Method & PC$_{H \geq 5.44}$ & PC$_{H \geq 5.40}$ & PC$_{H \geq 5.35}$ & PC$_{H \geq 5.25}$ & PC$_{H \geq 5.15}$ \\
\midrule
vMF & $97.7$\eH{5.44}  & $94.9$\eH{5.40} & $88.5$\eH{5.36} & $61.1$\eH{5.26} & $51.1$\eH{5.17} \\
VP  & $117.1$\eH{5.47} & $94.6$\eH{5.41} & $80.4$\eH{5.36} & $70.1$\eH{5.31} & $70.1$\eH{5.31} \\
\midrule
\multicolumn{6}{l}{\textit{+ self-conditioning ($p = 0.25$)}} \\
vMF & $\mathbf{55.5}$\eH{5.44} & $49.5$\eH{5.40} & $38.9$\eH{5.35} & $33.3$\eH{5.26} & $32.2$\eH{5.22} \\
VP  & $59.3$\eH{5.44} & $\mathbf{47.3}$\eH{5.41} & $\mathbf{36.2}$\eH{5.35} & $\mathbf{32.4}$\eH{5.31} & $\mathbf{21.8}$\eH{5.16} \\
\bottomrule
\end{tabular}
\vspace{2pt}
\caption{\textbf{OpenWebText} predictor-corrector (PC) at $\mathrm{NFE} = 1024$: best gen.\ PPL per entropy floor, the leading column held to $H_{\mathrm{data}} = 5.44$. Self-conditioned cells select over the onset $\tau \in \{0, 1/3, 1/2\}$. Best per column in bold. Configurations in Table~\ref{tab:owt-configs}.}
\label{tab:owt_pc}
\end{table}

The OpenWebText sweep follows the grids of Table~\ref{tab:sweep_grids}, at $\mathrm{NFE} \in \{128, 1024\}$. Figure~\ref{fig:owt_ppl_vs_entropy} draws the $(512, 1)$ split with warp-aware spacing. Tables~\ref{tab:owt_pc} and~\ref{tab:owt_pc_128} report the predictor-corrector results per entropy floor, Table~\ref{tab:owt_ode_variants} the ODE results over both spacings and all onsets, and Table~\ref{tab:owt-configs} the configurations behind the reported entries.

\begin{table}[htbp]
\centering
\small
\setlength{\tabcolsep}{5pt}
\begin{tabular}{llcll}
\toprule
Method & Cell & gen.\ PPL ($\downarrow$) \eH{$H$ ($\uparrow$)} & Configuration & Onset \\
\midrule
\multicolumn{5}{l}{\textit{$\mathrm{NFE} = 1024$}} \\
\multirow{5}{*}{vMF}
 & PC, $H \geq 5.44$ & $97.7$\eH{5.44} & $(128, 7)$, $\varepsilon = 10^{-4}$, $\mathrm{d}$ & $0$ \\
 & PC, $H \geq 5.40$ & $94.9$\eH{5.40} & $(512, 1)$, $\varepsilon = 10^{-4}$, $\mathrm{wd}$ & $0$ \\
 & PC, $H \geq 5.35$ & $88.5$\eH{5.36} & $(128, 7)$, $\varepsilon = 10^{-4}$, $\mathrm{wd}$ & $0$ \\
 & PC, $H \geq 5.25$ & $61.1$\eH{5.26} & $(512, 1)$, $\varepsilon = 3 \cdot 10^{-4}$, $\mathrm{d}$ & $0$ \\
 & PC, $H \geq 5.15$ & $51.1$\eH{5.17} & $(256, 3)$, $\varepsilon = 3 \cdot 10^{-4}$, $\mathrm{d}$ & $0$ \\
\addlinespace
\multirow{4}{*}{VP}
 & PC, $H \geq 5.44$ & $117.1$\eH{5.47} & $(128, 7)$, $\varepsilon = 10^{-3}$, $\mathrm{wd}$ & $0$ \\
 & PC, $H \geq 5.40$ & $94.6$\eH{5.41} & $(512, 1)$, $\varepsilon = 10^{-1}$, $\mathrm{d}$ & $0$ \\
 & PC, $H \geq 5.35$ & $80.4$\eH{5.36} & $(512, 1)$, $\varepsilon = 10^{-1}$, $\mathrm{wd}$ & $0$ \\
 & PC, $H \geq 5.25$ & $70.1$\eH{5.31} & $(128, 7)$, $\varepsilon = 10^{-1}$, $\mathrm{wd}$ & $0$ \\
\addlinespace
\multirow{5}{*}{vMF +SC}
 & PC, $H \geq 5.44$ & $55.5$\eH{5.44} & $(512, 1)$, $\varepsilon = 10^{-5}$, $\mathrm{wd}$ & $1/2$ \\
 & PC, $H \geq 5.40$ & $49.5$\eH{5.40} & $(256, 3)$, $\varepsilon = 10^{-5}$, $\mathrm{wd}$ & $1/2$ \\
 & PC, $H \geq 5.35$ & $38.9$\eH{5.35} & $(512, 1)$, $\varepsilon = 10^{-5}$, $\mathrm{wd}$ & $1/3$ \\
 & PC, $H \geq 5.25$ & $33.3$\eH{5.26} & $(512, 1)$, $\varepsilon = 10^{-4}$, $\mathrm{wd}$ & $1/2$ \\
 & PC, $H \geq 5.15$ & $32.2$\eH{5.22} & $(128, 7)$, $\varepsilon = 10^{-5}$, $\mathrm{d}$ & $0$ \\
\addlinespace
\multirow{5}{*}{VP +SC}
 & PC, $H \geq 5.44$ & $59.3$\eH{5.44} & $(512, 1)$, $\varepsilon = 10^{-3}$, $\mathrm{wd}$ & $1/2$ \\
 & PC, $H \geq 5.40$ & $47.3$\eH{5.41} & $(512, 1)$, $\varepsilon = 10^{-2}$, $\mathrm{wd}$ & $1/2$ \\
 & PC, $H \geq 5.35$ & $36.2$\eH{5.35} & $(512, 1)$, $\varepsilon = 10^{-2}$, $\mathrm{wd}$ & $1/3$ \\
 & PC, $H \geq 5.25$ & $32.4$\eH{5.31} & $(128, 7)$, $\varepsilon = 10^{-2}$, $\mathrm{wd}$ & $1/3$ \\
 & PC, $H \geq 5.15$ & $21.8$\eH{5.16} & $(128, 7)$, $\varepsilon = 10^{-2}$, $\mathrm{wd}$ & $0$ \\
\midrule
\multicolumn{5}{l}{\textit{$\mathrm{NFE} = 128$}} \\
\multirow{3}{*}{vMF}
 & PC, $H \geq 5.44$ & $100.4$\eH{5.44} & $(32, 3)$, $\varepsilon = 3 \cdot 10^{-4}$, $\mathrm{wd}$ & $0$ \\
 & PC, $H \geq 5.40$ & $98.3$\eH{5.42} & $(16, 7)$, $\varepsilon = 3 \cdot 10^{-4}$, $\mathrm{wd}$ & $0$ \\
 & PC, $H \geq 5.25$ & $52.8$\eH{5.25} & $(16, 7)$, $\varepsilon = 10^{-3}$, $\mathrm{d}$ & $0$ \\
\addlinespace
VP & PC, $H \geq 5.40$ & $122.5$\eH{5.50} & $(32, 3)$, $\varepsilon = 10^{-1}$, $\mathrm{wd}$ & $0$ \\
\addlinespace
\multirow{4}{*}{vMF +SC}
 & PC, $H \geq 5.40$ & $52.7$\eH{5.44} & $(32, 3)$, $\varepsilon = 10^{-4}$, $\mathrm{wd}$ & $1/3$ \\
 & PC, $H \geq 5.35$ & $42.8$\eH{5.37} & $(16, 7)$, $\varepsilon = 10^{-4}$, $\mathrm{wd}$ & $0$ \\
 & PC, $H \geq 5.25$ & $33.8$\eH{5.27} & $(16, 7)$, $\varepsilon = 3 \cdot 10^{-4}$, $\mathrm{wd}$ & $0$ \\
 & PC, $H \geq 5.15$ & $32.6$\eH{5.24} & $(32, 3)$, $\varepsilon = 3 \cdot 10^{-4}$, $\mathrm{wd}$ & $0$ \\
\addlinespace
\multirow{3}{*}{VP +SC}
 & PC, $H \geq 5.44$ & $54.1$\eH{5.44} & $(16, 7)$, $\varepsilon = 10^{-3}$, $\mathrm{wd}$ & $0$ \\
 & PC, $H \geq 5.40$ & $46.8$\eH{5.42} & $(16, 7)$, $\varepsilon = 10^{-2}$, $\mathrm{wd}$ & $0$ \\
 & PC, $H \geq 5.35$ & $45.2$\eH{5.39} & $(32, 3)$, $\varepsilon = 10^{-2}$, $\mathrm{wd}$ & $0$ \\
\bottomrule
\end{tabular}
\vspace{2pt}
\caption{Configurations behind every PC cell of Tables~\ref{tab:owt_pc} and~\ref{tab:owt_pc_128} and Figure~\ref{fig:owt_ppl_vs_entropy}. Notation as in Section~\ref{sec:sweep_choices}; the onset column gives the fraction $\tau$ of the trajectory over which the self-conditioning input is suppressed. Corrector damping is on for all OpenWebText cells.}
\label{tab:owt-configs}
\end{table}

\begin{table}[t]
\centering
\small
\setlength{\tabcolsep}{5pt}
\begin{tabular}{lccccc}
\toprule
 & \multicolumn{5}{c}{gen.\ PPL ($\downarrow$) \eH{$H$ ($\uparrow$)}} \\
\cmidrule(lr){2-6}
Method & PC$_{H \geq 5.44}$ & PC$_{H \geq 5.40}$ & PC$_{H \geq 5.35}$ & PC$_{H \geq 5.25}$ & PC$_{H \geq 5.15}$ \\
\midrule
vMF & $100.4$\eH{5.44} & $98.3$\eH{5.42}  & $98.3$\eH{5.42}  & $52.8$\eH{5.25}  & $52.8$\eH{5.25} \\
VP  & $122.5$\eH{5.50} & $122.5$\eH{5.50} & $122.5$\eH{5.50} & $122.5$\eH{5.50} & $122.5$\eH{5.50} \\
\midrule
\multicolumn{6}{l}{\textit{+ self-conditioning ($p = 0.25$)}} \\
vMF & $\mathbf{52.7}$\eH{5.44} & $52.7$\eH{5.44} & $\mathbf{42.8}$\eH{5.37} & $\mathbf{33.8}$\eH{5.27} & $\mathbf{32.6}$\eH{5.24} \\
VP  & $54.1$\eH{5.44} & $\mathbf{46.8}$\eH{5.42} & $45.2$\eH{5.39} & $45.2$\eH{5.39} & $45.2$\eH{5.39} \\
\bottomrule
\end{tabular}
\vspace{2pt}
\caption{\textbf{OpenWebText} predictor-corrector (PC) at $\mathrm{NFE} = 128$; otherwise as Table~\ref{tab:owt_pc}.}
\label{tab:owt_pc_128}
\end{table}

\begin{table}[htbp]
\centering
\small
\setlength{\tabcolsep}{5pt}
\begin{tabular}{lllcc}
\toprule
 & & & \multicolumn{2}{c}{gen.\ PPL ($\downarrow$) \eH{$H$ ($\uparrow$)}} \\
\cmidrule(lr){4-5}
Method & NFE & Onset & ODE (uniform) & ODE (warp-aware) \\
\midrule
\multicolumn{5}{l}{\textit{Ours}} \\
\multirow{2}{*}{vMF}
 & $1024$ & $0$ & $140.4$\eH{5.518} & $\mathbf{139.3}$\eH{5.518} \\
 & $128$ & $0$ & $156.8$\eH{5.549} & $\mathbf{146.9}$\eH{5.537} \\
\addlinespace
\multirow{2}{*}{Geodesic}
 & $1024$ & $0$ & $149.9$\eH{5.537} & $\mathbf{139.0}$\eH{5.514} \\
 & $128$ & $0$ & $249.8$\eH{5.684} & $\mathbf{144.4}$\eH{5.536} \\
\addlinespace
\multirow{2}{*}{VP}
 & $1024$ & $0$ & $123.5$\eH{5.455} & $\mathbf{123.0}$\eH{5.456} \\
 & $128$ & $0$ & $134.2$\eH{5.473} & $\mathbf{128.6}$\eH{5.480} \\
\midrule
\multicolumn{5}{l}{\textit{+ self-conditioning ($p = 0.25$)}} \\
\multirow{6}{*}{vMF}
 & $1024$ & $0$ & $41.6$\eH{5.244} & $\mathbf{36.0}$\eH{5.198} \\
 &        & $1/3$ & $116.1$\eH{5.533} & $55.3$\eH{5.381} \\
 &        & $1/2$ & $146.1$\eH{5.559} & $\mathbf{73.9}$\eH{5.456} \\
\cmidrule(lr){2-5}
 & $128$ & $0$ & $77.0$\eH{5.436} & $\mathbf{53.6}$\eH{5.350} \\
 &        & $1/3$ & $138.4$\eH{5.581} & $\mathbf{70.7}$\eH{5.453} \\
 &        & $1/2$ & $166.1$\eH{5.600} & $88.4$\eH{5.506} \\
\addlinespace
\multirow{6}{*}{Geodesic}
 & $1024$ & $0$ & $64.9$\eH{5.409} & $\mathbf{36.3}$\eH{5.233} \\
 &        & $1/3$ & $159.0$\eH{5.563} & $52.7$\eH{5.386} \\
 &        & $1/2$ & $159.0$\eH{5.563} & $\mathbf{70.3}$\eH{5.449} \\
\cmidrule(lr){2-5}
 & $128$ & $0$ & $258.8$\eH{5.726} & $\mathbf{55.0}$\eH{5.378} \\
 &        & $1/3$ & $275.0$\eH{5.714} & $\mathbf{67.5}$\eH{5.447} \\
 &        & $1/2$ & $275.0$\eH{5.714} & $83.1$\eH{5.491} \\
\addlinespace
\multirow{6}{*}{VP}
 & $1024$ & $0$ & $38.1$\eH{5.229} & $\mathbf{36.3}$\eH{5.224} \\
 &        & $1/3$ & $38.6$\eH{5.236} & $52.5$\eH{5.372} \\
 &        & $1/2$ & $41.6$\eH{5.278} & $\mathbf{66.2}$\eH{5.429} \\
\cmidrule(lr){2-5}
 & $128$ & $0$ & $60.9$\eH{5.361} & $\mathbf{50.1}$\eH{5.348} \\
 &        & $1/3$ & $61.3$\eH{5.355} & $63.0$\eH{5.433} \\
 &        & $1/2$ & $64.3$\eH{5.384} & $\mathbf{76.1}$\eH{5.475} \\
\bottomrule
\end{tabular}
\vspace{2pt}
\caption{Per-variant OpenWebText ODE results: uniform and warp-aware spacing at $n \in \{128, 1024\}$, across the self-conditioning onset $\tau$. Bold marks the cells reported in Table~\ref{tab:owt_main}.}
\label{tab:owt_ode_variants}
\end{table}
%---------------------------------------------------------------------
% (a) Time-conditioned
%---------------------------------------------------------------------
 
%---------------------------------------------------------------------
% (b) + self-conditioning
%---------------------------------------------------------------------
\subsubsection{SDE Sampler}\label{sec:sde_results}

\paragraph{Protocol.} We evaluate Algorithm~\ref{alg:sde_app} on the same checkpoints and under
the protocol of Section~\ref{sec:exp}: gen.\ PPL and entropy $H$ on LM1B and OpenWebText, GSM8K
accuracy after training on TinyGSM. The number of Euler--Maruyama steps equals the budget,
$S = \mathrm{NFE}$; there is no predictor-corrector split. Steps are placed uniformly in the
warp coordinate, on the grid clamped below $u = 1$ (up to $1 - 10^{-3}$), and tokens are decoded
from the posterior at the final state, so no step to $\kappa_{\max}$ is taken.

\paragraph{Sweep.} We sweep the diffusion scale
$\sigma \in \{0.003, 0.01, 0.03, 0.05, 0.1, 0.2, 0.3, 0.5, 1, 3\}$ and the schedule exponent
$q \in \{1, 2\}$ of $\sigma_t = \sigma (1 - u)^q$, with the path progress variable $u$ of
Section~\ref{sec:sweep_choices} ($u = \kappa_t / \kappa_{\max}$ for vMF, $u = t$ for VP,
$u = 1 - \sigma_t / \sigma_{\max}$ for VE). Entropy floors, sample counts and scoring follow the
per-dataset protocol of Table~\ref{tab:sweep_grids}.

\paragraph{Reporting.} All SDE runs use a single seed. Table~\ref{tab:lm1b_sde} through Table~\ref{tab:gsm8k_sde} report the results per entropy floor, and Table~\ref{tab:sde_configs} lists the winning $(\sigma, q)$ behind each entry. Geodesic admits no closed-form score and is omitted.

\begin{table}[htbp]
\centering
\small
\setlength{\tabcolsep}{5pt}
\begin{tabular}{lccccc}
\toprule
 & \multicolumn{5}{c}{gen.\ PPL ($\downarrow$) \eH{$H$ ($\uparrow$)}} \\
\cmidrule(lr){2-6}
Method & SDE$_{H \geq 4.34}$ & SDE$_{H \geq 4.30}$ & SDE$_{H \geq 4.25}$ & SDE$_{H \geq 4.20}$ & SDE$_{H \geq 4.15}$ \\
\midrule
vMF & $136.1$\eH{4.35} & $136.1$\eH{4.35} & $77.1$\eH{4.28} & $77.1$\eH{4.28} & $77.1$\eH{4.28} \\
VP & $137.3$\eH{4.34} & $137.3$\eH{4.34} & $137.3$\eH{4.34} & $137.3$\eH{4.34} & $137.3$\eH{4.34} \\
VE & $145.3$\eH{4.35} & $145.3$\eH{4.35} & $145.3$\eH{4.35} & $145.3$\eH{4.35} & $145.3$\eH{4.35} \\
\midrule
\multicolumn{6}{l}{\textit{+ self-conditioning ($p = 0.25$)}} \\
vMF & $\mathbf{70.0}$\eH{4.34} & $\mathbf{49.7}$\eH{4.30} & $\mathbf{49.7}$\eH{4.30} & $\mathbf{37.2}$\eH{4.24} & $\mathbf{26.9}$\eH{4.18} \\
VP & $118.2$\eH{4.35} & $91.8$\eH{4.32} & $91.8$\eH{4.32} & $91.8$\eH{4.32} & $91.8$\eH{4.32} \\
VE & $160.2$\eH{4.36} & $160.2$\eH{4.36} & $160.2$\eH{4.36} & $160.2$\eH{4.36} & $160.2$\eH{4.36} \\
\bottomrule
\end{tabular}
\vspace{2pt}
\caption{LM1B SDE results at $\mathrm{NFE} = 128$: best gen.\ PPL ($H$) per entropy floor; methods that plateau repeat the same value at every floor. Best per column in bold.}
\label{tab:lm1b_sde}
\end{table}

\begin{table}[htbp]
\centering
\footnotesize
\setlength{\tabcolsep}{5pt}
\begin{tabular}{llccccc}
\toprule
 & & \multicolumn{5}{c}{gen.\ PPL ($\downarrow$) \eH{$H$ ($\uparrow$)}} \\
\cmidrule(lr){3-7}
Method & NFE & SDE$_{H \geq 5.44}$ & SDE$_{H \geq 5.40}$ & SDE$_{H \geq 5.35}$ & SDE$_{H \geq 5.25}$ & SDE$_{H \geq 5.15}$ \\
\midrule
\multirow{2}{*}{vMF}
 & $1024$ & $\mathbf{112.7}$\eH{5.47} & $\mathbf{93.6}$\eH{5.43} & $88.1$\eH{5.36} & $88.1$\eH{5.36} & $49.4$\eH{5.19} \\
 & $128$  & $118.9$\eH{5.51} & $118.9$\eH{5.51} & $68.3$\eH{5.38} & $68.3$\eH{5.38} & $52.8$\eH{5.20} \\
\addlinespace
\multirow{2}{*}{VP}
 & $1024$ & $119.4$\eH{5.44} & $105.0$\eH{5.42} & $105.0$\eH{5.42} & $105.0$\eH{5.42} & $105.0$\eH{5.42} \\
 & $128$  & $128.7$\eH{5.48} & $128.7$\eH{5.48} & $128.7$\eH{5.48} & $128.7$\eH{5.48} & $128.7$\eH{5.48} \\
\midrule
\multicolumn{7}{l}{\textit{+ self-conditioning ($p = 0.25$)}} \\
\multirow{2}{*}{vMF}
 & $1024$ & -- & -- & -- & -- & $33.0$\eH{5.17} \\
 & $128$  & -- & -- & $\mathbf{43.5}$\eH{5.35} & $\mathbf{37.4}$\eH{5.30} & $27.1$\eH{5.21} \\
\addlinespace
\multirow{2}{*}{VP}
 & $1024$ & -- & -- & -- & -- & $\mathbf{26.3}$\eH{5.17} \\
 & $128$  & $149.3$\eH{5.54} & $149.3$\eH{5.54} & $50.9$\eH{5.35} & $50.9$\eH{5.35} & $50.9$\eH{5.35} \\
\bottomrule
\end{tabular}
\vspace{2pt}
\caption{OpenWebText SDE results at $\mathrm{NFE} \in \{1024, 128\}$: best gen.\ PPL ($H$) per entropy floor; methods that plateau repeat the same value at every floor. Empty cells (--) mark floors that no configuration in the sweep stays above. Best per column in bold.}
\label{tab:owt_sde}
\end{table}

\begin{table}[htbp]
\centering
\small
\setlength{\tabcolsep}{5pt}
\begin{tabular}{lcc}
\toprule
Method & $\mathrm{NFE} = 1024$ & $\mathrm{NFE} = 512$ \\
\midrule
vMF & $33.4$ & $32.6$ \\
VP  & $9.6$  & $9.2$ \\
VE  & $2.8$  & $2.5$ \\
\midrule
\multicolumn{3}{l}{\textit{+ self-conditioning ($p = 0.25$)}} \\
vMF & $\mathbf{42.4}$ & $\mathbf{43.1}$ \\
VP  & $22.6$ & $22.7$ \\
VE  & $3.9$  & $4.5$ \\
\bottomrule
\end{tabular}
\vspace{2pt}
\caption{GSM8K test accuracy ($\uparrow$, \%) after training on TinyGSM, under the SDE sampler at $\mathrm{NFE} \in \{1024, 512\}$. Best per column in bold.}
\label{tab:gsm8k_sde}
\end{table}

Across all three tasks the SDE tracks the predictor-corrector results closely at matched budgets, with $\sigma$ acting as the same quality--entropy trade-off knob as the corrector step size. On GSM8K for example it reaches $43.1\%$ at $\mathrm{NFE} = 512$, matching the predictor-corrector headline ($43.5\%$ at $\mathrm{NFE} = 1024$, Table~\ref{tab:gsm8k_main}). The path ordering is unchanged: vMF leads at nearly every floor and budget, VP requires large $\sigma$ and trails on GSM8K, and VE remains flat.

\begin{table}[htbp]
\centering
\scriptsize
\setlength{\tabcolsep}{3pt}
\begin{tabular}{llllll}
\toprule
Dataset & Method & Floor / budget & Value ($H$) & $\sigma$ & $q$ \\
\midrule
\multirow{9}{*}{LM1B (128)}
 & vMF (TC) & $H \geq 4.34$ / $4.30$ & $136.1$ (4.35) & $0.03$ & $1$ \\
 & vMF (TC) & $H \geq 4.25$ & $77.1$ (4.28) & $0.05$ & $1$ \\
 & vMF (SC) & $H \geq 4.34$ & $70.0$ (4.34) & $0.03$ & $1$ \\
 & vMF (SC) & $H \geq 4.30$ / $4.25$ & $49.7$ (4.30) & $0.05$ & $1$ \\
 & vMF (SC) & $H \geq 4.20$ & $37.2$ (4.24) & $0.1$ & $2$ \\
 & vMF (SC) & $H \geq 4.15$ & $26.9$ (4.18) & $0.1$ & $1$ \\
 & VP (TC) & all floors & $137.3$ (4.34) & $0.2$ & $1$ \\
 & VP (SC) & $H \geq 4.34$ & $118.2$ (4.35) & $3$ & $1$ \\
 & VP (SC) & $H \geq 4.30$ and below & $91.8$ (4.32) & $0.1$ & $1$ \\
\addlinespace
 & VE (TC / SC) & all floors & $145.3$ (4.35) / $160.2$ (4.36) & $0.1$ / $0.003$ & $1$ \\
\midrule
\multirow{10}{*}{OWT}
 & vMF (TC), $1024$ & $H \geq 5.44$ & $112.7$ (5.47) & $0.03$ & $2$ \\
 & vMF (TC), $1024$ & $H \geq 5.40$ / $5.35$ / $5.15$ & $93.6$ (5.43) / $88.1$ (5.36) / $49.4$ (5.19) & $0.03$ / $0.05$ / $0.05$ & $1$ / $2$ / $1$ \\
 & vMF (TC), $128$ & $H \geq 5.40$ / $5.35$ / $5.15$ & $118.9$ (5.51) / $68.3$ (5.38) / $52.8$ (5.20) & $0.03$ / $0.03$ / $0.05$ & $2$ / $1$ / $2$ \\
 & vMF (SC), $1024$ & $H \geq 5.15$ & $33.0$ (5.17) & $0.003$ & $1$ \\
 & vMF (SC), $128$ & $H \geq 5.35$ / $5.25$ / $5.15$ & $43.5$ (5.35) / $37.4$ (5.30) / $27.1$ (5.21) & $0.01$ / $0.03$ / $0.03$ & $1$ / $2$ / $1$ \\
 & VP (TC), $1024$ & $H \geq 5.44$ & $119.4$ (5.44) & $1$ & $1$ \\
 & VP (TC), $1024$ & $H \geq 5.40$ and below & $105.0$ (5.42) & $3$ & $1$ \\
 & VP (TC), $128$ & all floors & $128.7$ (5.48) & $0.1$ & $1$ \\
 & VP (SC), $1024$ & $H \geq 5.15$ & $26.3$ (5.17) & $3$ & $1$ \\
 & VP (SC), $128$ & $H \geq 5.35$ & $50.9$ (5.35) & $0.1$ & $1$ \\
\midrule
\multirow{6}{*}{GSM8K}
 & vMF (TC) & $1024$ / $512$ & $33.4\%$ / $32.6\%$ & $0.5$ / $0.3$ & $2$ / $1$ \\
 & vMF (SC) & $1024$ / $512$ & $42.4\%$ / $43.1\%$ & $0.5$ / $0.2$ & $2$ / $1$ \\
 & VP (TC) & $1024$ / $512$ & $9.6\%$ / $9.2\%$ & $3$ & $1$ \\
 & VP (SC) & $1024$ / $512$ & $22.6\%$ / $22.7\%$ & $3$ & $1$ \\
 & VE (TC) & $1024$ / $512$ & $2.8\%$ / $2.5\%$ & $0.01$ / $3$ & $1$ \\
 & VE (SC) & $1024$ / $512$ & $3.9\%$ / $4.5\%$ & $0.03$ / $3$ & $1$ \\
\bottomrule
\end{tabular}
\vspace{2pt}
\caption{Winning $(\sigma, q)$ configurations behind each SDE cell of Tables~\ref{tab:lm1b_sde}--\ref{tab:gsm8k_sde}. Values that plateau across floors are listed once. VP selects large $\sigma$ throughout GSM8K and on self-conditioned OpenWebText.}
\label{tab:sde_configs}
\end{table}

\FloatBarrier
\section{Inference-Seed Variability}\label{sec:seed_bars}

The configurations behind the main tables were re-run with five generation seeds ($42$--$46$), holding everything else fixed. Seed $42$ is the draw printed in the main text. Table~\ref{tab:seed_bars} reports the mean and standard deviation over the five seeds for $59$ main-text entries.

\begin{table}[htbp]
\centering
\footnotesize
\setlength{\tabcolsep}{5pt}
\begin{tabular}{llcc}
\toprule
Path & Setting & Seed $42$ & Mean $\pm$ s.d. \\
\midrule
\multicolumn{4}{l}{\textit{LM1B, ODE, $\mathrm{NFE}=128$ (Table~\ref{tab:lm1b-comparison})}} \\
vMF & ODE & $171.8$ & $171.4 \pm 2.0$ \\
Geodesic & ODE & $167.9$ & $168.9 \pm 1.5$ \\
VP & ODE & $140.4$ & $139.4 \pm 1.0$ \\
VE & ODE & $149.7$ & $147.7 \pm 1.9$ \\
vMF\,+SC & ODE & $92.4$ & $93.1 \pm 0.8$ \\
Geodesic\,+SC & ODE & $107.1$ & $106.9 \pm 1.5$ \\
VP\,+SC & ODE & $92.3$ & $92.1 \pm 1.3$ \\
VE\,+SC & ODE & $163.4$ & $164.4 \pm 1.4$ \\
\midrule
\multicolumn{4}{l}{\textit{LM1B, PC, $\mathrm{NFE}=128$ (Table~\ref{tab:lm1b_pc})}} \\
vMF & PC, $H \geq 4.30$ & $112.7$ & $112.4 \pm 2.2$ \\
vMF & PC, $H \geq 4.25$ & $66.0$ & $66.2 \pm 0.5$ \\
vMF & PC, $H \geq 4.20$ & $52.4$ & $51.6 \pm 0.5$ \\
vMF & PC, $H \geq 4.15$ & $45.1$ & $44.5 \pm 0.4$ \\
VP & PC, $H \geq 4.30$--$4.15$ & $138.4$ & $139.3 \pm 1.8$ \\
VE & PC, $H \geq 4.30$--$4.15$ & $151.7$ & $151.0 \pm 2.5$ \\
vMF\,+SC & PC, $H \geq 4.30$ & $55.6$ & $56.0 \pm 0.7$ \\
vMF\,+SC & PC, $H \geq 4.25$ & $38.4$ & $38.6 \pm 0.2$ \\
vMF\,+SC & PC, $H \geq 4.20$ & $33.8$ & $33.9 \pm 0.1$ \\
vMF\,+SC & PC, $H \geq 4.15$ & $27.0$ & $26.7 \pm 0.2$ \\
VP\,+SC & PC, $H \geq 4.30$--$4.15$ & $83.1$ & $83.5 \pm 0.3$ \\
VE\,+SC & PC, $H \geq 4.30$--$4.15$ & $164.6$ & $167.2 \pm 1.9$ \\
\midrule
\multicolumn{4}{l}{\textit{OpenWebText, ODE, $\mathrm{NFE}=1024$ (Table~\ref{tab:owt_main})}} \\
vMF & ODE & $139.3$ & $137.9 \pm 1.2$ \\
Geodesic & ODE & $139.0$ & $140.3 \pm 2.7$ \\
VP & ODE & $123.0$ & $124.7 \pm 1.5$ \\
vMF\,+SC & ODE & $36.0$ & $36.0 \pm 0.5$ \\
Geodesic\,+SC & ODE & $36.3$ & $35.7 \pm 0.7$ \\
VP\,+SC & ODE & $36.3$ & $37.0 \pm 0.5$ \\
vMF\,+SC & ODE, delayed onset & $73.9$ & $75.3 \pm 0.8$ \\
Geodesic\,+SC & ODE, delayed onset & $70.3$ & $71.5 \pm 1.0$ \\
VP\,+SC & ODE, delayed onset & $66.2$ & $66.1 \pm 0.7$ \\
\midrule
\multicolumn{4}{l}{\textit{OpenWebText, PC, $\mathrm{NFE}=1024$ (Table~\ref{tab:owt_pc})}} \\
vMF & PC, $H \geq 5.15$ & $51.1$ & $49.9 \pm 0.9$ \\
VP & PC, $H \geq 5.15$ & $70.1$ & $68.0 \pm 1.6$ \\
vMF\,+SC & PC, $H \geq 5.15$ & $32.2$ & $31.7 \pm 0.4$ \\
VP\,+SC & PC, $H \geq 5.15$ & $21.8$ & $22.0 \pm 0.3$ \\
\midrule
\multicolumn{4}{l}{\textit{TinyGSM, $\mathrm{NFE}=1024$, accuracy in \% (Table~\ref{tab:gsm8k_main})}} \\
vMF & ODE & $6.4$ & $6.6 \pm 0.3$ \\
Geodesic & ODE & $5.0$ & $5.3 \pm 0.3$ \\
VP & ODE & $7.7$ & $7.6 \pm 0.5$ \\
VE & ODE & $2.1$ & $2.5 \pm 0.4$ \\
vMF & PC & $33.0$ & $32.9 \pm 0.4$ \\
VP & PC & $12.7$ & $12.1 \pm 0.6$ \\
VE & PC & $1.9$ & $2.4 \pm 0.5$ \\
vMF\,+SC & ODE & $23.7$ & $23.6 \pm 1.0$ \\
Geodesic\,+SC & ODE & $14.9$ & $15.1 \pm 0.2$ \\
VP\,+SC & ODE & $18.3$ & $18.7 \pm 0.8$ \\
VE\,+SC & ODE & $3.5$ & $4.1 \pm 0.4$ \\
vMF\,+SC & PC & $43.5$ & $42.2 \pm 1.0$ \\
VP\,+SC & PC & $28.5$ & $28.2 \pm 1.1$ \\
VE\,+SC & PC & $4.1$ & $4.0 \pm 0.3$ \\
\bottomrule
\end{tabular}
\vspace{2pt}
\caption{Inference-seed variability: mean and standard deviation of gen.\ PPL (TinyGSM: accuracy) over five generation seeds.}
\label{tab:seed_bars}
\end{table}

\FloatBarrier
\section{Qualitative Samples}\label{sec:qual_samples}

\subsection{LM1B}

Decoded LM1B samples (BERT tokenizer, $L = 128$) at $\mathrm{NFE} = 128$, from the runs behind Table~\ref{tab:lm1b-comparison} and Figure~\ref{fig:ppl_vs_entropy}.

\begin{samplecard}{vMF + self-conditioning}

\sampleheader{ODE, warp-aware, $n = 128$}%
{gen.\ PPL = $92.4$, $H = 4.34$.}
\begin{sampletext}
[CLS] [CLS] the office of the consumer agreed that workers must still be in danger of falling incomes. [CLS] opec, which currently has a 25 cent stake in opel, has continued to face up to a " horrisome " auction yesterday that has been dominated by optimism about production costs in recent years with a general bid commitment. [CLS] boutvinges - reid was scheduled to attend the opening hearing friday and fetch a hearing later on wednesday morning. [CLS] but her father, overlooking, told reporters in australia she didn ' t want to find out anything about her medal numbers. [CLS] ( ap ) - alynn communications said thursday that it [CLS]
\end{sampletext}
\begin{sampletext}
[CLS] - time music star, who died monday in a train crash from a orange county, calif., hospital after her real sister was injured. [CLS] he said more than a quarter of asset operations held by shareholders in hbos were " fundamentally uncontrolled. " [CLS] other than just new jersey ' s reflay, mancini is flying to redkna as a potential derby to impress. [CLS] dozens of people are reportedly still dead every day. [CLS] my skills with maximum curiosity and aptitude sabotage affects the utter order were also with greater resonance. [CLS] " the six - hourors " hosts a series of family - dance [CLS]
\end{sampletext}
\begin{sampletext}
[CLS] kids have been putting through something. " [CLS] west yorkshire police said a 17 - year - old man was being treated for very serious injuries after a woman was snatched in st george ' s address in august last year. [CLS] a second trial, which began in may 2007, involves unlawfully fleeting lehman brothers from spain last year, but say mr ijyu is unlikely to stay in line with citigroup ' s mainholders. [CLS] this article was made by glitj global markets ( bst on friday 27 july 2010. bbc home city part al. humek, kan. [CLS] i ' m just a hawk. [CLS]
\end{sampletext}

\end{samplecard}

\begin{samplecard}{vMF + self-conditioning}

\sampleheader{PC, $(32, 3)$, $\varepsilon = 10^{-3}$, $\mathrm{wd}$}%
{gen.\ PPL = $55.6$, $H = 4.30$.}
\begin{sampletext}
[CLS] the most basic surroundings. [CLS] nasoe ' s brilliant strike from tevez gave arsenal a crucial lead but didier drogba, after watching a frustrating equaliser only nine minutes later, sent his side back into the net. [CLS] the franchise retail stores ( www. thewinepark. com / fairfield hd. com ) is a leading consumer electronics retailer of specialty apparel, accessories, apparel, accessories, and accessories. walmart. com offers a wide array of consumer products, including specialtywear, shoes, necklaces, game rings. [CLS] in afternoon trading on thursday, the ftse eurofirst 300 added 7. 58 [CLS]
\end{sampletext}
\begin{sampletext}
[CLS] authorities say suspect umar farouk abdulmutallab and his deputy, umar farouk jabayiri, were terror suspects who had been blamed for amsterdam ' s september 11 terror attacks on yemen. [CLS] mr thomas, 51, from harrow, east london, was awarded the medal. [CLS] wachovia, which has been hit hard by the subprime mortgage crisis, is one of the few banks to step up deposits that make it easier for the borrowers to borrow, a move that could help them buy into their own homes and refinance them in exchange for at least two years. [CLS] sarkozy, who has [CLS]
\end{sampletext}
\begin{sampletext}
[CLS] tribal areas near the border with afghanistan and along pakistan ' s border border. [CLS] " the country will travel to copenhagen this month to travel to london to set up a new strategy to tackle global warming and fight climate change, " he said. [CLS] the obamas ' spending plan will provide up to \$ 114 billion to help them avert large failures in the country ' s financial markets. [CLS] u. s. officials have questioned amiri ' s suggestion that it would not only stop refining oil and gasoline in iran, and help strengthen his government ' s courage and line the international community for imposing sanctions on tehran. [CLS] they sometimes discussed [CLS]
\end{sampletext}

\end{samplecard}

\begin{samplecard}{Geodesic + self-conditioning}

\sampleheader{ODE, warp-aware, $n = 128$}%
{gen.\ PPL = $107.1$, $H = 4.33$.}
\begin{sampletext}
[CLS], egypt has not yet endorsed the arab civil board. [CLS] an authority official said he concluded the bomb was not being fully treated. [CLS] guilgivichs - - whom she later joined from her college training - - fled to sweden after her refusal to take in the equipment. [CLS] the same thing in which everyone wants to stick has spoken. [CLS] church schools in the city are also on the hook to rebunk the fragile social divide between gay and lesbian women and the catholic community. [CLS] ochoa, who started pulling back on a stretcher, got edging on the next hole. [CLS] ( cbs ) a bad air [CLS]
\end{sampletext}
\begin{sampletext}
[CLS] that doesn ' t mean. [CLS] but its identity is unusually low. [CLS] pelosi, reid and feinstein broke talking quickly wednesday to republicans and skeptical that they would not create a \$ 1 billion spending bill enacted at the end of july. [CLS] these rules are bills that would really require. [CLS] the move, inscores the response of lloyds, has triggered concerns almost taken by the credit card agency to an overwhelm the company ' s control. [CLS] after their jabing against the bag, " the likelihood of a single child dropped dramatically, " he explained, only the 49 - year - old smorist [CLS]
\end{sampletext}
\begin{sampletext}
[CLS] on the number of posts who could have been working at children ' s allowance? [CLS] hutesens chose newly groomed secretary - president blika el saytim to have a victory over government - dominated election friday ' s gathering to try to reduce fears that the earthquake could be caused by apparent problems in kenya. [CLS] also years later, those were identified as unfair encountering "orro racer, " whose name is friends of the u. s. and who emigrated to the united states. [CLS] which has experience fuel that by simply not understanding what the chinese principles they would be very careful. [CLS] 12 ( bna ) - iraq [CLS]
\end{sampletext}

\end{samplecard}

\begin{samplecard}{VP + self-conditioning}

\sampleheader{PC, $(16, 7)$, $\varepsilon = 10^{-2}$, $\mathrm{wd}$}%
{gen.\ PPL = $83.1$, $H = 4.33$.}
\begin{sampletext}
[CLS] are usually not to be discrued upon garnett ' s success. [CLS] a document which leads to the rule of speech, such as law and literature, is more more important than any practice or constitution. [CLS] that contest was 6 - 0, against berkeley state. [CLS] two packages each sells for \$ 3. [CLS] the burmese state media appointed suu kyi ' s chief military adviser as the former military director in late 2005 when he was released on trial. [CLS] in fact, the chocolate isn ' t missing by snogakers. [CLS] the discovery was based on other models of advanced data to create safe, fragmented communities [CLS]
\end{sampletext}
\begin{sampletext}
[CLS] ft, the company said that according to the daily telegraph, sir ronnie prepared for his review of financial rules, " recounting " the recent report on the corporation ' s annual council tax package for a maximum of five years which presents a " grave and grave concern for the economy. [CLS] she was confirmed dead in serious condition at the hospital at fort bellalco on oct. [CLS] " they are supported by the external state which can involve people at other disciplines in london. [CLS] new tools will allow employers to make final monthly annuity checks monthly but also, if they don ' t save their husbands or pay off their loan, [CLS]
\end{sampletext}
\begin{sampletext}
[CLS] to the all - star roster. [CLS] he put his face in the lungs and had three broken fingers at his leg. [CLS] " this is not about a prank, " foreign secretary david miliband said on saturday. [CLS] assets bearing price on non - performing interests increased by \$ 5. 5 million to \$ 1. 0 million during the second quarter of 2009. the write - bearing loss on commercial investment banking investments was \$ 17. 8 million to \$ 11. 5 million in the first quarter of 2009. these increase resulted in the decrease of these deferred tax assets during the first nine months of fiscal 2008. the decrease [CLS]
\end{sampletext}

\end{samplecard}

\subsection{TinyGSM}

Generated programs at $\mathrm{NFE} = 1024$ on the full $1{,}319$-problem GSM8K test set, under the configurations of Table~\ref{tab:gsm8k_main}. 

\begin{samplecard}{vMF + self-conditioning}

\sampleheader{PC, $(512, 1)$, $\varepsilon = 10^{-2}$, $\mathrm{w}$, seed $46$}%
{accuracy $= 43.0\%$ ($567/1319$).}
\begin{sampletext}
{\scriptsize\itshape Test item 0, correct.}\\[2pt]
\texttt{def simple\_math\_problem() -> int:}\\
\texttt{~~~~"""}\\
\texttt{~~~~Janet’s ducks lay 16 eggs per day.}\\
\texttt{~~~~She eats three for breakfast every morning and bakes muffins for her friends every day with four.}\\
\texttt{~~~~She sells the remainder at the farmers' market daily for \$2 per fresh duck egg.}\\
\texttt{~~~~How much in dollars does she make every day at the farmers' market?}\\
\texttt{~~~~"""}\\
\texttt{~~~~eggs\_left = 16 - 3 - 4}\\
\texttt{~~~~}\\
\texttt{~~~~revenue = eggs\_left * 2}\\
\texttt{~~~~}\\
\texttt{~~~~result = revenue}\\
\texttt{}\\
\texttt{~~~~return result}
\end{sampletext}
\begin{sampletext}
{\scriptsize\itshape Test item 1, correct.}\\[2pt]
\texttt{def simple\_math\_problem() -> int:}\\
\texttt{~~~~'''}\\
\texttt{~~~~A robe takes 2 bolts of blue fiber and half that much white fiber.}\\
\texttt{~~~~~How many bolts in total does it take?}\\
\texttt{~~~~'''}\\
\texttt{~~~~blue\_bolts = 2}\\
\texttt{~~~~white\_bolts = blue\_bolts / 2}\\
\texttt{~~~~total\_bolts = blue\_bolts + white\_bolts}\\
\texttt{~~~~result = total\_bolts}\\
\texttt{~~~~return result}
\end{sampletext}

\end{samplecard}

\begin{samplecard}{vMF + self-conditioning}

\sampleheader{ODE, warp-aware, $n = 1024$, seed $45$}%
{accuracy $= 23.0\%$ ($303/1319$).}
\begin{sampletext}
{\scriptsize\itshape Test item 0, \emph{wrong} --- returns $-72$, gold $18$.}\\[2pt]
\texttt{def simple\_math\_problem() -> int:}\\
\texttt{~~~~'''}\\
\texttt{~~~~Janet’s ducks lay 16 eggs per day.}\\
\texttt{~~~~She eats three for breakfast every morning and bakes muffins for her friends every day with four.}\\
\texttt{~~~~She sells the remainder at the farmers' market daily for \$2 per fresh duck egg.}\\
\texttt{~~~~How much in dollars does she make every day at the farmers' market?}\\
\texttt{~~~~'''}\\
\texttt{~~~~eggs\_sold = (16 - 3) * 4}\\
\texttt{~~~~eggs\_left = 16 - eggs\_sold}\\
\texttt{~~~~money\_made = eggs\_left * 2}\\
\texttt{~~~~result = money\_made}\\
\texttt{~~~~return result}
\end{sampletext}
\begin{sampletext}
{\scriptsize\itshape Test item 1, correct.}\\[2pt]
\texttt{def simple\_math\_problem() -> int:}\\
\texttt{~~~~'''}\\
\texttt{~~~~A robe takes 2 bolts of blue fiber and half that much white fiber.}\\
\texttt{~~~~~How many bolts in total does it take?}\\
\texttt{~~~~'''}\\
\texttt{~~~~blue\_bolts = 2}\\
\texttt{~~~~white\_bolts = blue\_bolts / 2}\\
\texttt{~~~~total\_bolts = blue\_bolts + white\_bolts}\\
\texttt{~~~~result = total\_bolts}\\
\texttt{~~~~return result}
\end{sampletext}

\end{samplecard}

\begin{samplecard}{Geodesic + self-conditioning}

\sampleheader{ODE, $n = 1024$}%
{accuracy $= 10.8\%$ ($143/1319$).}
\begin{sampletext}
{\scriptsize\itshape Test item 14, correct.}\\[2pt]
\texttt{def simple\_math\_problem() -> int:}\\
\texttt{~~~~'''}\\
\texttt{~~~~In a dance class of 20 students, 20\% enrolled in contemporary dance, 25\% of the remaining enrolled in jazz dance, and the rest enrolled in hip-hop dance.}\\
\texttt{~~~~What percentage of the entire students enrolled in hip-hop dance?}\\
\texttt{~~~~'''}\\
\texttt{~~~~total\_students = 20}\\
\texttt{~~~~contemporary\_students = total\_students * 0.20}\\
\texttt{~~~~remaining\_students = total\_students - contemporary\_students}\\
\texttt{~~~~jazz\_students = remaining\_students * 25/100}\\
\texttt{~~~~hiphop\_students = total\_students - contemporary\_students - jazz\_students}\\
\texttt{~~~~percentage\_hiphip\_hop\_students = (hiphop\_students / total\_students) * 100}\\
\texttt{~~~~result = percentage\_hiphip\_hop\_students}\\
\texttt{~~~~return result}
\end{sampletext}
\begin{sampletext}
{\scriptsize\itshape Test item 15, \emph{wrong} --- returns $-4992.0$, gold $125$.}\\[2pt]
\texttt{def simple\_math\_problem() -> int:}\\
\texttt{~~~~"""}\\
\texttt{~~~~A merchant wants to make a choice of purchase between 2 purchase plans: jewelry worth \$5,000 or electronic gadgets worth \$8,000.}\\
\texttt{~~~~His financial advisor speculates that the jewelry market will go up 2.5\% while the electronic gadgets market will rise 1.2\% within the same month.}\\
\texttt{~~~~If the merchant is looking to maximize profit at the end of this month by making a choice, how much profit would this be?}\\
\texttt{~~~~"""}\\
\texttt{~~~~jewelry\_price = 5000000*2}\\
\texttt{~~~~electronic\_price = 80000*1.012}\\
\texttt{~~~~new\_price = 5000 + electronic\_price}\\
\texttt{~~~~total\_price = 8000/1000 + electronic\_price}\\
\texttt{~~~~profit = total\_price - new\_price}\\
\texttt{~~~~result = profit}\\
\texttt{}\\
\texttt{~~~~return result}
\end{sampletext}

\end{samplecard}

\begin{samplecard}{VP + self-conditioning}

\sampleheader{PC, $(512, 1)$, $\varepsilon = 10^{-1}$, $\mathrm{d}$, seed $46$}%
{accuracy $= 28.6\%$ ($377/1319$).}
\begin{sampletext}
{\scriptsize\itshape Test item 1, correct.}\\[2pt]
\texttt{def simple\_math\_problem() -> int:}\\
\texttt{~~~~'''}\\
\texttt{~~~~A robe takes 2 bolts of blue fiber and half that much white fiber.}\\
\texttt{~~~~~many many bolts in total does it take?}\\
\texttt{~~~~'''}\\
\texttt{~~~~blue\_bolts = 2}\\
\texttt{~~~~white\_bolts = blue\_bolts / 2}\\
\texttt{~~~~total\_bolts = blue\_bolts + white\_bolts}\\
\texttt{~~~~result = total\_bolts}\\
\texttt{~~~~return result}
\end{sampletext}
\begin{sampletext}
{\scriptsize\itshape Test item 3, correct.}\\[2pt]
\texttt{def simple\_math\_problem() -> int:}\\
\texttt{~~~~'''}\\
\texttt{~~~~James decides to run 3 sprints 3 times a week.}\\
\texttt{~~~~~He runs 60 meters each sprint.}\\
\texttt{~~~~~many many total meters does he run a week?}\\
\texttt{~~~~'''}\\
\texttt{~~~~meters\_per\_sprint = 60}\\
\texttt{~~~~sprints\_per\_week = 3}\\
\texttt{~~~~num\_sprints\_per\_week = 3}\\
\texttt{~~~~total\_meters\_run = meters\_per\_sprint * sprints\_per\_week * num\_sprints\_per\_week}\\
\texttt{~~~~result = total\_meters\_run}\\
\texttt{~~~~return result}
\end{sampletext}

\end{samplecard}

\subsection{OpenWebText}

Complete decoded OpenWebText samples (GPT-2 tokenizer, $L = 1024$), from the runs behind Tables~\ref{tab:owt_main} and~\ref{tab:owt_pc} and Figure~\ref{fig:owt_ppl_vs_entropy}; each header gives its own sampler and budget. Each is one unedited sequence of $1024$ tokens, shown in full; \texttt{<|endoftext|>} is the document boundary emitted by the model. Sixteen samples were drawn per setting and one is reproduced here. The cards span the entropy range of the same self-conditioned models: the first two are delayed-onset configurations that hold the data entropy $H_{\mathrm{data}} = 5.44$, the last two sit well below it.

\begin{samplecard}{vMF + self-conditioning}

\sampleheader{ODE, warp-aware, $n = 128$, delayed onset $\tau = 1/3$}%
{gen.\ PPL = $70.7$, $H = 5.45$.}
\begin{sampletext}
{\ttfamily\footnotesize\textless{}|endoftext|\textgreater{}}, but it's expected that next week 100 people will be forced to reunite to the nuclear plant in Fukushima Laeu.

Last June, Japan's Daima Electric Plant experienced a nuclear leak from the Monroe River's northern-east Port. Three days afterwards, the New Zealand Electric Company's Tokichi plant attempted to impueance 15,000 electrical service workers.

The much-caated Fukushima catastrophe also toppled 1,400 jobs.

READ MORE: Sugo in New Zealand, despite New boom firing on

Tamushima will be bankruptcy on Tuesday, more than a year and two weeks after taking in.

"Based on orders received from one company active, Japanese Ministry of Defence, there is a planned shutdown regarding the Tok nuclear power plant operation," the company said in a statement.

"The shutdown is very effective. Severe alarm systems will continue to operate in the course of days, although additional stations will be shut down wherever appropriate.

"We take a precaution for incidents to affect the safety of serviceages and prevent disruption of our radio networks."

There would be "no limit on whether a grid construct makes short malfunctions and Japan will do everything it takes to ensure our power safety and levels security".

Japan spokesman Adrian
The Earth Group, who represents the environment and an environment group campaigning for nuclear regulation, said on Tuesday the plant released gigigrams of nuclear water, causing the loss of 2,700 lives.

READ MORE: Fukushima warns in Fukushima Laeu plant

Mr Gani released a statement on Wednesday saying that "defcipled parties are sworn to blame".

"Our tribe and colleagues were appointed for the democratic votes, while others may have failed popular campaigns for atomic power construction already signed.

"Our peoples rely on telling the facts to make decisions, which is why we are at the heart of this tragedy running up as it seems is the thousands of deaths that have been reported across Australia and Europe.

“Our indigenous staff in the country have recently informed others we could do our part to finish, design, engineer, repair or mitigate this accident."

“Some of the financial studies anticipated in the work will be prescribed red leather faces and hardware over time and will be added to public order in perhaps few weeks."

Mr Gani said he was "poursing to get as much support here as possible from the New Zealand Liberal government".

"We will act here to help against this, and this impoverishes this island," he said. “To us, as everybody else, we’ve always been a people, and we’re proud to be a part that’s a New thing.”{\ttfamily\footnotesize\textless{}|endoftext|\textgreater{}}“I’m proud to be heard,” - Donald Donald Trump

Hundreds of thousands of ad members attended a Jupiter, Florida fundraiser and tickets to Donald J. Trump seeking hearings range from \$1,500 to \$10,000, according to WSB-TV.The Jupiter fundraising sign-up was buzzing: “Our Tickets Go Nice to support our fellow Chovers does not support Donald Trump.”Unfortunately, the crowds were overflowing in the Byzantine fundraising Hall.“Many people spend time with family and want to hear Inside trust about their life and sit in the mirror, to who plays and what they love,” said Susan Marallova, Florida fundraiser chairman.The fundraiser quoted the former entrepreneur: “We’re aware of a lot of big people to us as a donors. They like to hear from each other, and we’re always proud to be heard.” (Watch ABC News)“And if you’re a political example if you run into a [...] plan Social Security, we’re also going to make you pro-life. So you have to make sure your values are strictly,” according to the 68-year-old.When Does Trump admitted that he thought he'd use the fundraiser a lot of cash and decided to shell out Trump junior "promises," he said, “But we don’t make a lot of money.”{\ttfamily\footnotesize\textless{}|endoftext|\textgreater{}}Scientists wonder how arms could be applied to a human robots

We could pull the challenge if we could interpose humans into another Predator

Once in life, humans are able to recognize or recognize objects from their materials. To deal with this challenge is to make the robot a human.

Earlier this year, researchers at the Google UCSF National Institute for Advanced Science demonstrated the need to patent a robot-in-lectrol chip to humans recognize more objects like their arms via an accelerated brain operation by their brains.

E' Xiao and his undergraduate colleagues, a Research Science in Biology for the Robotics and Engineering Science Management Office at Carnegie Mellon University, predict current developments in Autonomous A Human (VID). The findings were published last month in the journal Science (Feb. 6{\ttfamily\footnotesize\textless{}|endoftext|\textgreater{}}
\end{sampletext}

\end{samplecard}

\begin{samplecard}{vMF + self-conditioning}

\sampleheader{PC, $(32, 3)$, $\varepsilon = 10^{-4}$, $\mathrm{wd}$, delayed onset $\tau = 1/3$}%
{gen.\ PPL = $52.7$, $H = 5.44$.}
\begin{sampletext}
{\ttfamily\footnotesize\textless{}|endoftext|\textgreater{}}

Criticism [ edit ]

For conspiracy scientists, there is a fundamental difference between admitting and examining several more explanations and not logically distinguishing what type of findings one feels as authentic. Thus, the proponents of conspiracy theories should abandon neither religious or religious truths and selectively believe the truth is all being "telling".[1][2][3][4] They must not counter the idea that conspiracy theories are fringe theories, or beliefs away from them. Inspire scientific analysis distinguishes conspiracy theories based on science. As a result, it is possible to reconcile their beliefs in a clinical manner.[5]

Psychic professionals, clinicians or even representatives of a general medical team or psychiatric practitioner should be reminded that you must practice or consideration conspiracy theories in order to make informed informed decisions prior to events contained in the evidence being clearly presented. Readers should also agree with the decision based on their beliefs, background, beliefs, and theories of the conspiracy suspects.[6]

Religion [ edit ]

Islamians feel many Muslims embrace conspiracy theories placing doubt into the "wrong" religion.[6][9]

Religion [ edit ]

When conspiracy theorists are asked to explain their sectarian beliefs, values and the regime, it is not to say that a Muslim holds an allegiance to Islamic State.[10]

Pro-cychvery [ edit ]

Contivants believe conspiracy theories focus on science rather than evidence.[10][12] This reasoning is not limited. The main problem with conspiracy theories is that they cannot match credible evidence publicly. For example, Internet searches maintained by the 2017 Federal Bureau of Investigation contain over 900 conspiracy theories and conspiracy theories.[13][13] Federal authorities report millions of them each year.[13] Furthermore, they lack the so-called "logic fallacy"[13] which states that a fallacy fallacy addresses a "unexpected number" possible lengths to disproport "facts from the sources".[16] Some websites promote "rastery claim" when people believe "that conspiracy theories based on false evidence have conflicting viewpoints".[17]

Random evidence shows the success of conspiracy theories on the internet.[17] A significant number of websites rely on experts to further verify such claims, because conspiracy theorists believe that "stages should only be given to read something big".[19] Psychological conspiracy theories have also been dubbed by the media "ill-informed".[21]

Pro-systematic theories [ edit ]

Popular, secular and statistical beliefs are perhaps most popular pedavaging conspiracy theories on concepts almost exclusively.[22] Although the conspiration is true, the theories tend to respectfully agree with belief ideas.[20][23][24]

For example, Sir Mamhan Kealday Infar, a 13-year-old school student, was allegedly attacked in Amrikhar district in 1999 by the terrorist group "dhhan Shirab" and murdered several girls.[23] The above article argues that pseudipcience values are pre-corrected and suggests that it is better to accept these pro-systematic, world-conpetual conspiracy theories.[27]

A new body of research from 1998–2010, which classified conspiracy theories as "confidence-mining", changed the term "skeptology" to "independentular science".[28][30][31] Conspiracy theories are esteemed among scientists such as psychologist Ralph Ross.[30][32][33][34] Conspiracy theories are more popular than among non-religious religions such as Islamic Zoroidhs.[32]

Religion in the United States [ edit ]

Religion and conspiracy theories also popularize regularly. According to a Demosium Group which polled voters in the United States during the preceding 2000 election, 60\% of Americans believe that these conspiracy theories "are too harmless for all Americans".

Foci theories \& non-religious beliefs [ edit ]

A 2013 Gallup poll survey showed the majority of Americans as wary of conspiracy theories as either secular or faithless. 72\% believe all children and 66\% believe in science:[9]

What you should say is that I'm putting on all who the faithly are...I know, because I believe in my own belief, that from death you live the belief you believe comes far from me. But I don't from religion - everything I believe comes far from nature because you choose all the facts."[38]

A 1998-15 survey conducted by the Gervaeb Centre for the Religious Science reported an "open grade on possible belief"[62] with 219 reporting that there was at least no connection between belief in children and adult science.[19][full citation needed]

In 2013 the American Academy of Sciences for Children and Kidcrators stated that there was no connection between belief in children and infants or in human beliefs.[89][11]

Those who believe in conspiracy beliefs and experience difficulties in babies' ability to understand science are more likely to adopt conspiracy theories due to the following:

attention and physical differences

honderance{\ttfamily\footnotesize\textless{}|endoftext|\textgreater{}}
\end{sampletext}

\end{samplecard}

\begin{samplecard}{VP + self-conditioning}

\sampleheader{PC, $(128, 7)$, $\varepsilon = 10^{-2}$, $\mathrm{wd}$}%
{gen.\ PPL = $21.8$, $H = 5.16$.}
\begin{sampletext}
{\ttfamily\footnotesize\textless{}|endoftext|\textgreater{}} far as it goes. We still didn’t have enough information to get into this whole thing exactly what we could have been of the two that hadn’t, like, read up on a scrap that didn’t all wrap up. We didn’t have to worry about my brothers’, either. So we, we didn’t have to worry about the papers of all of that.”

But the scrap cracked and he looked down at Martin, stunned. “uuaaaaaya! He was traded in the mid-30s. Obviously he got seriously hurt, we’re gonna have a problem with catching the pain a lot on his wrist, a little!”

“Yeah! Joel was right,” said Patrick, banging on her phone. “What would you do about that?”

Grind Huslan’s face was placed on a finger. “You’re such a man. I never asked to do that. I made him bruise. There was a guy who was gonna shoot twenty-five or six. You weren’t gonna get paid for it. You didn’t use the weapon. He was gonna kill sixteen.”

Steven rolled back. “That’s the dumbest thing I’ve ever known.”

“No,” said Gerry. “Maybe is that wrong with killing?”

“Oh, I’m not aware of it, Billy, but these guys have no problems with it—it’s a bit wrong with killing, aren’t they?”

“Yeah, Jeff Tummy, who shot twice,” Patrick said dismissively, “said, ‘I shot twenty-teen, I don’t want to tell you about it.' Her bruizer raised her gun at Connie and shouted, ‘I shot twenty-seven, eight—how did those guys do that? You shot twenty-three. I shot five and five. Fuck Chuck Huslan’s memory!”

Tummy looked at Kelly. “Did that? I shot sixty. I stole 20 pounds of meat, like I didn’t do that for any reason. These guys didn’t do this either.”

Sabotita frowned then. Kelly leaned into the window, slamming down the window.

“Why did you say that?” Mitchell stun pointed at Connie in an agitated manner. “Why didn’t they say something like that?”

“Did you say something like that?” Connie asked.

“I won’t give credit,” Jim said.

“Didn’t you give credit.”

Gosh, Chuck had nothing to do with that. Harry Lamaine had nothing to do with him.

“He hit me in the back,” Tummy quused. “Ay-whooly. How did Billy get shot twice?”

“Nothing,” Graham laughed, crying. “I wanted to kill him! Not all of this shit would really red scratch my character. My crew and nobody even wanted to stop him!”

“I don’t want to lock him up, fear cutting—I love catching!”

“Keeping that in mind. These guys were cool,” said Kelly. “I’m gonna do fine. Not bad for shooting these guys.”

“That sounds fucking silly,” said Connie.

“That’s too much.” He paused and took a turn to chant loudly. “That’s too big to go by.”

Kelly swung his fist at Sherrod, who tapped the knife again.

“Throw his gun.”

Centley stepped out of his way, “Jon Barteo’s a badass, right now!”

Mitchell looked at Sherrod’s cart. Timward three co-workers took heads, both hurled at each other. Tummy started to each other, joined by Sherrod’s jeer, smirking gunsters and guts.

“Listen!” continued Connie, “he’s the gun, okay? I’ve got no control over what I want.”

“Wait, isn’t that what I just said?”

Mitchell gave Mitchell a nod, leaning back at Tummy. “I’ve got no control. I’m having good at home. Good Christmas stores are really where people would buy it.{\ttfamily\footnotesize\textless{}|endoftext|\textgreater{}}
\end{sampletext}

\end{samplecard}

\begin{samplecard}{Geodesic + self-conditioning}

\sampleheader{ODE, warp-aware, $n = 1024$}%
{gen.\ PPL = $36.3$, $H = 5.23$.}
\begin{sampletext}
{\ttfamily\footnotesize\textless{}|endoftext|\textgreater{}} so let's really draw you on her thoughts on all of the things Manziel brings to the franchise.

Podcast: Jon Grick, Eric McCllethe: That's a National Report dynamic, and all of this history is celebratory, but what do you focus on?

Strkinger Well, man. You know, we don't care. These are things people all know. We've been following for a while. It's pretty important that we react to those comments as well.

Strkinger: Like, Marty is the kind of quarterback, he's a burner, he's a playmaker \#4.

But we talk about Johnny. Because Johnny knows what he knows, he sees a great role, but he's here.

And you're starting to see what he lacks or nobody else.

But he's a good, man. He's a runner, but he has every opportunity. He says he's a Wonderlinger or not a pass rusher.

And that's why you're starting to see him.

That's why you're starting to see him.

And that's why you guys are starting to see him.

Now, this is the reason Michael Manziel runs no.

You can see him, but he runs unnestly.

He seems to give off Tom Roethlisberger in a way. Manziel will build the quarterback that you all know, but if he knows who he'll be, it's tough to see him improve.

So when he comes out, deeper Manziel is a tough, athletic, kid. He knows nobody knows what he is.

And he's a tackling, man. He's a quarterback, but he has talent.

Rate's a burner. And he's a playmaker \#4. But he's dynamic, strong-type kid. He knows what's special, it's tough to see him play well.

Now, every minute he comes out. He's a great, athletic, kid.

And he's a burner. And he's a playmaker \#4.

He's a burner, but he definitely knows who he can be.

Or. Johnny Manziel is the kind of quarterback. He's tough, man. He's a burner, but he has talent.

And that is inseparable.

He does this. He's been an explosive. And in Manziel's life, he'll be very great. And he's going to excel in high school.

But that is inseparable.

But does he ever have to work his way. It doesn't mean he knows he is athletic and yet he knows he's total and finally smart down. That doesn't mean he needed a chance to rush because he's not anything. He's just tools. He still has to play one-on-one.

And that is inseparable.

He does this. He's been an explosive. And in Manziel's life, he'll be very great. And he's going to excel in high school.

But that is inseparable.

Well, does he do everything as quarterback. He didn't open up. And he's trying to play in a spectacular way.

This guy really fits. And he's been in the open position. Average, outstanding rating. He just wants to see those suitsers.

Does Johnny Manziel have everything around. He's all around. He just wants to see those suits.

Well, does Manziel have everything around. He does everything as quarterback. He didn't open up. And he is trying to play in a spectacular way.

This guy really fits. He's been in the open position. Average, outstanding rating. He just wants to see those suitsers.

And he's doing everything. And he's doing everything. And I think he's going to have an incredible receiver. And he's going to strip and a great top 1 threat. And it's not all he scares, but he's going to have an amazing receiver. And he's going to need some wide receivers for him in the offseason. And that's going to be a solid receiver.

And I'd hate to hear right now how the Colts use the combination of the AFC mix at the same time. He's going to need something bigger than Peyton Manning, he can take on all the plays and the competition. And that's going to be in the mix. And it's not all he gets. Incover and great top 1 threat. And I think he's going to have an easy defense. And he'll be a threat to everybody, but he's going to need some wide receivers for him in the offseason. And he's going to be a solid receiver. And that's going to counterbalance over his upside, which anyone is aware. And that's going to be in the mix.

Hell, I'd{\ttfamily\footnotesize\textless{}|endoftext|\textgreater{}}
\end{sampletext}

\end{samplecard}

%\newpage
%\input{checklist.tex}

\end{document}